\documentclass[10pt,twocolumn,letterpaper]{article}

\usepackage{iccv}
\usepackage{times}
\usepackage{epsfig}
\usepackage{graphicx}
\usepackage{amsmath}
\usepackage{amssymb}

\usepackage{tikz}
\usepackage{comment}
\usepackage{color}
\usepackage{multirow}
\usepackage{algorithm}
\usepackage{algorithmic}
\usepackage{bbding}
\usepackage{xspace}
\usepackage{graphicx}
\usepackage{graphics}
\usepackage{multirow}
\usepackage{epstopdf}
\usepackage{epsfig}
\usepackage{url}
\usepackage{subcaption}
\usepackage{booktabs}
\usepackage{enumitem}

\DeclareMathOperator*{\argmin}{arg\!\min}

\newtheorem{theorem}{Theorem}[section]

\newtheorem{property}[theorem]{Property}

\newenvironment{proof}[1][Proof]{\begin{trivlist}
\item[\hskip \labelsep {\bfseries #1}]}{\end{trivlist}}

\newcommand{\qed}{\nobreak \ifvmode \relax \else
      \ifdim\lastskip<1.5em \hskip-\lastskip
      \hskip1.5em plus0em minus0.5em \fi \nobreak
      \vrule height0.75em width0.5em depth0.25em\fi}
\newcommand{\Mark}[1]{\textnormal{\textsuperscript{#1}}}

\usepackage[pagebackref=true,breaklinks=true,letterpaper=true,colorlinks,bookmarks=false]{hyperref}

\usepackage[capitalize]{cleveref}
\crefname{section}{Sec.}{Secs.}
\Crefname{section}{Section}{Sections}
\Crefname{table}{Table}{Tables}
\crefname{table}{Tab.}{Tabs.}

\iccvfinalcopy 

\def\iccvPaperID{11639} 

\ificcvfinal\fi

\begin{document}

\title{Neural Vector Fields for Implicit Surface Representation and Inference}

\author{Edoardo Mello Rella\Mark{1} \quad Ajad Chhatkuli\Mark{1} \quad Ender Konukoglu\Mark{1} \quad Luc Van Gool\Mark{1,2} \\
\Mark{1}\,Computer Vision Lab, ETH Zurich \quad \Mark{2}\,VISICS, KU Leuven\\}

\maketitle
\ificcvfinal\thispagestyle{empty}\fi

\begin{abstract}
   Implicit fields have recently shown increasing success in representing and learning 3D shapes accurately. Signed distance fields and occupancy fields are decades old and still the preferred representations, both with well-studied properties, despite their restriction to closed surfaces. With neural networks, several other variations and training principles have been proposed with the goal to represent all classes of shapes. In this paper, we develop a novel and yet a fundamental representation considering unit vectors in 3D space and call it Vector Field (VF): at each point in $\mathbb{R}^3$, VF is directed at the closest point on the surface. We theoretically demonstrate that VF can be easily transformed to surface density by computing the flux density. Unlike other standard representations, VF directly encodes an important physical property of the surface, its normal. We further show the advantages of VF representation, in learning open, closed, or multi-layered as well as piecewise planar surfaces. We compare our method on several datasets including ShapeNet where the proposed new neural implicit field shows superior accuracy in representing any type of shape, outperforming other standard methods. Code is available at https://github.com/edomel/ImplicitVF.
\end{abstract}

\section{Introduction}
Representing 3D surfaces efficiently and conveniently has long been a challenge in computer graphics and 3D vision. A representation for 3D surfaces should be accurate, while being suitable for downstream tasks. Shape analysis, such as 3D shape correspondences~\cite{zheng2021deep,groueix20183d,litany2017deep}, 3D registration and deformations~\cite{yifan2020neural,ovsjanikov2012functional,eisenberger2020smooth}, or generation~\cite{rozen2021moser,chan2022efficient,or2022stylesdf,groueix2018papier,hui2022neural} rely on having the right representations. Recently, some of these questions were answered with implicit neural representations (INRs)~\cite{park2019deepsdf,mescheder2019occnet,saito2019implicitbin3,chen2019implicit}, which use the classical signed distance~\cite{Osher1988-fronts,osher2003level} or binary occupancy fields, learned with neural networks.
Implicit fields are also used in Neural Radiance Fields (NeRFs)~\cite{mildenhall2020nerf,yariv2021volume,yu2021plenoxels}, surface representation from point clouds~\cite{atzmon2020sal,ben2022digs}, image synthesis and processing~\cite{Shaham_2021_CVPR,anokhin2021image,chen2021learning,dupont2021coin}, and many others.

\begin{figure}[!ht]
\centering
\begin{subfigure}{0.9\linewidth}
    \includegraphics[width=\textwidth]{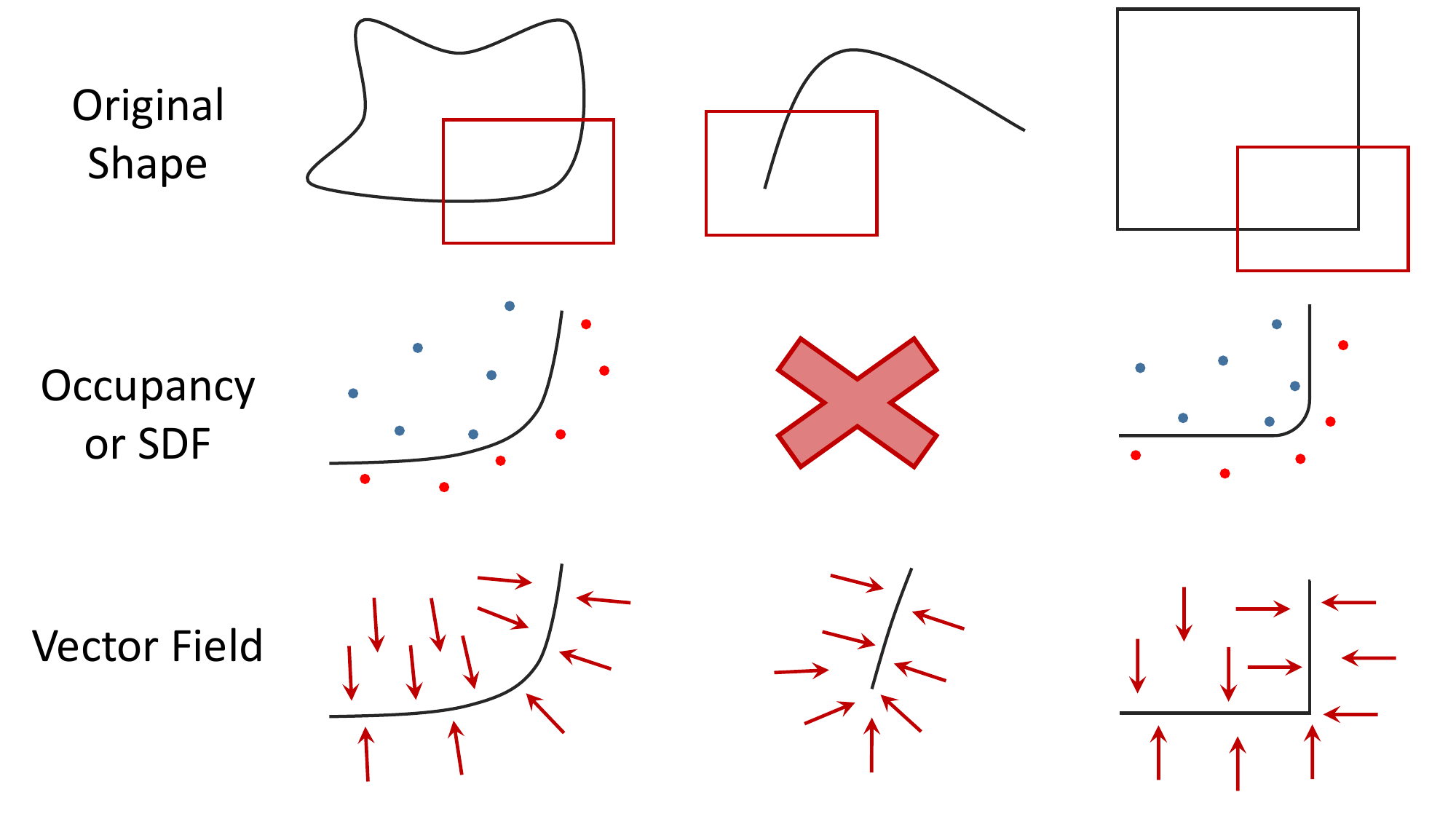}
\end{subfigure}
\caption{Vector Field (VF) visualization on 2D surfaces. Each column shows the shape to represent in black, and a zoomed in sample reconstruction of the surface inside the red box. VF can represent shapes similar to SDF or binary occupancy, but also open surfaces. As the surface normals are directly encoded, VF can reconstruct very sharp angles through additional regularization when needed.}
\label{fig:teaser}
\end{figure}

\begin{figure*}[t!]
\centering
\begin{subfigure}{0.19\linewidth}
\includegraphics[width=\textwidth]{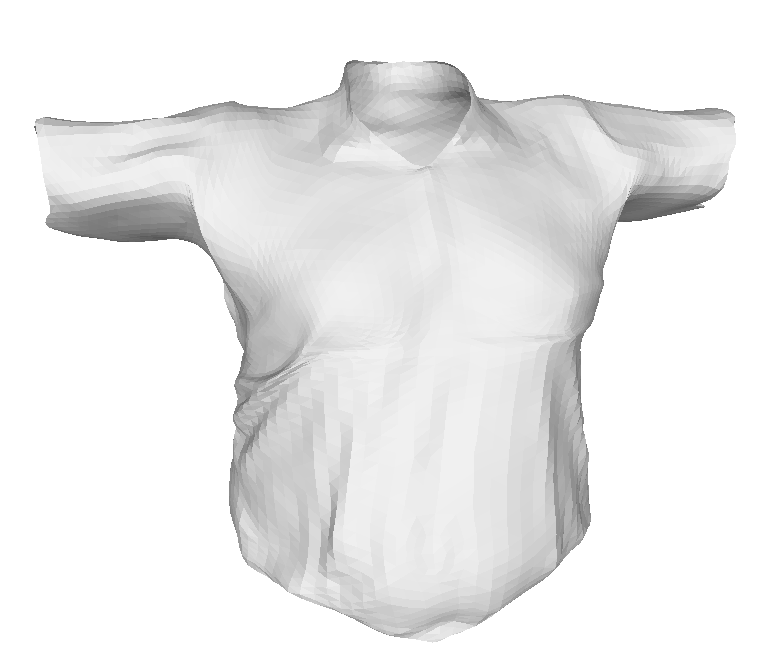}
\caption{Original Mesh}
\end{subfigure}
\hspace{0.03\linewidth}
\begin{subfigure}{0.20\linewidth}
\includegraphics[width=\textwidth]{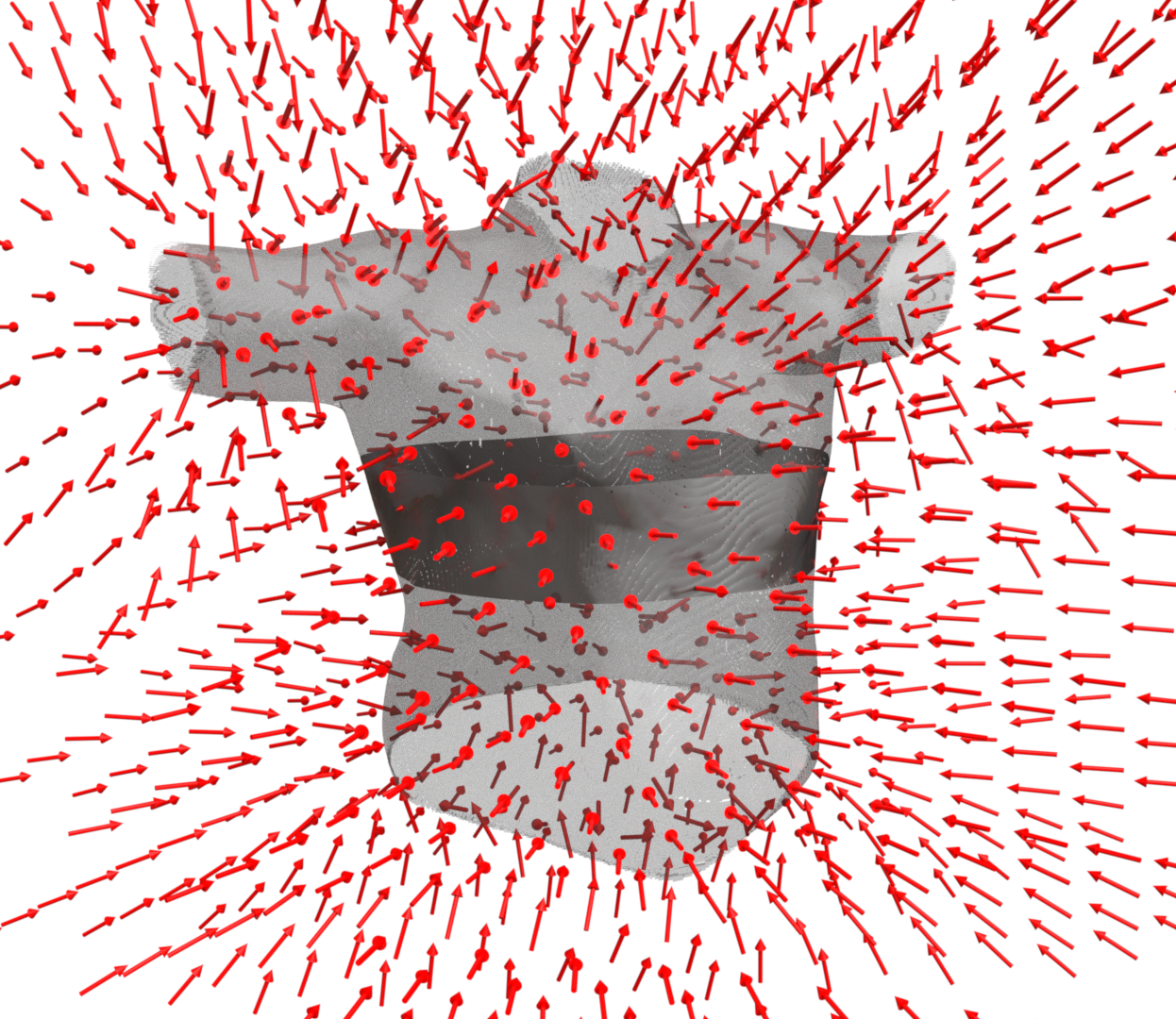}
\caption{VF prediction}
\end{subfigure}
\hspace{0.03\linewidth}
\begin{subfigure}{0.20\linewidth}
\includegraphics[width=\textwidth]{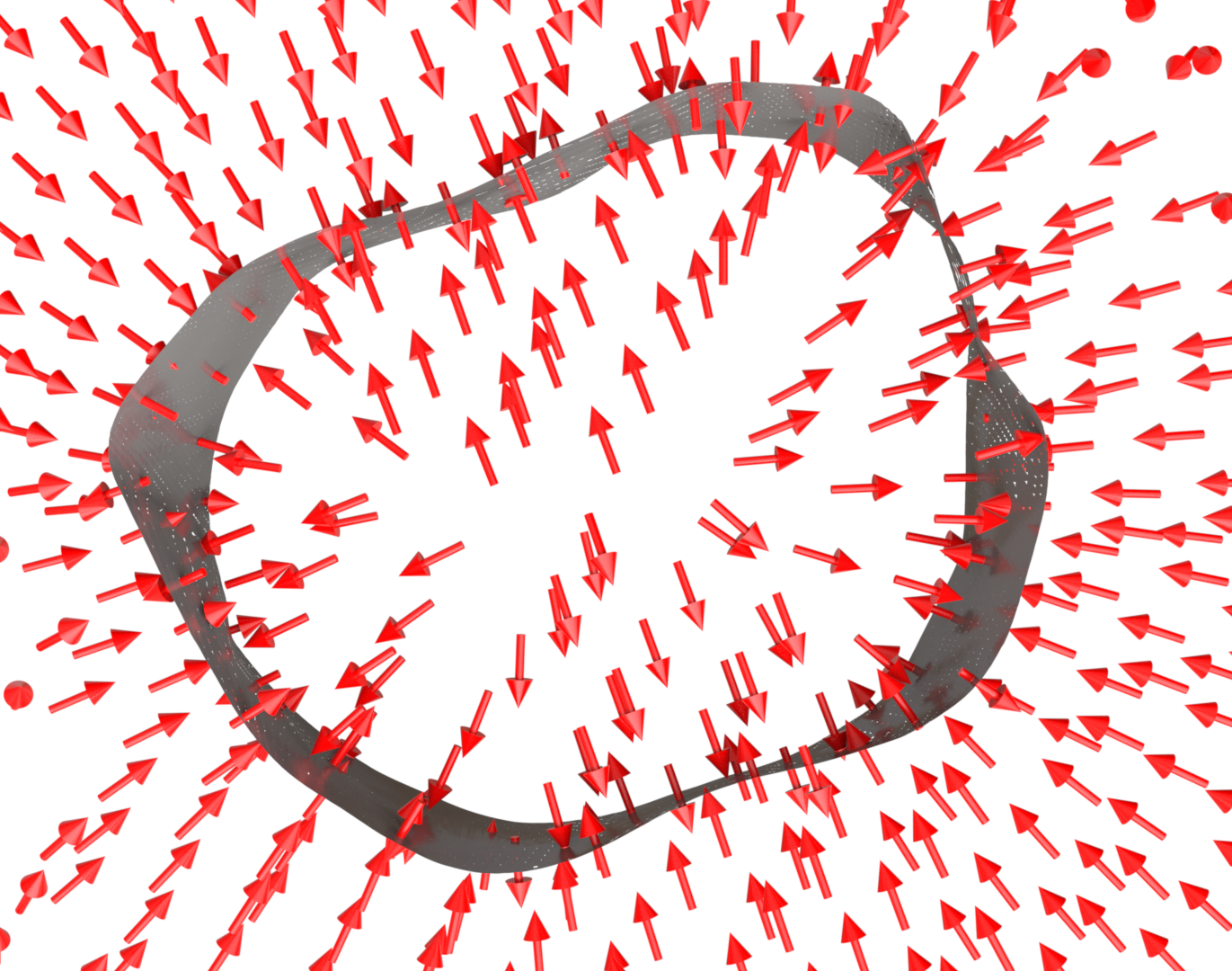} 
\caption{Local VF crop}
\end{subfigure}
\hspace{0.03\linewidth}
\begin{subfigure}{0.19\linewidth}
\includegraphics[width=\textwidth]{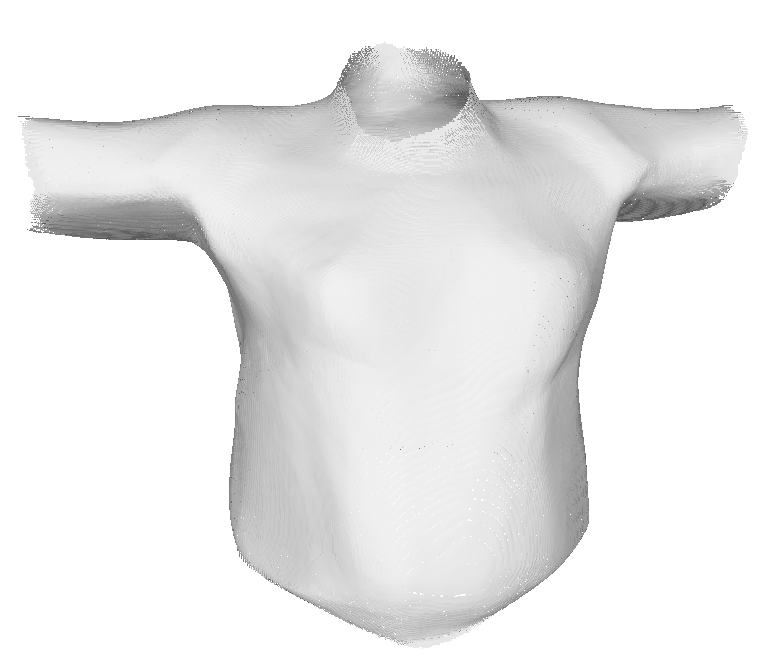} 
\caption{Reconstructed Mesh}
\end{subfigure}
\caption{Overview of our proposed Vector Field representation. Left to right we show the ground truth mesh, the predicted VF field and a zoom-in of a crop of it, and finally, the reconstructed object after running Marching Cubes on the prediction.}
\label{fig:open_surface}
\end{figure*}

Two different types of implicit representations are commonly used. Distance based representations~\cite{park2019deepsdf,chibane2020ndf} associate to each point in space the distance from the closest surface of the object. This is either the signed distance field (SDF)~\cite{park2019deepsdf,saito2019implicitbin3,chen2019implicit} or the unsigned one (UDF)~\cite{chibane2020ndf}, with the former that gives a negative sign to points inside the object and positive outside. The second type of representation is binary occupancy~\cite{mescheder2019occnet}, where each point in space is classified to be either inside or outside the object.
In both cases, a scalar field $f(\mathsf{x})$ is learned to represent the surface at a specific level $\lambda$.
In general, SDF is the prevalent representation used in shape correspondences~\cite{zheng2021deep}, NeRFs~\cite{yariv2021volume}, geometric regularization~\cite{atzmon2020sal,atzmon2021sald,ben2022digs}, etc. Recent work has proposed to supplement SDF with gradient information~\cite{sommersang2022}, which adds regularization to surface reconstruction using the normals. Similar work also uses binary occupancy fields~\cite{niemeyer2019occupancy,lei2022cadex,mueller2022instant}. UDF was recently used to learn open surfaces~\cite{chibane2020ndf}, with the downside being a much more complex meshing process.
Alternative strategies, such as ray-based sampling methods~\cite{Feng2022} and pairwise point sampling~\cite{Ye_2022gifs} solve open surface representation using a very different approach~\cite{Feng2022} or added training/architectural complexity~\cite{Ye_2022gifs}.

We introduce a novel field for shape representation analogous to SDF. Inspired by image boundary representations in \cite{rella2022zero}, we map each point in the $\mathbb{R}^3$ 3D space to a unit vector pointing towards the closest surface point. We call such representation Vector Field (VF). We mathematically demonstrate that the surface points can be recovered at the $-1$ level set of the flux density of VF. The proposed vector field allows us to represent both open and closed shapes and can be used for surface regularization, by directly encoding the normals. Instead, in standard implicit representations, normals can only be obtained through a differentiation step~\cite{guo2022neural,ben2022digs}. We show the potential application of normals to the representation of piecewise planar surfaces through VF, thus improving the accuracy for certain object classes. Moreover, we perform experiments and show comparisons with VF on the standard ShapeNet~\cite{cheng2015shapenet} subsets with additional classes as well as on open clothes surfaces from the MGN dataset~\cite{bhatnagar2019multi}. We obtain state-of-the-art results for surface representation on point accuracy.
In summary our contributions are threefold:
\begin{itemize}
    \item We propose a vector field, VF, for implicit representation of 3D shapes with mathematical proofs and empirical evidence for its representation capability.
    \item We show that the surface normals directly encoded in VF can be used to efficiently represent piecewise planar objects.
    \item We perform extensive tests for standard and general shapes, showing the strong performance of our method, achieving superior accuracy and generalization in multiple testing set-ups.
\end{itemize}

\section{Related Work}

We briefly review modern implicit functions for shape representation and analysis, followed by discussions on different modifications of the original approaches.

\paragraph{Standard representations.}
When comparing to traditional voxel-based representations for 3D shapes, the INR~\cite{park2019deepsdf,mescheder2019occnet,chen2019implicitsdf1,saito2019implicitbin3} have been able to solve, to a large extent, the high memory issues and limitations on downstream tasks~\cite{zheng2021deep,Xiu_2022_CVPR,yariv2021volume}. Recent methods in INRs generally either use SDF~\cite{park2019deepsdf,chen2019implicit,zheng2021deep} or binary occupancy~\cite{mescheder2019occnet,niemeyer2019occupancy,lei2022cadex}, with very little consensus on where and how they differ. SDF and binary occupancy have the advantage that they represent closed surfaces and generate watertight meshes with the fast Marching Cubes (MC) algorithm. However, they cannot represent general shapes as \emph{open surfaces}, non-manifold structures, or multi-layered objects cannot be defined using the inside-outside separation. An equally fundamental representation, the distance field~\cite{Osher1988-fronts}, was proposed as a solution for INR of open surfaces~\cite{chibane2020ndf}. On the other hand, \cite{chibane2020ndf} achieves it by forgoing quick and simple mesh inference by MC. A recent work~\cite{Ye_2022gifs} tackles open surface representation through classifying whether point-pairs are on the same side of the surface; it enables fast inference but at the cost of increased training complexity. In some cases, the standard INRs are augmented with positional encoding~\cite{mildenhall2020nerf,tancik2020fourier} or sinusoidal activations~\cite{sitzmann2020implicit,ben2022digs} replacing ReLU layers in the neural architecture. 

\paragraph{INR in NeRF and 3D Reconstructions.}
Implicit fields are also used to construct NeRFs~\cite{mildenhall2020nerf,yariv2021volume}, where the 3D surface is defined by a term called volume density $\sigma$. However, these methods are trained for accurate view synthesis rather than scene representation. Several works~\cite{yariv2021volume,guo2022neural,ueda2022neural} have proposed the use of a proper SDF in NeRF in order to improve the implicitly represented 3D scene. On the other hand, a ray-based implicit representation~\cite{Feng2022} was recently proposed, inspired from the so-called light field networks~\cite{sitzmann2021light}. Another recent work~\cite{sommersang2022} proposes to use SDF along with its gradient for the task of 3D reconstruction. A very similar approach is presented in \cite{venkatesh2020dude}. Even though in previous work~\cite{ben2022digs,atzmon2021sald} the gradient information is used for surface regularization, it does not contribute to localize the surface, contrary to what we propose.

\paragraph{Grid-based representations and modifications.}
An important class of 3D shape representations uses the voxel grid, which is essentially the discretized binary occupancy. \cite{ji2017voxel1,kar2017voxel2} directly extended the pixel-like image structure to volumetric data and were used with straightforward extensions to 3 dimensions of image processing techniques, such as convolutions. However, traditional voxel-based methods scale cubically with the resolution in terms of memory usage and computation. Therefore, the first methods proposed \cite{choy2016voxelsmall1,tulsiani2017voxelsmall2,wu2015smallvoxel3} could only work with $32^3$ voxel grids. It was later improved to $128^3$ \cite{wu2017bigvoxel1,zhang2018bigvoxel2} with drawbacks in terms of network sizes and training speed. Recently, voxel-grids are used in implicit representation, particularly in NeRF, by using octree-based sparse voxels~\cite{liu2020neural}, sparse spherical harmonic functions on the grid~\cite{yu2021plenoxels}, voxel occupancy and hashing~\cite{mueller2022instant} or simply interpolating voxel occupancy~\cite{sun2022direct}. Such grid-based interpolation and proxy functions may help improve the standard representations discussed above.

\section{Method}
We first formally define VF and list the properties which make it suitable for the representation task. We outline how it can be used in practice and compare its properties to other fundamental representations. Finally, we show how using VF allows us to impose a planar prior on the predicted shapes, which is useful for several object/scene classes.

\subsection{Vector Field Representation}
\label{sec:prop}
Suppose we have a 3D object or scene in a subset of 3D space, $\Omega \subseteq \mathbb{R}^3$, embedding the set of 2D surfaces, $\Pi \subset \Omega$, with $\mathsf{x}_S \in \Pi$ a surface point. To define the VF representation, we map each point $\mathsf{x}\in\Omega$ to the normalized direction $\mathsf{v}\in\Gamma\subset\mathbb{R}^3$ towards the closest surface point $\mathsf{x}_S$. Given the definition, the points $\{\mathsf{x}_S\}$ are discontinuities in the field where the direction of $\mathsf{v}$ converges and flips. At the surface points, the field $\mathsf{v}$ is mapped arbitrarily to either of the two opposite directional normals of the surface. Generalizing to non-differentiable surface points where no surface normal is defined, this is equivalent to treating the point $\mathsf{x}_S$ as a point $\mathsf{x}' = \mathsf{x}_S + \mathsf{\epsilon}$ s.t.\ $\mathsf{x}' \notin \Pi$ and $\Vert \mathsf{\epsilon} \Vert_2 \to 0$ and computing the field values for point $\mathsf{x}'$.

The VF representation is thus the mapping from positions in space to directions $f : \Omega \to \Gamma$. Figure \ref{fig:open_surface} visualizes our representation for an example object. We define the VF representation formally as follows:
\begin{align}
\begin{split}
    f(\mathsf{x}) &= \mathsf{v} \quad \text{with} \quad \mathsf{x}\in\Omega, \quad \mathsf{v}\in\Gamma \\
    \mathsf{v} &= -\frac{\mathsf{x}- \mathsf{\hat{x}}_S}{\Vert \mathsf{x} - \mathsf{\hat{x}}_S\Vert} \ \ \text{and}\\
    \mathsf{\hat{x}}_S &= \argmin _{\mathsf{x}_S\in \Pi } \Vert \mathsf{x}-\mathsf{x}_S\Vert \quad \text{if} \quad \mathsf{x} \notin \Pi,\  \text{otherwise:} \\
    \mathsf{v} &= -\frac{\mathsf{x}'- \mathsf{\hat{x}}_S}{\Vert \mathsf{x}'- \mathsf{\hat{x}}_S\Vert} \quad \text{with} \\ \mathsf{x}' &= \lim_{\Vert \mathsf{\epsilon} \Vert \to 0}{\mathsf{x} + \mathsf{\epsilon}}; \ \mathsf{x}' \notin \Pi \ \text{and} \ \epsilon > 0.
\end{split}
\label{eq:vectransform}
\end{align}
Here, $\Vert.\Vert$ is the $\ell_2$ norm operator. Eq.~\eqref{eq:vectransform} defines the representation that maps each point in the subset $\Omega$ of space to a normalized direction vector $\mathsf{v} \in \Gamma$. $\mathsf{\epsilon} \in \mathbb{R}^3$ is an infinitesimal displacement. We note that when multiple points $\mathsf{\hat{x}}_S$ satisfy the argmin condition (for example at equal distance between surfaces), one of them can be chosen arbitrarily. Similarly, the point $\mathsf{x}'$ is chosen as a point with higher coordinate value than $\hat{\mathsf{x}}_S$, with the only constraint that $\mathsf{x}' \notin \Pi$. This particular choice is also an arbitrary one.

We now need to ensure that VF can properly represent surfaces embedded in $\mathbb{R}^3$. Specifically, with the three following properties, we establish a one-to-one map between surface points and the $-1$ level set of flux density computed on VF. Here and throughout the paper, we define flux density as the flux through a spherical surface divided by its cross-sectional area when the radius of such sphere tends to zero.

\begin{property}
The vector field $\mathsf{v} = f(\mathsf{x})$ is equal to the negative gradient of the unsigned distance field, \ie,  $-\nabla u(\mathsf{x})$, except at the discontinuities at the surface points $\{\mathsf{x}_S\}$ and at points at equal distance from multiple surface points.
\end{property}

\begin{property}
The vector field $\mathsf{v} = f(\mathsf{x})$ is equal to the surface normal as it approaches the surface.
\end{property}

\begin{property}
Consider the VF representation $\mathsf{v} = f(\mathsf{x})$ of a piecewise smooth surface as defined in Eq.~\eqref{eq:vectransform}, and the following transform:
\begin{align}
\label{eq:divvt}
\begin{split}
    & g(\mathsf{x}) = D_\Phi f(\mathsf{x}). \\
\end{split}
\end{align}
Here, $D_\Phi$ is the operator for flux density, defined as flux per unit cross-sectional area \cite{maxwell1873treatise}, measured using an infinitesimal spherical surface.
A point $\mathsf{x} \in \mathbb{R}^3$ then is a surface point, $\mathsf{x} \in \Pi$, if and only if it belongs to the zero level set of $g(\mathsf{x})+1$, \ie, 
\begin{equation}
\label{eq:invtransform}
\Pi = L_0(g+1).    
\end{equation}
\end{property}
The properties listed above hold only for piecewise smooth surfaces \cite{osher2003level}. However, in practice, the flux density goes slightly lower than $-1$ at discontinuous surface points. Nonetheless, a simple analysis of the field reveals that elsewhere, the flux density is either around 0 or positive, making the thresholding operation for surface inference simple and robust. Following these properties, we show how VF can be used for INR to represent any type of surface.

\subsubsection{Field Creation and Training}
VF can be used on top of any INR architecture without significant changes. We follow standard practice~\cite{park2019deepsdf,zheng2021deep} and use the auto-decoder architecture for training the INR with VF. The field prediction $\hat{\mathsf{v}}\in\Gamma$ at each point $\mathsf{x}$ is conditioned on a high dimensional embedding vector $\mathsf{c}\in \mathbb{R}^M$. Mathematically, we parameterize the neural VF prediction, with the network parameters $\theta$, as follows:
\begin{equation}
\label{eq:arch}
    f(\mathsf{x};\theta,\mathsf{c}) = \hat{\mathsf{v}}.
\end{equation}
At training time, the embedding vector $\mathsf{c}$ is optimized together with the network parameters $\theta$ to a shape class; at test time, only the embedding vector is optimized on a small set of observed points. This allows one to test the reconstruction accuracy, while also considering the generalization capabilities. Our INR architecture only differs from \cite{park2019deepsdf} in that the output dimension is now 3 instead of 1.
Strictly speaking, $D_\Phi f(\mathsf{x})+1$ is the scalar-valued implicit function that defines the surface. However, a discrete flux density measure can be easily performed on $f$ at the post-processing step, and therefore we use the term INR more broadly to discuss the original function $f$ itself.

For training and optimization, the $\ell_1$ loss\footnote{$\ell_2$ or cosine loss may also be used with a similar outcome.} is computed between the predicted $\mathsf{\hat{v}}$ and ground truth vector $\mathsf{v^{gt}}$:
\begin{equation}
    \label{eq:vt_loss}
    \ell_{VF} = \Vert\mathsf{v^{gt}} - \mathsf{\hat{v}}\Vert_1.
\end{equation}
More specifically, for each object, the loss is computed on a set of coordinates randomly sampled in space with a higher density around the surface~\cite{park2019deepsdf}.
For inference, surfaces can be defined as points in space with flux density lower than a fixed threshold $\alpha$. Given the theoretical value of $-1$ for surface points, we set $\alpha = -0.7$ to account for small prediction inconsistencies. The flux density computation allows us to adapt the marching cubes (MC) algorithm to run on a VF voxel grid. With this, similarly to an independent work \cite{guillard2021meshudf} which uses the gradient of UDF, it is then possible to obtain accurate meshes. The $\alpha$ threshold on the flux density identifies the voxels containing a surface and the VF values are used to assign on which side of the surface the vertices lie. For more details about our implementation, refer to the Supplementary Material.

\begin{figure}[ht]
\centering
\begin{subfigure}{0.14\linewidth}
    \includegraphics[width=\textwidth]{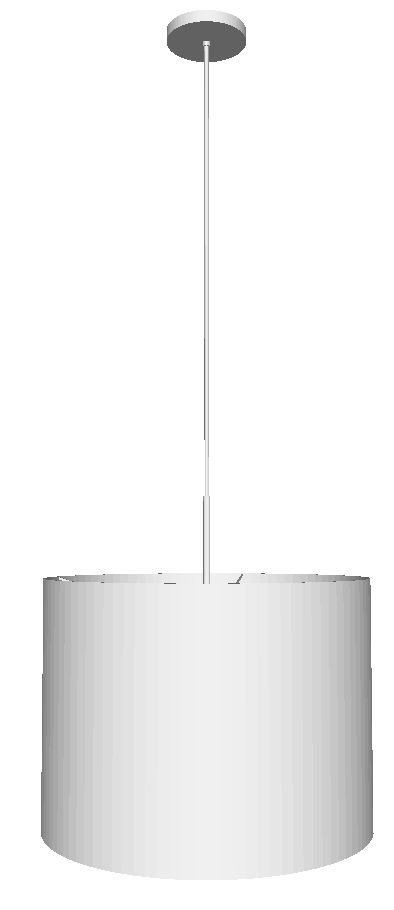}
    \caption{GT}
\end{subfigure}
\hspace{0.08\linewidth}
\begin{subfigure}{0.14\linewidth}
    \includegraphics[width=\textwidth]{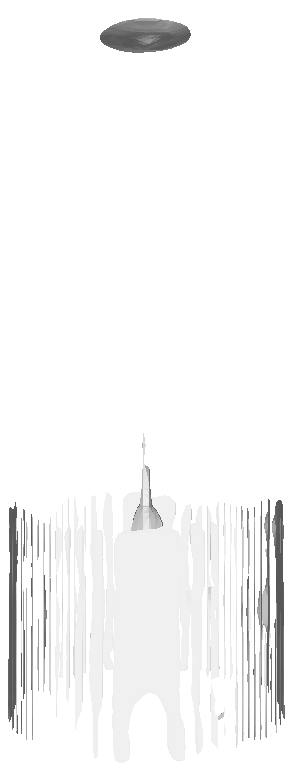}
    \caption{SDF}
\end{subfigure}
\hspace{0.08\linewidth}
\begin{subfigure}{0.14\linewidth}
    \includegraphics[width=\textwidth]{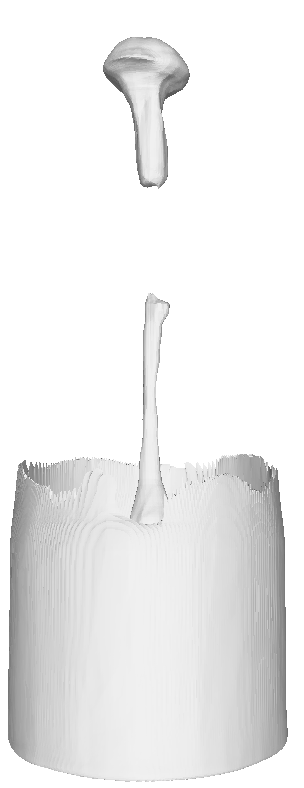}
    \caption{Occ}
\end{subfigure}
\hspace{0.08\linewidth}
\begin{subfigure}{0.14\linewidth}
    \includegraphics[width=\textwidth]{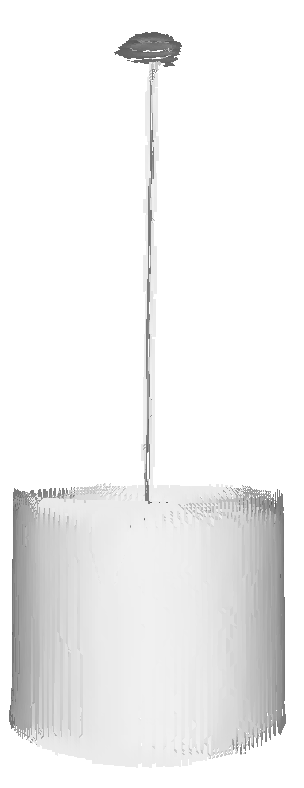}
    \caption{VF}
\end{subfigure} \\
\caption{\textbf{Representation comparison}: VF is compared to SDF and binary occupancy on thin surface reconstruction. Thanks to its ability to represent open surfaces, VF accurately preserves complete thin surfaces.}
\label{fig:thin_surf}
\end{figure}

\subsection{Piecewise Planar Prior on VF}

Applying surface priors on an INR requires discovery and regularization of the surface normals, which are conventionally obtained in SDF and UDF through differentiation. In contrast, VF representation allows one to enforce priors on surface normals directly on the prediction.
We show this advantage of VF by using it to represent piecewise planar objects.
Piecewise planar surface representation or reconstruction is an application well explored in the literature \cite{gallup2010piecewise, romanoni2019tapa, xu2020planar,deng2020hybrid_cvxnet} as large low-textured planar parts are common in man-made environment. The Manhattan-world assumption \cite{coughlan1999manhattan} was recently used in INR\cite{guo2022neural} to obtain multi-view 3D reconstruction of Manhattan worlds. We show that, using VF, it is straightforward to extend the Manhattan assumption to arbitrary directions and numbers of planes which can be specific to the object in a category.

Due to the explicit normals in the VF implicit representation, it is possible to jointly predict a set of bases and select between them at each point in space to obtain the final VF. The direction bases represent a very strong prior on the dominant directions that are present in the scene. The new architecture is described by the following:
\begin{align}
\label{eq:vfplanar}
    \begin{split}
        &f(\mathsf{x}; \theta, \mathsf{c}, \mathsf{b}) = \mathsf{s}_b \\
        &h(\mathsf{c}_b; \theta_h) = \mathsf{b} \\
        &\hat{\mathsf{v}} = \mathsf{s}_b^\top \mathsf{b}.
    \end{split}
\end{align}
We use a similar auto-decoder $f$ with the latent embedding $\mathsf{c}$, and an additional small network branch $h$\footnote{3 hidden layers with $128$ hidden size.}; its input is a different latent vector $\mathsf{c}_b$, and the output is the set of bases $b\in\mathbb{R}^{3\times k}$, where $k$ is the number of bases. We fix $k=10$ for our implementation. The main branch $f$ takes as input the latent vector $\mathsf{c}$ and the predicted bases $\mathsf{b}$, together with the query position $\mathsf{x}$, and selects which of the bases to use at that queried position $\mathsf{x}$. The selection is done using the binary one-hot vector output $\mathsf{s}_b$. In this way, it is possible to learn shape-specific bases, which are then selected to form the VF prediction.

The main branch of the network is trained as a classification network - with the number of classes equal to $k$ - to predict the best matching basis $\mathsf{b}_i$ using the Cross Entropy loss. Basis prediction, instead, is trained with an $\ell_1$ loss applied to the basis that best matches the ground truth VF. The final loss is given by:
\begin{align}
    \begin{split}
        \label{eq:planarloss}
        \ell_{pl} = &\left(- \log \frac{\exp(\mathsf{s}_{\mathsf{b}_i})}{\sum_{n=1}^{k}exp(\mathsf{s}_{\mathsf{b}_n})}\right) \\
        &+ w_{\mathsf{b}} \Vert\mathsf{b}_i - \mathsf{v}^{\mathsf{gt}}\Vert_1 \\
        &i = \argmin_{n \in [1, k]} \Vert\mathsf{b}_n - \mathsf{v}^{\mathsf{gt}}\Vert_1
    \end{split}
\end{align}
Here, we fix the scalar hyperparameter $w_\mathsf{b} = 0.1$.

\begin{figure}[ht]
\centering
\begin{subfigure}{0.22\linewidth}
    \includegraphics[width=\textwidth]{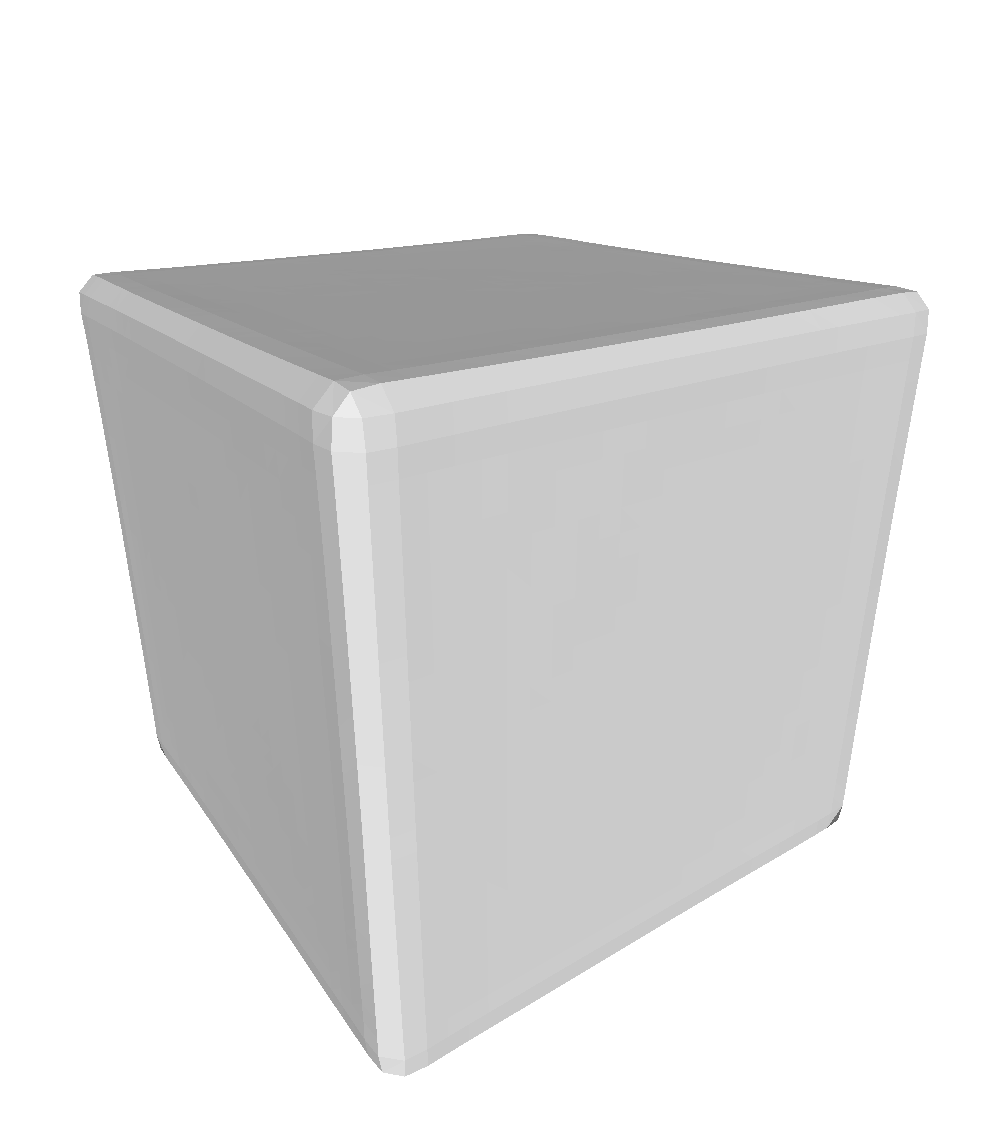}
    \caption{UDF}
    \label{fig:udf_cube}
\end{subfigure}
\hspace{0.01\linewidth}
\begin{subfigure}{0.22\linewidth}
    \includegraphics[width=\textwidth]{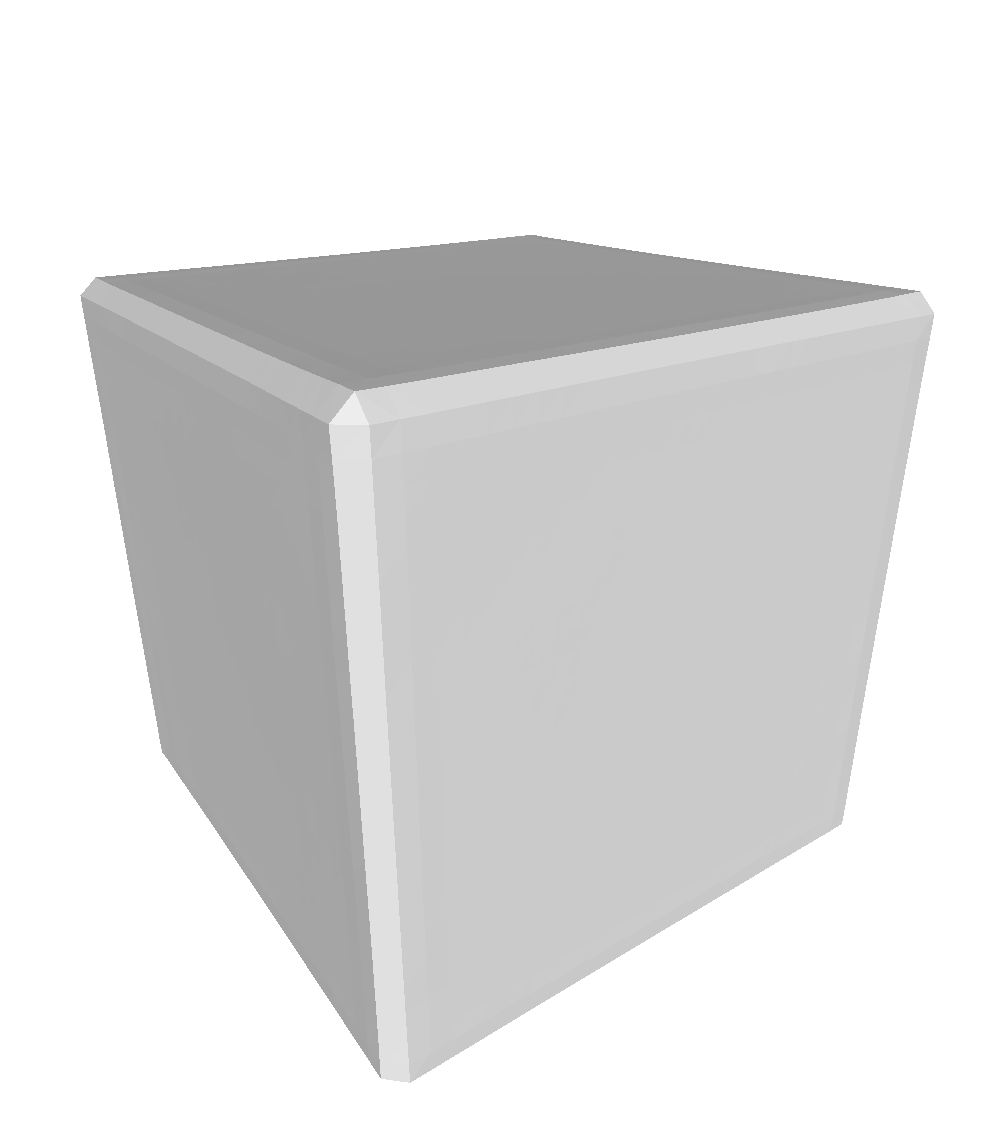}
    \caption{SDF}
\end{subfigure}
\hspace{0.01\linewidth}
\begin{subfigure}{0.22\linewidth}
    \includegraphics[width=\textwidth]{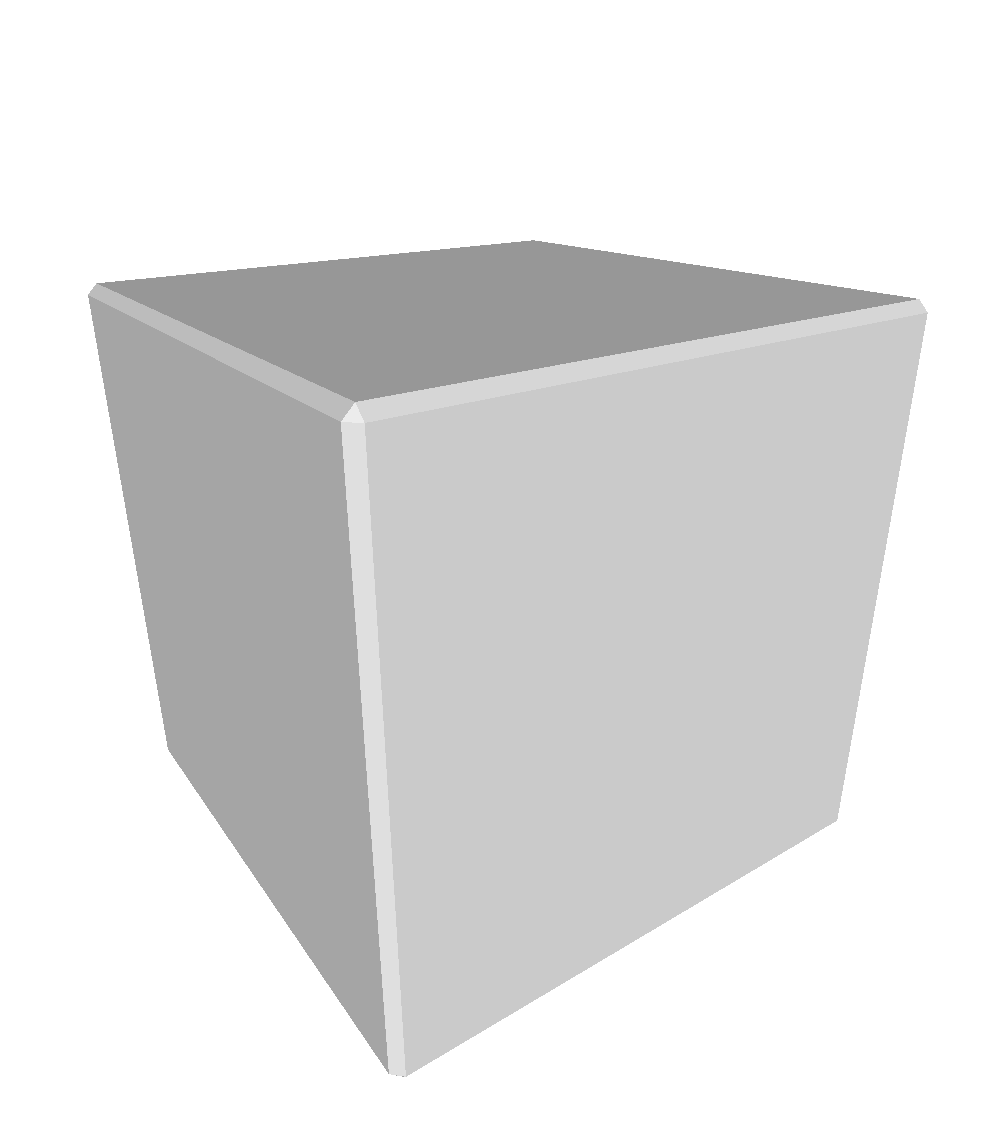}
    \caption{Occ}
\end{subfigure}
\hspace{0.01\linewidth}
\begin{subfigure}{0.22\linewidth}
    \includegraphics[width=\textwidth]{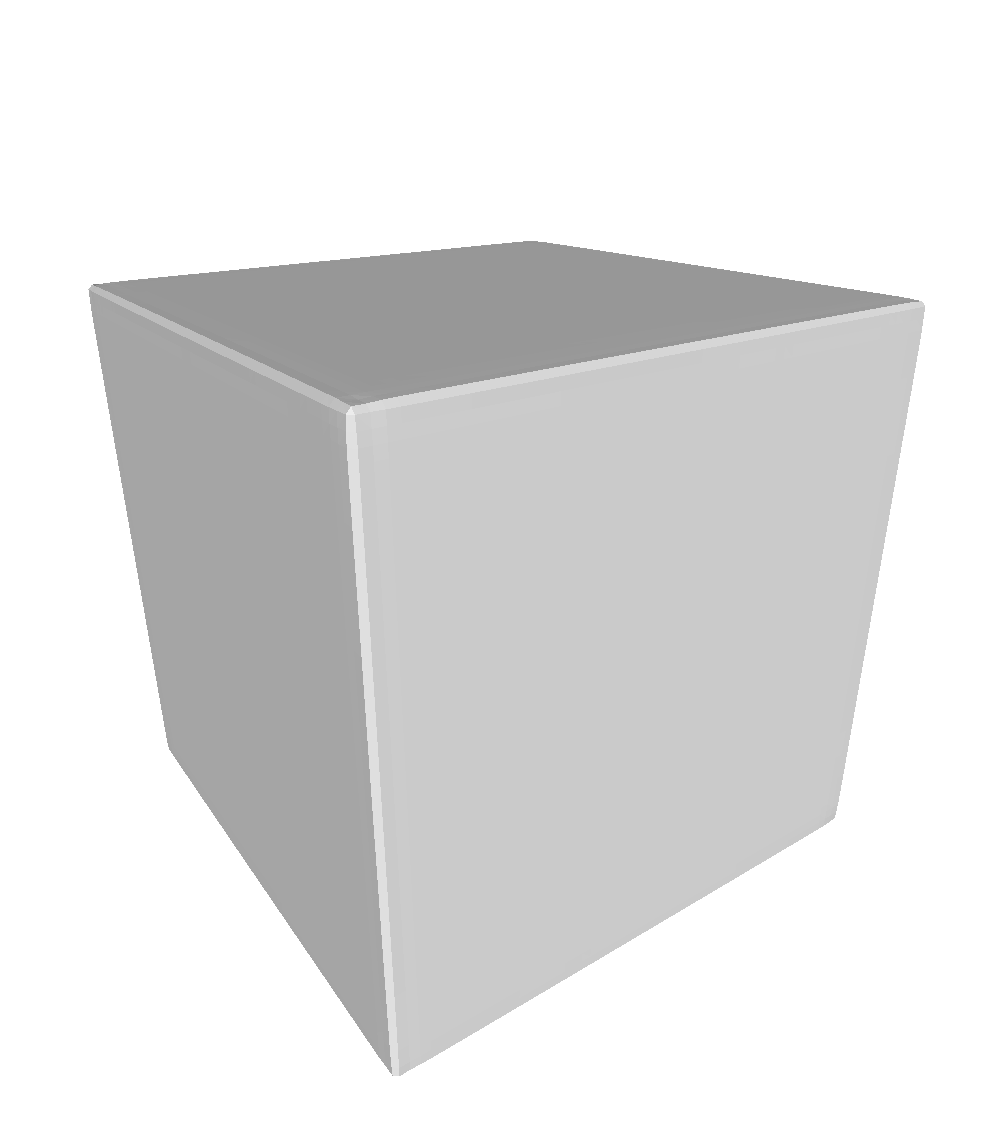}
    \caption{VF}
\end{subfigure} \\
\begin{subfigure}{0.9\linewidth}
    \includegraphics[width=\textwidth]{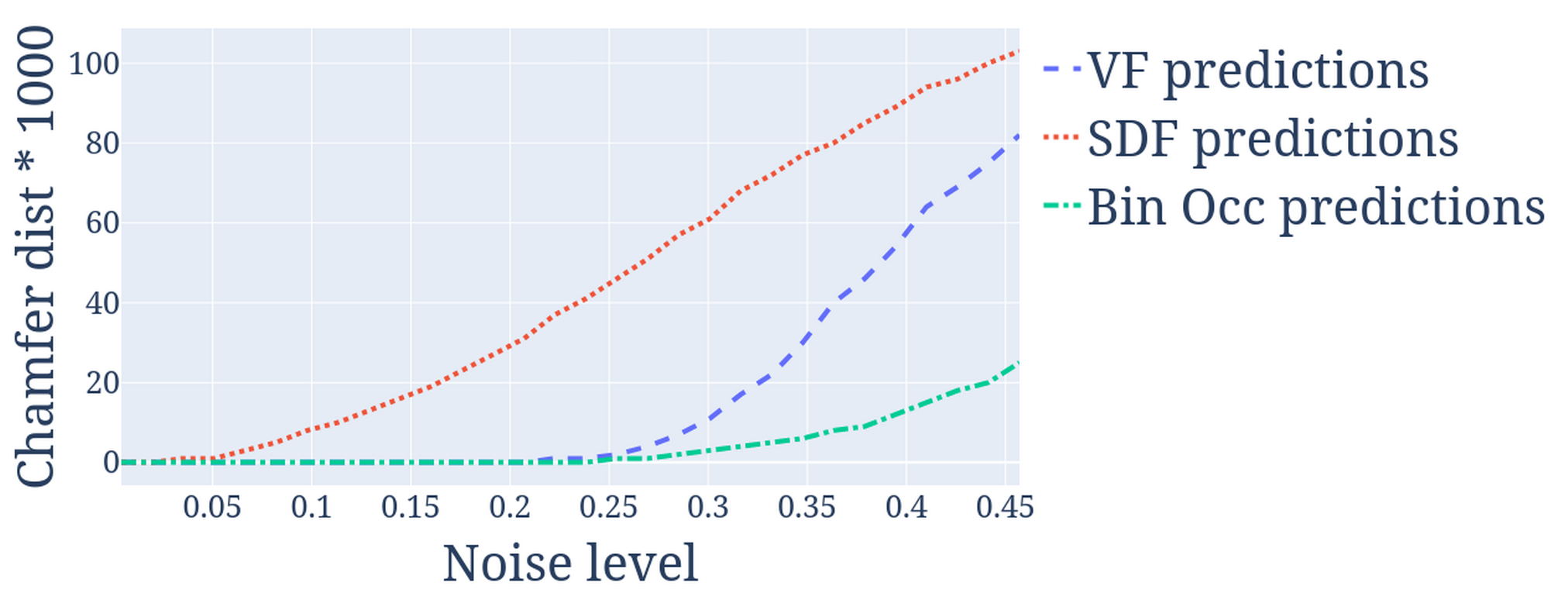}
    \caption{Reconstruction accuracy under varying noise levels}
    \label{fig:noise_effect}
\end{subfigure} \\
\caption{\textbf{Cube reconstruction}: Top: VF is compared to UDF, SDF and binary occupancy on reconstructing a cube. Note the sharper edges on the cube with VF. Bottom: we show the effect of noise added to predicted VF, SDF, and binary occupancy on the final shape after MC. The toy-experiment mimics the effect of prediction errors showing how the final shape may be affected.}
\label{fig:cube_exp}
\end{figure}

\subsection{Discussion on Implicit Representations}
\label{sec:discussion}
Unlike SDF and UDF~\cite{Osher1988-fronts}, the VF as an implicit function is not discussed in the literature. Here we provide more insights into the main differences with existing representations.

\paragraph{VF, SDF and binary occupancy.}
The properties in Section~\ref{sec:prop} show that VF can represent any piecewise smooth surface, including open and non-orientable surfaces, proving its expressiveness.
Besides the stronger representation power, we show in Section \ref{sec:exp} that VF is able to outperform SDF \cite{park2019deepsdf} and binary occupancy \cite{mescheder2019occnet} also on watertight meshes. The difference can still be partially attributed to the more general representation power. A very thin object, even though represented as watertight, can be better modeled as a thin-shelled surface. When modeled as watertight, instead, small prediction errors may lead to holes in the surface. We show this effect on the inference of a lamp from ShapeNet \cite{cheng2015shapenet} in Figure \ref{fig:thin_surf}. Only VF can fully preserve the lampshade and the wire structure.

Additionally, the different performance can be attributed to the effect of extracting the surfaces by thresholding the flux density of VF rather than the predicted level in itself as in SDF and binary occupancy. We conduct a small experiment with a very simple shape shown in Figure \ref{fig:cube_exp}. We use a small 4-layered network to represent a cube. As it is visible from the edges, VF can best preserve the sharpness of the shape. Furthermore, in Figure \ref{fig:noise_effect} we show the effect of artificially adding Gaussian noise on the field predictions. We plot the reconstruction error as rescaled Chamfer distance while increasing the noise standard deviation. As it can be seen, VF and binary occupancy are less sensitive to noise, which makes them less affected by small errors in the prediction. The worse performance of VF compared to binary occupancy at high noise levels is expected, and is due to the addition of noise on three components, which makes the likelihood of flipping at least one direction far higher; at lower and more realistic levels, no discernible difference can be noted. However, it is not clear how the lack of surface context in binary occupancy affects the overall learning process in comparison to others. At each point, away from the surface, the field value contains information regarding the surface distance (in SDF) and direction (in VF), no such information are encoded in binary occupancy.

\begin{table*}[t]
    \centering
    \resizebox{0.95\linewidth}{!}{
    \begin{tabular}{ c | c c | c c | c c | c c | c c }\toprule
         \multirow{3}{*}{\textbf{Method}} & \multicolumn{2}{c|}{\textit{chairs}} & \multicolumn{2}{c|}{\textit{lamps}} & \multicolumn{2}{c|}{\textit{planes}} & \multicolumn{2}{c|}{\textit{sofas}} & \multicolumn{2}{c}{\textit{cars}} \\
         & Chamfer & F1-Score & Chamfer & F1-Score & Chamfer & F1-Score & Chamfer & F1-Score & Chamfer & F1-Score \\
         & mean/median & 0.01 & mean/median & 0.01 & mean/median & 0.01 & mean/median & 0.01 & mean/median & 0.01 \\ \midrule
         OccNet \cite{mescheder2019occnet} & 0.420/0.173 & 87.10 & 3.002/0.448 & 76.83 & 0.162/0.027 & 87.62 & 0.149/0.093 & 86.00 & 1.619/0.275 & 91.14 \\ 
         DeepSDF \cite{park2019deepsdf} & 0.337/\textbf{0.078} & \textbf{89.31} & 0.798/0.183 & 78.10 & 0.100/0.047 & 87.34 & \textbf{0.120}/\textbf{0.073} & 86.74 & 1.417/0.236 & 91.71 \\
         NDF \cite{chibane2020ndf} & 0.418/0.275 & 82.47 & 0.640/0.235 & 75.34 & 0.177/0.085 & 86.49 & 0.286/0.160 & 84.21 & 0.430/0.357 & 86.23 \\
         MC NDF & 0.629/0.347 & 78.59 & 0.844/0.278 & 70.05 & 0.347/0.161 & 84.90 & 0.575/0.206 & 81.75 & 1.913/0.572 & 78.40 \\
         PRIF \cite{Feng2022} & 0.982/0.267 & - & 3.276/0.534 & - & 0.389/0.125 & - & 1.236/0.222 & - & 1.961/0.347 & - \\
         GIFS \cite{Ye_2022gifs} & 0.612/0.252 & 83.32 & 0.749/0.296 & 68.84 & 0.295/0.041 & 88.84 & 0.302/0.147 & 87.87 & 0.573/0.392 & 82.66 \\ \midrule
         \textbf{VF} & \textbf{0.193}/0.092 & 87.57 & \textbf{0.299}/\textbf{0.163} & \textbf{79.26} & \textbf{0.074}/\textbf{0.024} & \textbf{90.11} & 0.152/0.076 & \textbf{88.43} & \textbf{0.374}/\textbf{0.230} & \textbf{93.97} \\ \bottomrule
    \end{tabular}}
    \caption{\emph{Evaluation on watertight surfaces.} Reconstruction results on watertight ShapeNet~\cite{cheng2015shapenet} categories. The latent vector used to represent the object has size 256 and is optimized for 800 iterations.}
    \label{tab:reconstruction_result}
\end{table*}

\begin{table*}[!t]
    \centering
    \resizebox{0.8\linewidth}{!}{
    \begin{tabular}{ c | c c | c c | c c | c c }\toprule
         \multirow{3}{*}{\textbf{Method}} & \multicolumn{2}{c|}{\textit{cars}} & \multicolumn{2}{c|}{\textit{busses}} & \multicolumn{2}{c|}{\textit{lamps}} & \multicolumn{2}{c}{\textit{clothes}} \\
         & Chamfer & F1-Score & Chamfer & F1-Score & Chamfer & F1-Score & Chamfer & F1-Score \\
         & mean/median & 0.01 & mean/median & 0.01 & mean/median & 0.01 & mean/median & 0.01 \\ \midrule
         NDF \cite{chibane2020ndf} & 0.194/0.166 & 83.82 & 0.178/0.157 & 88.72 & 0.573/0.287 & 67.12 & 0.786/0.733 & 80.50 \\
         MC NDF & 0.483/0.319 & 75.04 & 0.632/0.457 & 84.77 & 0.893/0.309 & 66.96 & 1.165/0.979 & 77.41 \\ 
         GIFS \cite{Ye_2022gifs} & 0.397/0.289 & 76.60 & 0.288/0.125 & 89.43 & 0.729/0.285 & 67.82 & 0.629/0.582 & 81.47 \\ \midrule
         \textbf{VF} & \textbf{0.137}/\textbf{0.106} & \textbf{86.10} & \textbf{0.153}/\textbf{0.091} & \textbf{90.48} & \textbf{0.312}/\textbf{0.144} & \textbf{78.46} & \textbf{0.128}/\textbf{0.112} & \textbf{89.26} \\ \bottomrule
    \end{tabular}}
    \caption{\emph{Evaluation on open surfaces.} Reconstruction results on unprocessed open ShapeNet~\cite{cheng2015shapenet} categories and clothes from MGN dataset \cite{Bhatnagar_2019_ICCV}. The latent vector used to represent the object has size 256 and is optimized for 800 iterations.}
    \label{tab:open_reconstruction_result}
\end{table*}

\paragraph{VF and open surface representations.}
While considering distance based representations, the UDF, used by \cite{chibane2020ndf}, can represent any type of surface, similar to VF. At a first glance, one may expect VF and UDF to perform similarly, after all they are related by Property 3.1. However, as shown by \cite{Ye_2022gifs}, predicting UDF \cite{chibane2020ndf} suffers from slow inference times and is prone to producing results with holes in the reconstructed shapes due to prediction noise, a major weakness in UDF.

There are some major differences when comparing VF and UDF in practice. Inferring a surface with UDF requires computing its gradient \cite{chibane2020ndf}, which makes the process more time-consuming and memory-intensive than SDF/VF.
Furthermore, by evaluating normal consistency at the surface, we show that VF predictions are more accurate then deriving the directions by differentiating UDF. By computing the cosine similarity to the ground truth at the surface across all the categories used in Section \ref{sec:exp}, VF performs over $8\%$ better than the directions obtained by differentiating UDF. Distinct results for all the categories are available in the Supplementary Material. The same is observed on the cube reconstruction in Fig. \ref{fig:cube_exp}, where the $\ell_1$\footnote{Better than cosine similarity for highlighting differences in the simple toy experiment} distance from the ground truth of VF is $0.125$, better than $0.149$ obtained by UDF.
This performance gap results in lower accuracy at the edges in Figure \ref{fig:udf_cube} and makes VF prediction more suitable than UDF to be applied with MC.

Additionally, as UDF is not symmetric around $0$, it suffers from a bias towards the middle of the prediction range between $0$ and the maximum distance threshold. This is a factor which leads to high prediction noise as shown in \cite{rella2022zero} for 2D boundary detection.
We observe the same when representing a cube (Fig. \ref{fig:cube_exp}) with predicted UDF values at the surface that fluctuate between $0$ and $0.048$.
We provide further analysis on this aspect in the Supplementary Material.

Point-pair training for surface detection, proposed by \cite{Ye_2022gifs}, can represent any type of surface as VF. However, it requires a more complex network as it predicts UDF together with the surface detection. Furthermore, for each candidate cube at inference, it requires computing the boundary presence for each 2 vertices couples. The prediction with point couples does not allow to trivially apply prior knowledge on the surface properties, something that is straightforward with VF.
Finally, as shown in the Section \ref{sec:exp}, it achieves quantitatively lower performance compared to competitor methods. A hypothesis for this difference, might be the higher complexity of learning a function that relates point couples, when compared to single point predictions as SDF, binary occupancy, UDF, and VF.

\section{Experiments}
\label{sec:exp}
First, we describe the experimental details for network structure and the training set-up, followed by the tasks and the metrics used. We then provide the results on each task comparing the performance of VF to other representations. Finally, we show more qualitative results and provide a discussion on the performance.

\begin{figure*}[ht]
\centering
\begin{subfigure}{0.13\linewidth}
\includegraphics[width=\textwidth]{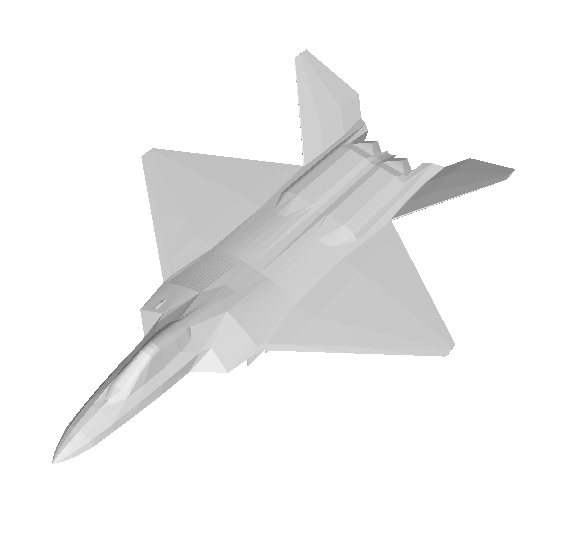}
\end{subfigure}
\hspace{0.02\linewidth}
\begin{subfigure}{0.13\linewidth}
\includegraphics[width=\textwidth]{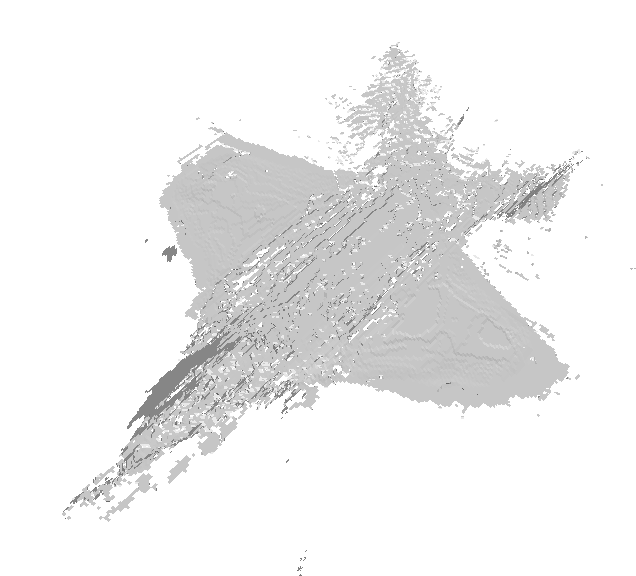}
\end{subfigure}
\hspace{0.02\linewidth}
\begin{subfigure}{0.13\linewidth}
\includegraphics[width=\textwidth]{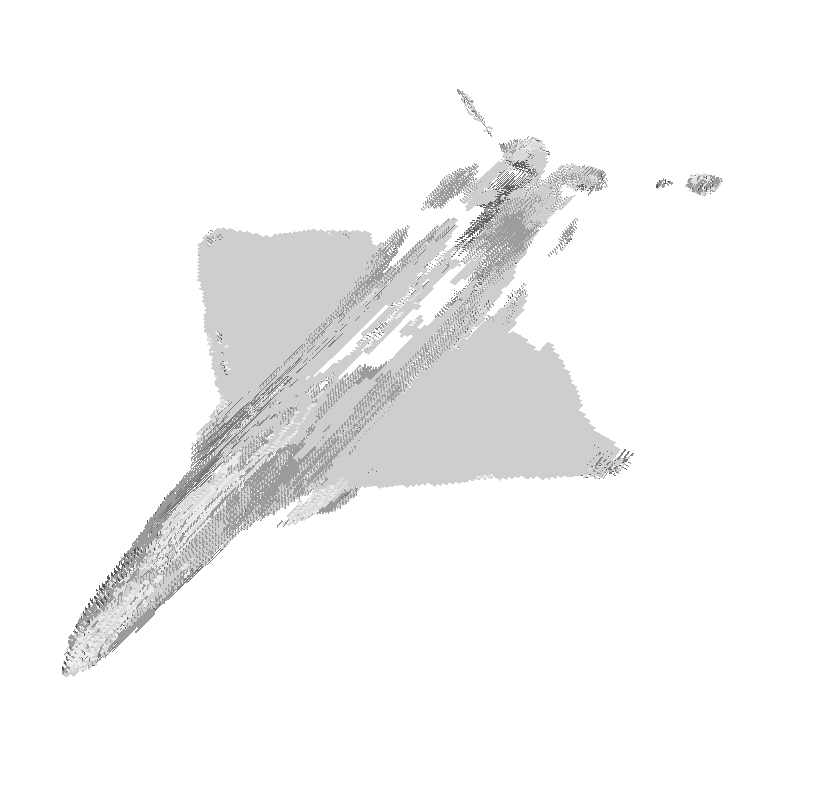} 
\end{subfigure}
\hspace{0.02\linewidth}
\begin{subfigure}{0.13\linewidth}
\includegraphics[width=\textwidth]{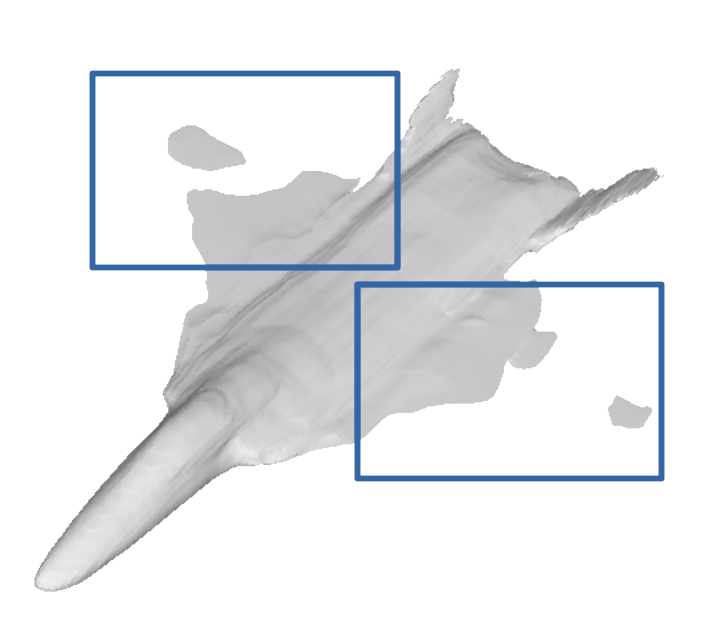} 
\end{subfigure}
\hspace{0.02\linewidth}
\begin{subfigure}{0.13\linewidth}
\includegraphics[width=\textwidth]{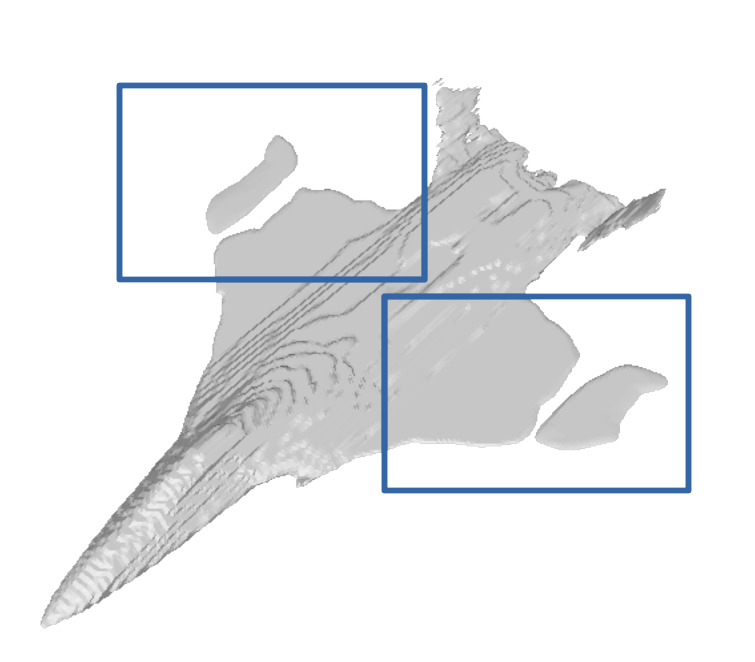} 
\end{subfigure}
\hspace{0.02\linewidth}
\begin{subfigure}{0.13\linewidth}
\includegraphics[width=\textwidth]{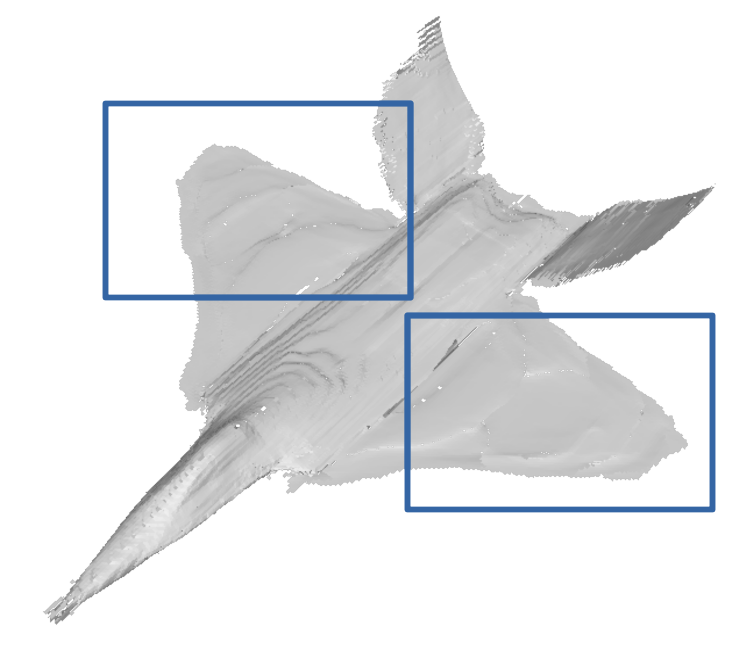} 
\end{subfigure} 
\begin{subfigure}{0.13\linewidth}
\includegraphics[width=\textwidth]{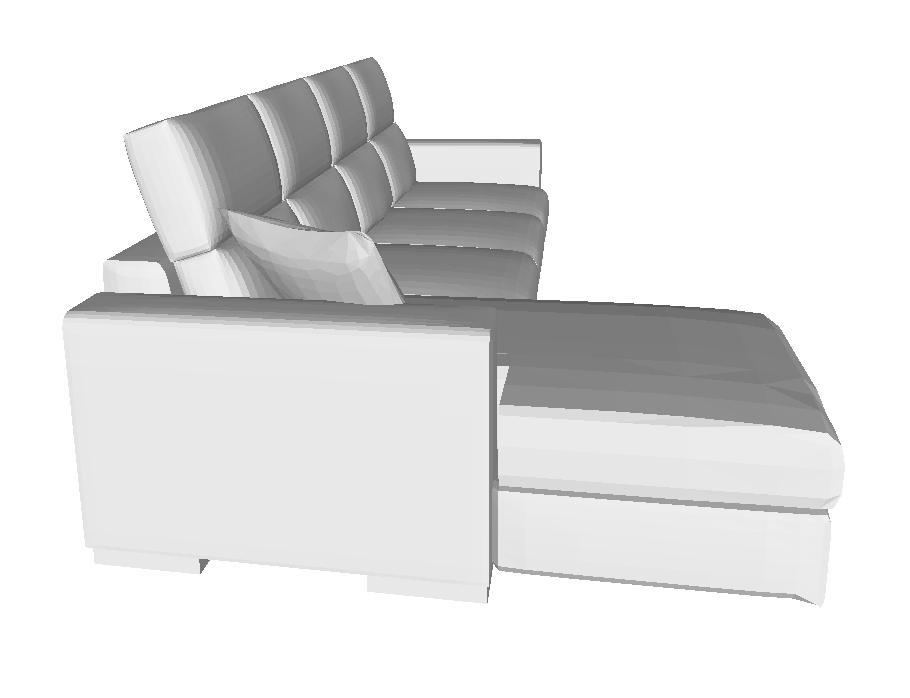}
\end{subfigure}
\hspace{0.02\linewidth}
\begin{subfigure}{0.13\linewidth}
\includegraphics[width=\textwidth]{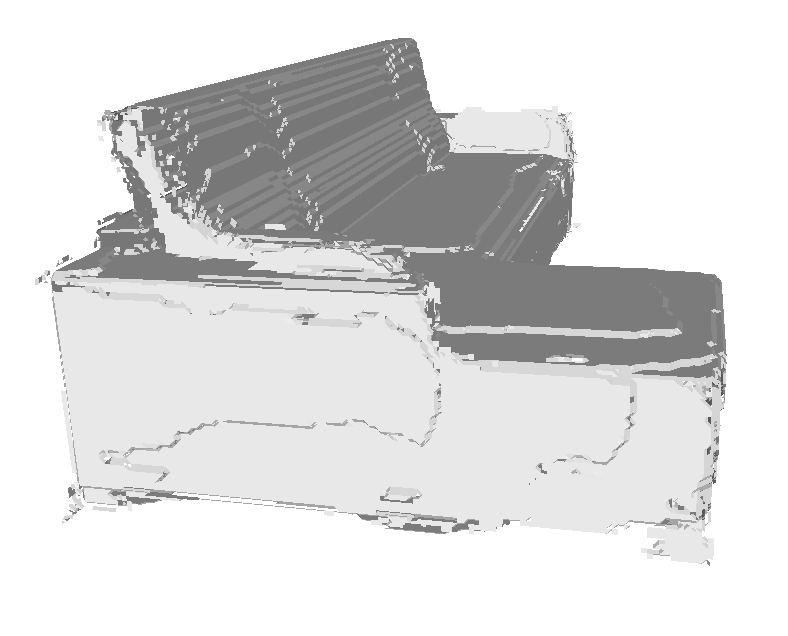}
\end{subfigure}
\hspace{0.02\linewidth}
\begin{subfigure}{0.13\linewidth}
\includegraphics[width=\textwidth]{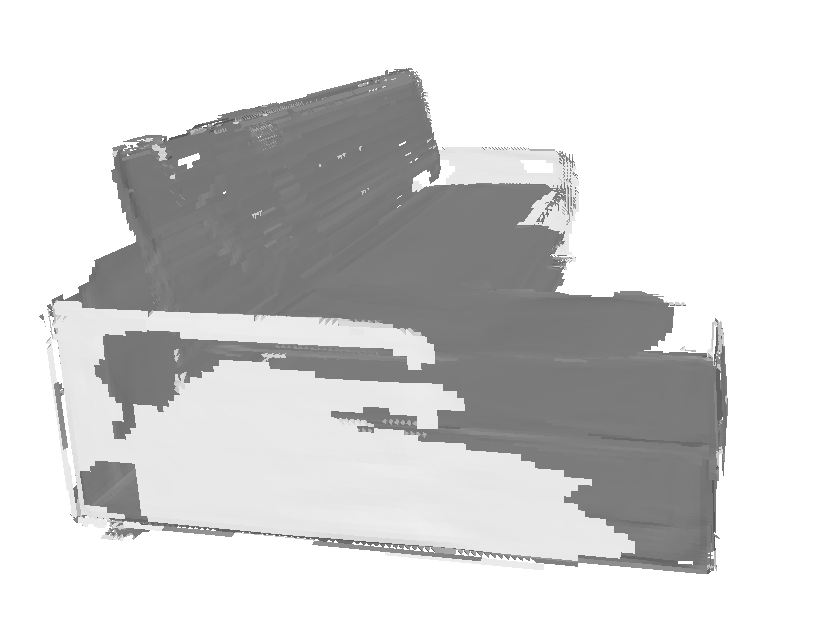} 
\end{subfigure}
\hspace{0.02\linewidth}
\begin{subfigure}{0.13\linewidth}
\includegraphics[width=\textwidth]{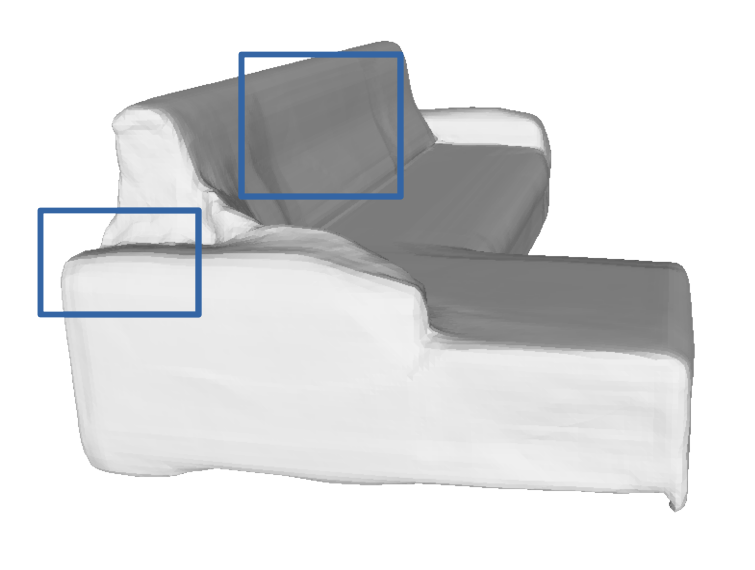} 
\end{subfigure}
\hspace{0.02\linewidth}
\begin{subfigure}{0.13\linewidth}
\includegraphics[width=\textwidth]{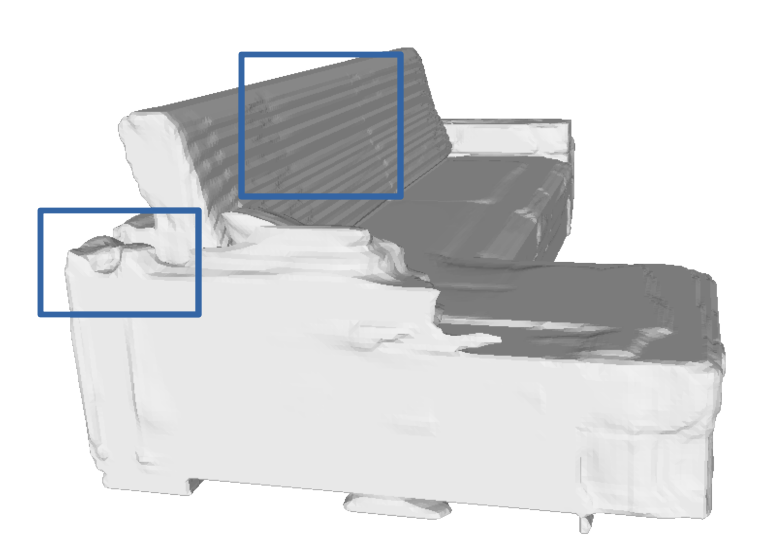} 
\end{subfigure}
\hspace{0.02\linewidth}
\begin{subfigure}{0.13\linewidth}
\includegraphics[width=\textwidth]{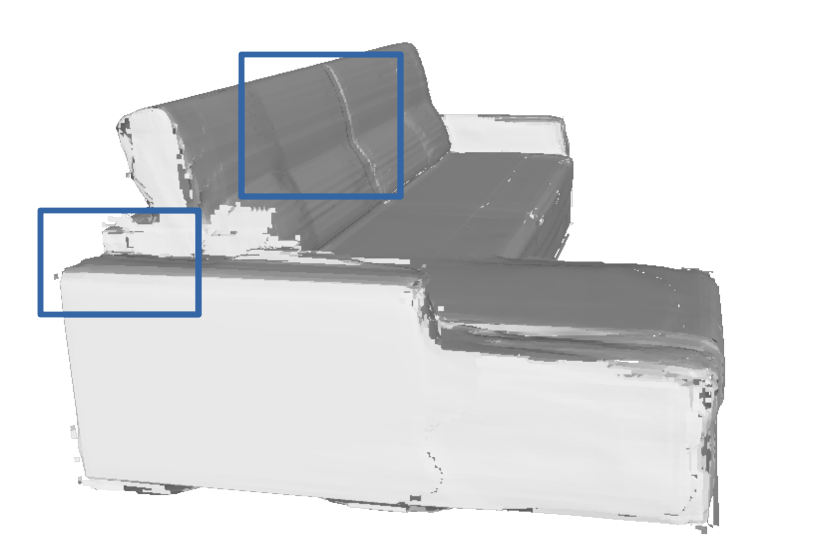} 
\end{subfigure} 
\begin{subfigure}{0.13\linewidth}
\includegraphics[width=\textwidth]{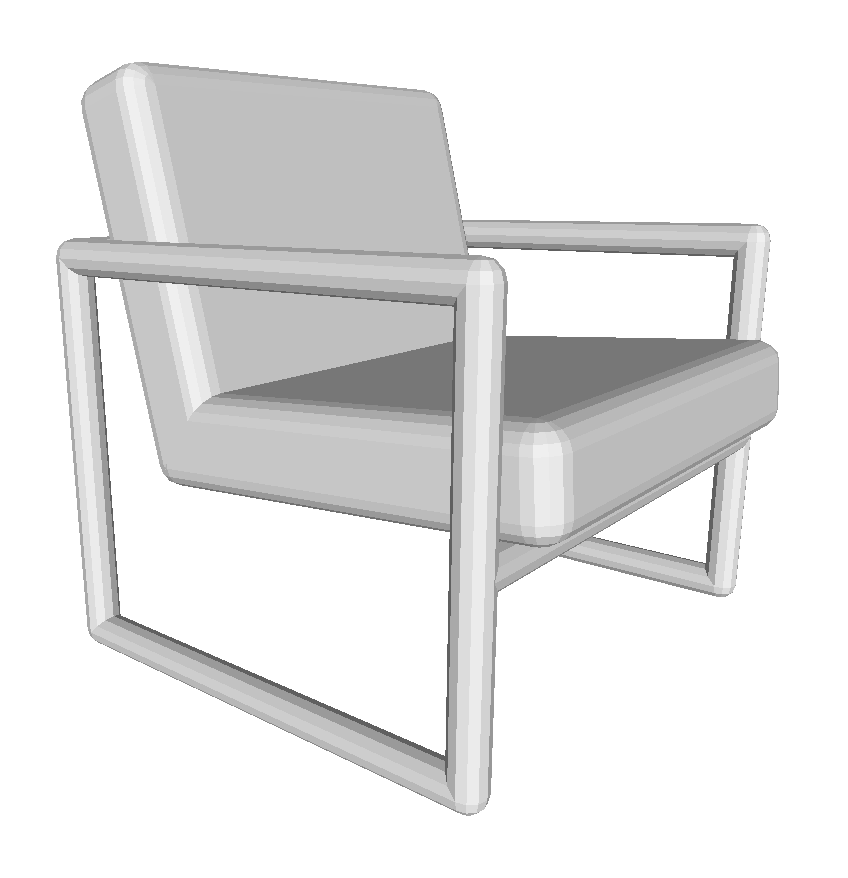}
\caption{GT}
\end{subfigure}
\hspace{0.02\linewidth}
\begin{subfigure}{0.13\linewidth}
\includegraphics[width=\textwidth]{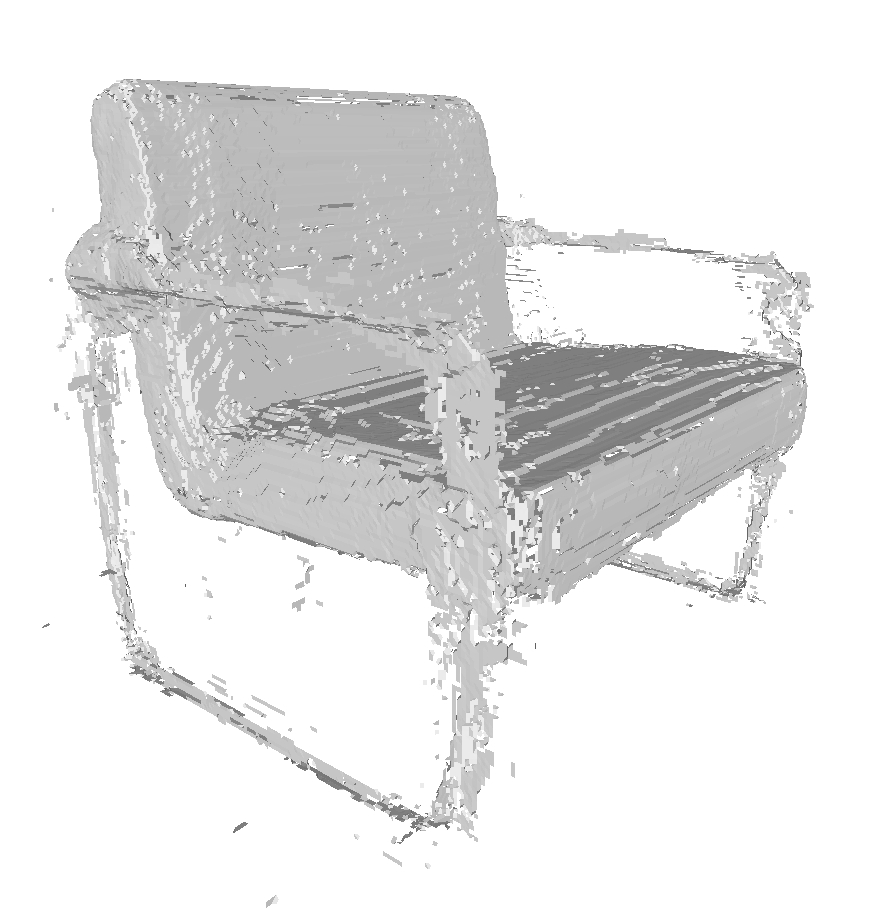}
\caption{NDF \cite{chibane2020ndf}}
\end{subfigure}
\hspace{0.02\linewidth}
\begin{subfigure}{0.13\linewidth}
\includegraphics[width=\textwidth]{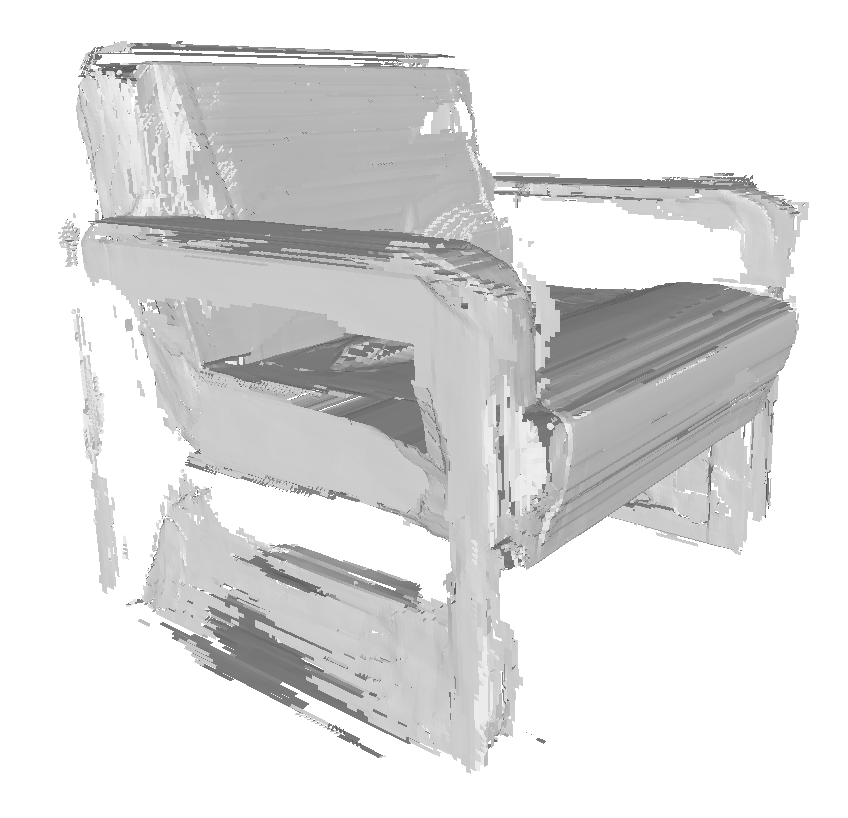} 
\caption{GIFS \cite{Ye_2022gifs}}
\end{subfigure}
\hspace{0.02\linewidth}
\begin{subfigure}{0.13\linewidth}
\includegraphics[width=\textwidth]{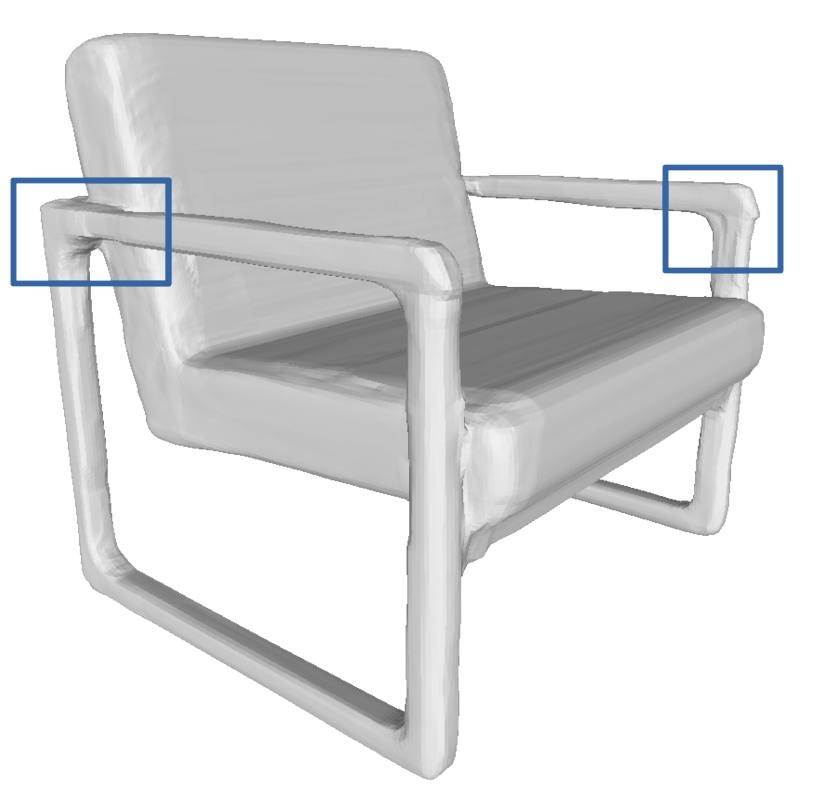} 
\caption{DeepSDF \cite{park2019deepsdf}}
\end{subfigure}
\hspace{0.02\linewidth}
\begin{subfigure}{0.13\linewidth}
\includegraphics[width=\textwidth]{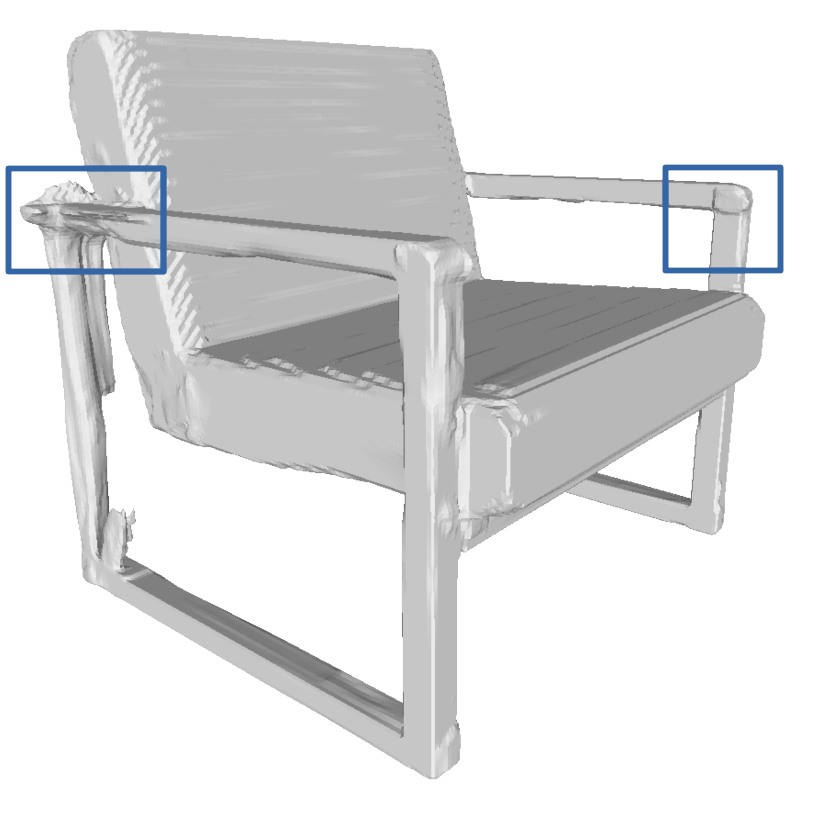} 
\caption{OccNet \cite{mescheder2019occnet}}
\end{subfigure}
\hspace{0.02\linewidth}
\begin{subfigure}{0.13\linewidth}
\includegraphics[width=\textwidth]{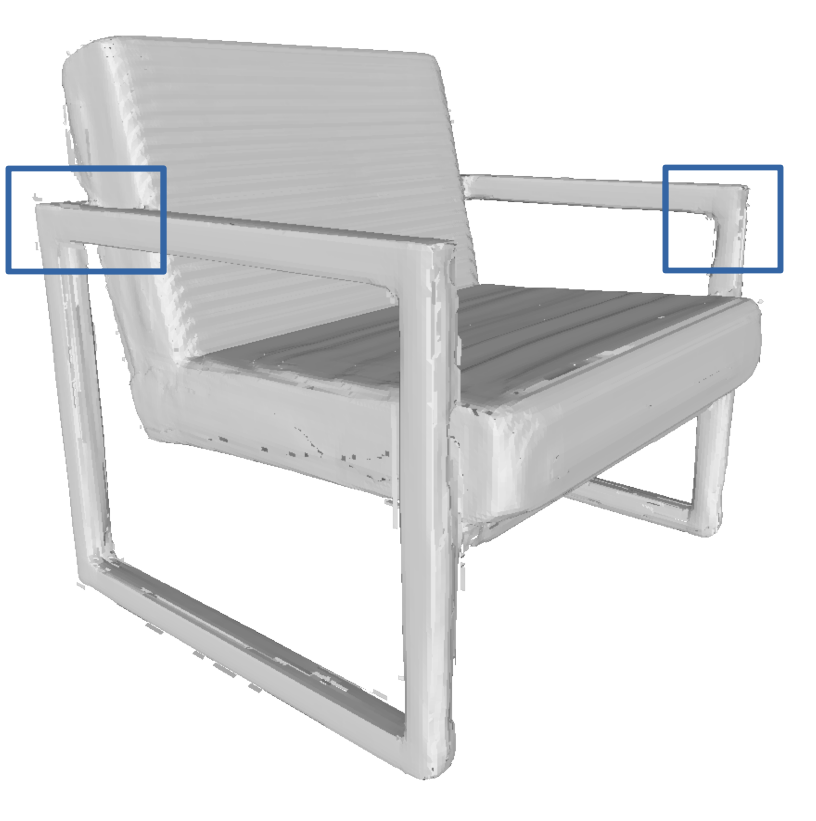}
\caption{VF}
\end{subfigure} 
\caption{\textbf{Watertight shape representation}: VF can represent a wide range of shapes with accuracy and aesthetically pleasing results. In thin surfaces and straight angles, VF achieves the best performance preserving the complete surface and without oversmoothing. See, for example, the separation between cushions of the sofa. We highlight the major differences in blue rectangles.}
\label{fig:qual_water}
\end{figure*}

\subsection{Network Structure and Methods}
In order to fairly evaluate our method against others, we follow standard architecture and training methods. For shape reconstruction, we use the popular architecture proposed in \cite{park2019deepsdf} in all methods, together with its training and sampling protocol, unless a change is required for the compared method. It is a fully connected auto-decoder network that takes as input a latent vector together with the queried position to predict the field. Both the network and the latent code are optimized at training time, while only the latent vector is optimized at test time. The network is trained for $2\,000$ epochs with samples from $64$ scenes in each batch and $16\,384$ points per scene. Each method uses a latent vector of size 256 which is optimized for 800 iterations on 800 points each, at test time. For further details concerning the network structure and the training, please refer to the original work \cite{park2019deepsdf}.

The auto-decoder network structure is used together with the state-of-the-art representations currently available, namely the binary occupancy (OccNet \cite{mescheder2019occnet}), the thresholded signed and unsigned distance field (DeepSDF \cite{park2019deepsdf} and NDF \cite{chibane2020ndf}), the point-pair method of surface detection (GIFS \cite{Ye_2022gifs}), and the proposed VF. Each method is trained with its proposed loss; specifically, OccNet \cite{mescheder2019occnet} uses the binary cross-entropy loss, DeepSDF \cite{park2019deepsdf}, NDF \cite{chibane2020ndf}, and GIFS \cite{Ye_2022gifs} the $\ell_1$ loss, which we use for VF as well. The threshold value for the signed and unsigned distance field is also set based on previous work~\cite{chibane2020ndf,park2019deepsdf}. Additionally, we compare to PRIF~\cite{Feng2022} by taking the results reported on their work. No qualitative comparison was possible for PRIF~\cite{Feng2022} as no code was provided for reproducing the results.

At inference, we first sample the fields on a $256^3$ voxel grid and then apply marching cubes (MC) algorithm on it. OccNet \cite{mescheder2019occnet} and DeepSDF \cite{park2019deepsdf} can be directly used with the standard version of MC \cite{lewiner2003efficient} to provide the resulting mesh. GIFS \cite{Ye_2022gifs} uses an adapted MC algorithm available in their public code. NDF \cite{chibane2020ndf} is evaluated with their proposed evaluation protocol which projects points in space to the surface based on the predicted NDF \cite{chibane2020ndf}; additionally, for fair comparison with the other methods, we also evaluate NDF \cite{chibane2020ndf} with the developed adaptation of MC and call this method MC NDF. The same MC variant is applied to our method, VF, to obtain a resulting mesh.

\subsection{Tasks, Metrics, and Datasets}
We apply the INR methods on three different tasks: watertight shape reconstruction, open shape reconstruction, and piecewise planar objects reconstruction. In each task, the network is first trained on a shape class, and then used to reconstruct a set of unseen objects belonging to the same class.

All reconstructions are evaluated using the symmetric Chamfer distance (CD) on $30\,000$ points with the results written as $CD \times 1\,000$. Specifically, $30\,000$ points are uniformly sampled on the ground truth and the predicted mesh, and the average distances from each point in the set to the closest in the other are summed. To reduce the random effect of points sampling, each result is obtained by averaging the results over three different samplings. Additionally, each method is evaluated also in terms of F1-score \cite{Genova_2020_CVPR}.

We use multiple classes of the popular ShapeNet~\cite{cheng2015shapenet} dataset for evaluation. We measure watertight mesh reconstruction performance on the \textit{chairs}, \textit{lamps}, \textit{planes}, \textit{sofas}, and \textit{cars} classes separately; \textit{chairs}, \textit{lamps}, \textit{planes}, \textit{sofas} are made watertight following the protocol in \cite{park2019deepsdf} and, for \textit{cars}, we use the popular \cite{choy2016voxelsmall1} split. Open shapes are evaluated on the unprocessed mesh from the ShapeNet~\cite{cheng2015shapenet} classes, \textit{lamps}, \textit{cars}, \textit{buses} and the clothes of the MGN dataset \cite{Bhatnagar_2019_ICCV}.
For every task, each class is divided into a training, validation and testing set, with respectively $70\%$, $10\%$, and $20\%$ of the data.

\subsection{Shape Reconstruction}

First, in Table \ref{tab:reconstruction_result}, we evaluate the different representations on watertight shape reconstruction from different ShapeNet \cite{cheng2015shapenet} categories. Second, in Table \ref{tab:open_reconstruction_result}, inference is run on non-watertight meshes from ShapeNet \cite{cheng2015shapenet} and MGN \cite{Bhatnagar_2019_ICCV} dataset.

Table \ref{tab:reconstruction_result} and \ref{tab:open_reconstruction_result} show that VF is overall the best performing method, outperforming any other representation in most cases. The differences can be attributed to the higher capability of VF to preserve sharp details and its robustness as discussed in Section \ref{sec:discussion}. Furthermore, the better performance of VF compared to NDF \cite{chibane2020ndf}, together with the worse results achieved when applying MC to it (MC NDF), suggests that directly predicting directions is more effective than deriving them from a distance function.

\begin{figure}[ht]
\centering
\begin{subfigure}{0.21\linewidth}
\includegraphics[width=\textwidth]{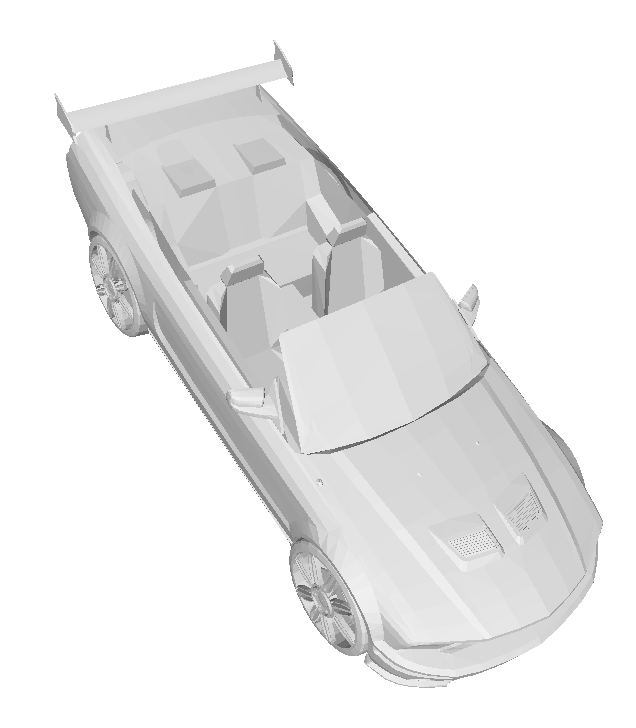}
\end{subfigure}
\hspace{0.02\linewidth}
\begin{subfigure}{0.21\linewidth}
\includegraphics[width=\textwidth]{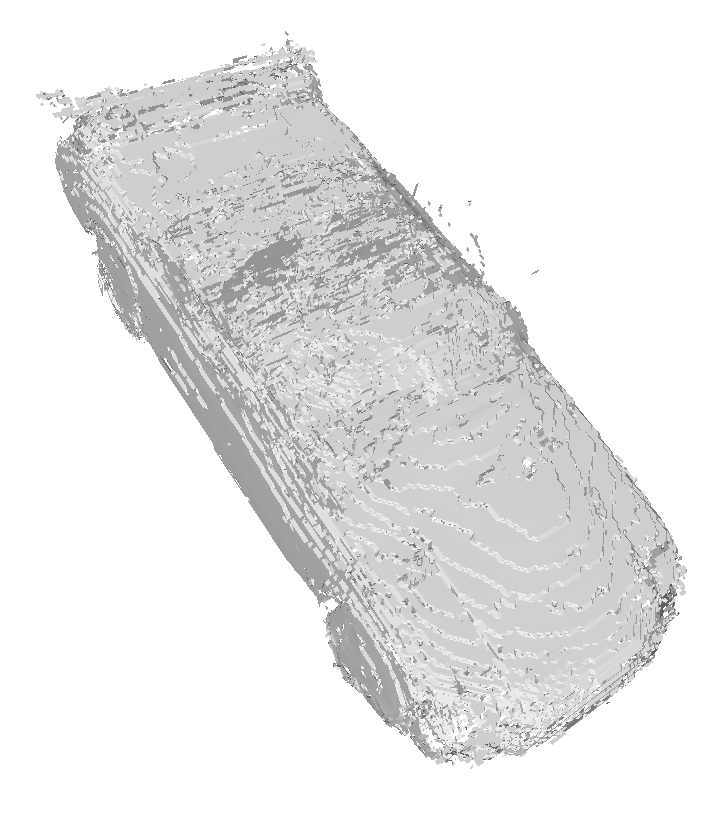}
\end{subfigure}
\hspace{0.02\linewidth}
\begin{subfigure}{0.21\linewidth}
\includegraphics[width=\textwidth]{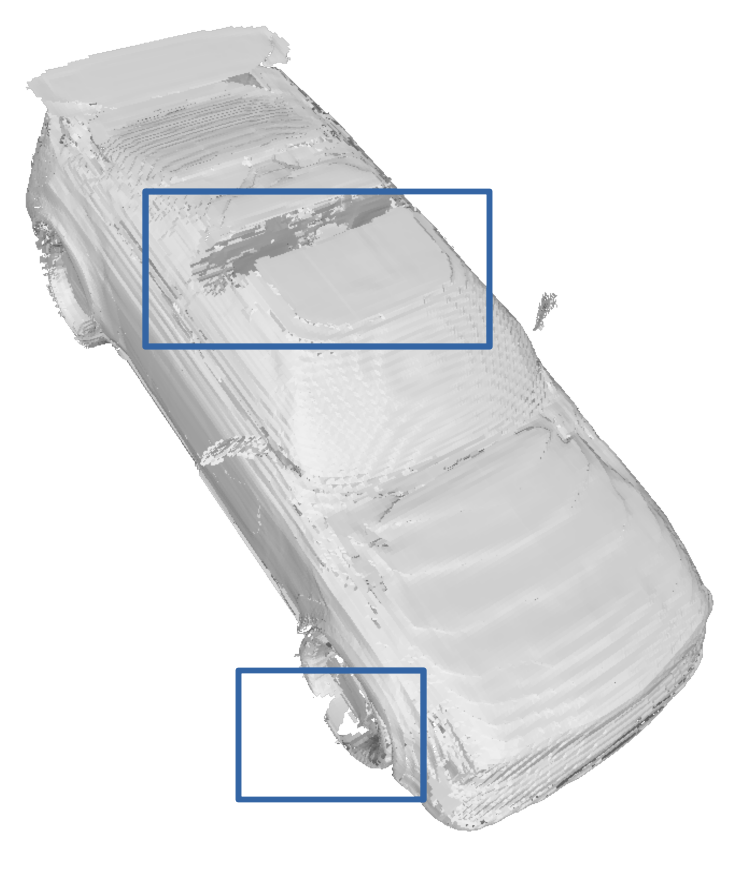} 
\end{subfigure}
\hspace{0.02\linewidth}
\begin{subfigure}{0.21\linewidth}
\includegraphics[width=\textwidth]{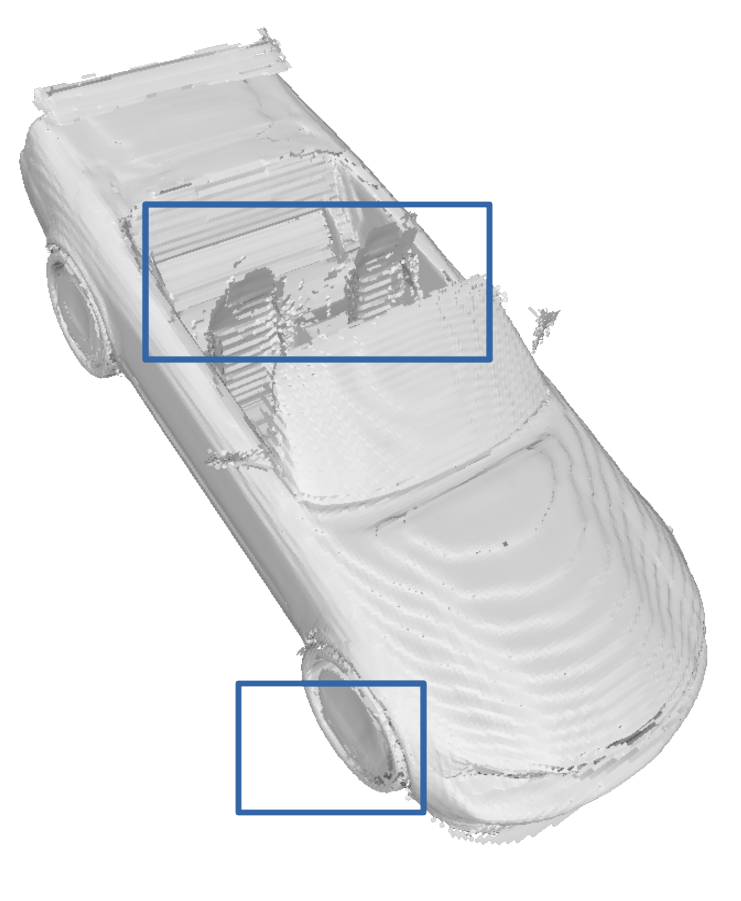} 
\end{subfigure} 
\begin{subfigure}{0.21\linewidth}
\includegraphics[width=\textwidth]{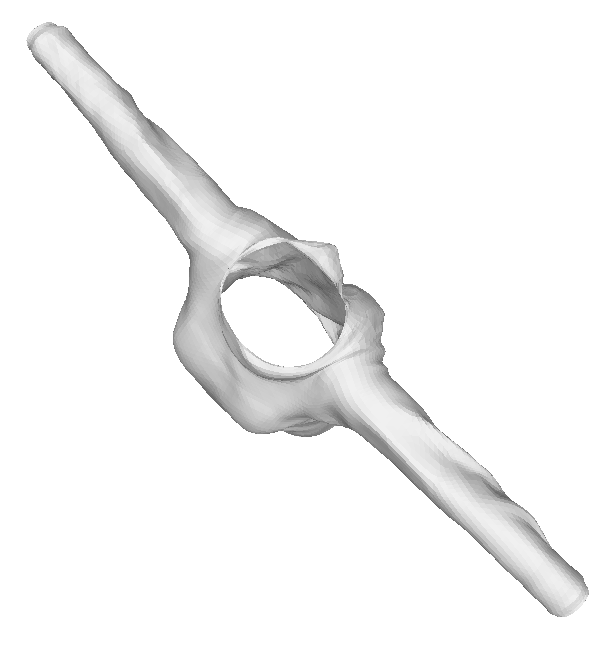}
\end{subfigure}
\hspace{0.02\linewidth}
\begin{subfigure}{0.21\linewidth}
\includegraphics[width=\textwidth]{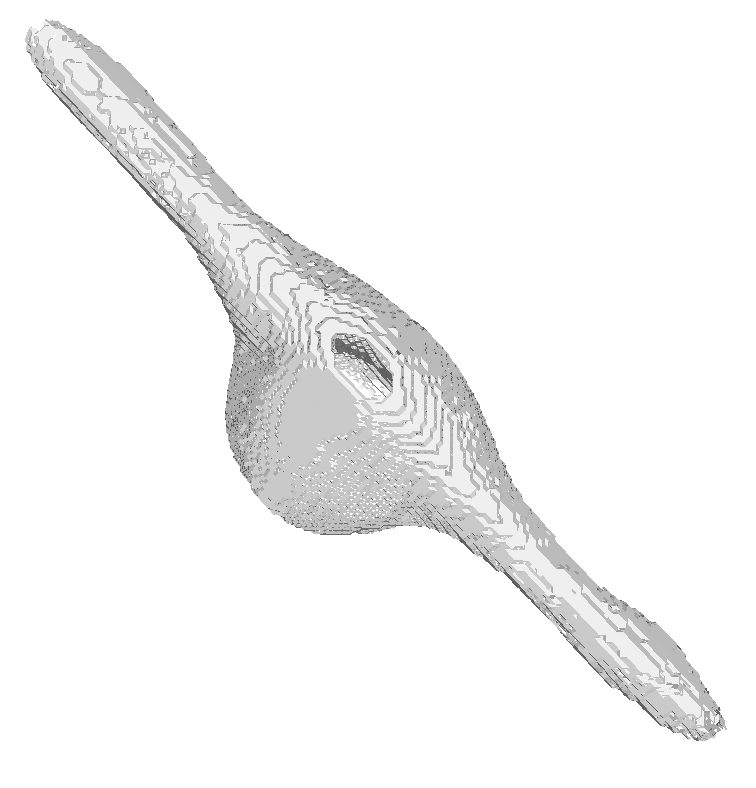}
\end{subfigure}
\hspace{0.02\linewidth}
\begin{subfigure}{0.21\linewidth}
\includegraphics[width=\textwidth]{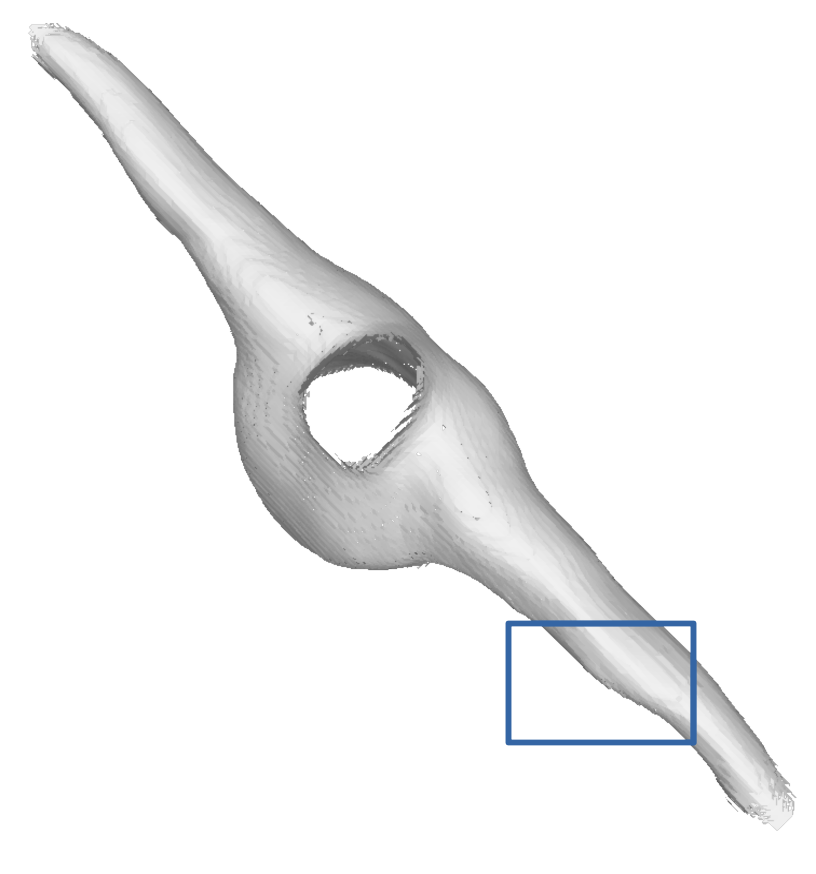} 
\end{subfigure}
\hspace{0.02\linewidth}
\begin{subfigure}{0.21\linewidth}
\includegraphics[width=\textwidth]{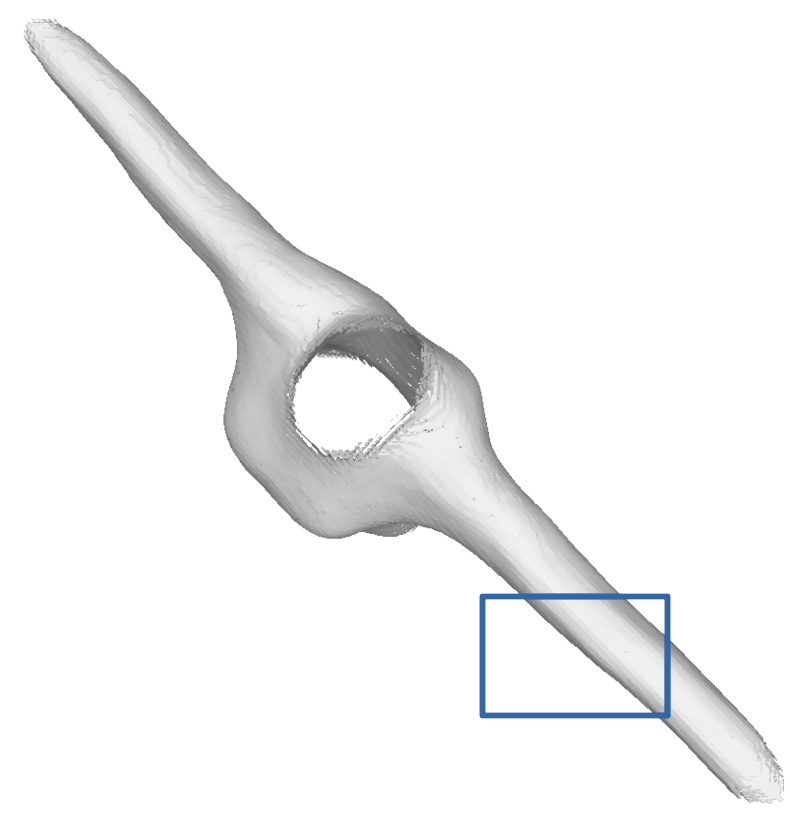} 
\end{subfigure} 
\begin{subfigure}{0.21\linewidth}
\includegraphics[width=\textwidth]{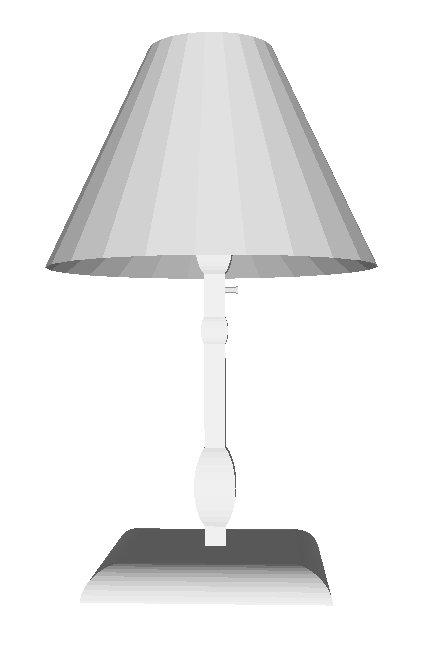}
\caption{GT}
\end{subfigure}
\hspace{0.03\linewidth}
\begin{subfigure}{0.22\linewidth}
\includegraphics[width=\textwidth]{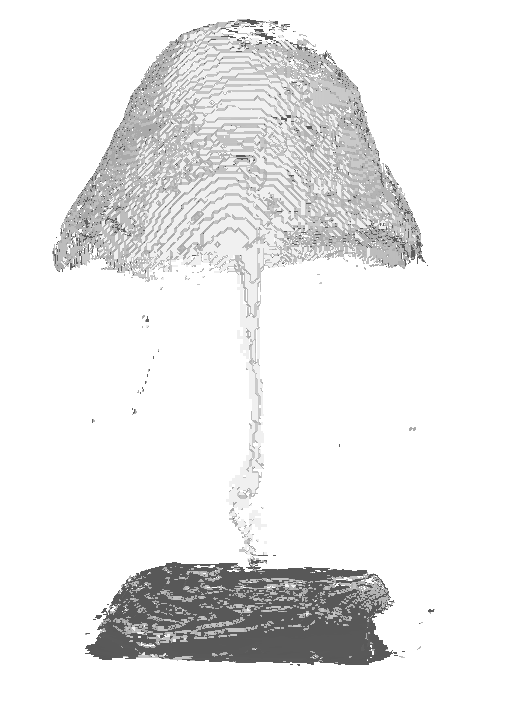}
\caption{NDF \cite{chibane2020ndf}}
\end{subfigure}
\hspace{0.03\linewidth}
\begin{subfigure}{0.21\linewidth}
\includegraphics[width=\textwidth]{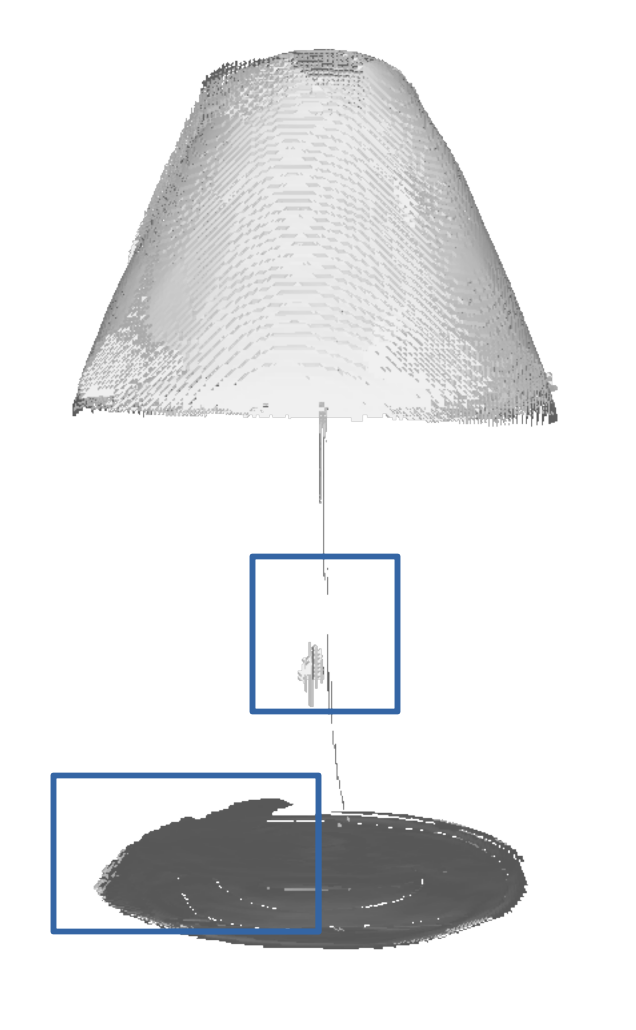} 
\caption{GIFS \cite{Ye_2022gifs}}
\end{subfigure}
\hspace{0.03\linewidth}
\begin{subfigure}{0.2\linewidth}
\includegraphics[width=\textwidth]{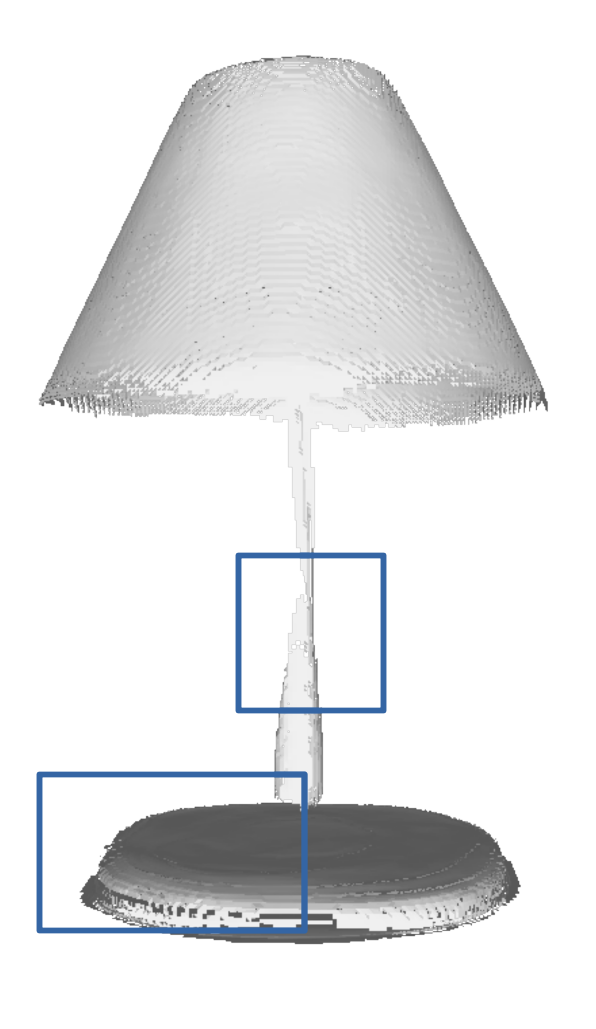}
\caption{VF}
\end{subfigure} 
\caption{\textbf{General shape representation}: VF can represent both open and multi-layered shapes. VF shows very accurate reconstructions even at the fine details outperforming recent methods in the reconstruction quality.}
\label{fig:open_qual}
\end{figure}

\subsection{Qualitative Results}

Overall, we can see that every method is able to accurately predict the target shapes with smooth results as shown in Figure \ref{fig:qual_water}. Furthermore, VF, together with NDF \cite{chibane2020ndf} and GIFS \cite{Ye_2022gifs} are able to generalize to open and more complex shapes, as in Figure \ref{fig:open_qual}.
Looking at Fig.~\ref{fig:qual_water} more closely, it can be seen that DeepSDF \cite{park2019deepsdf} and OccNet \cite{mescheder2019occnet} struggle in predicting thin surfaces as in the \textit{plane} reconstruction example. They also struggle to predict sharp angles in the \textit{chair} and \textit{sofa} examples; highlighted in blue. In comparison to NDF \cite{chibane2020ndf}, VF can predict smoother surfaces and does not suffer from small holes or artifacts, with a possible reason being the higher level of noise on the UDF gradient with respect to VF as described in Section \ref{sec:discussion}. Despite smoother than NDF \cite{chibane2020ndf}, GIFS \cite{Ye_2022gifs} still suffers from inconsistencies in the surface details and struggles to accurately reconstruct tubular parts as in \textit{chair} and \textit{lamp} examples. We provide additional visualizations in the supplementary material.

\subsection{Piecewise Planar Shape Representation}

\begin{table}[ht]
    \centering
    \resizebox{0.98\columnwidth}{!}{
    \begin{tabular}{ c | c c | c c | c c }\toprule
         \multirow{3}{*}{\textbf{Method}} & \multicolumn{2}{c|}{\textit{bookshelves}} & \multicolumn{2}{c|}{\textit{cabinets}} & \multicolumn{2}{c}{\textit{laptops}} \\
         & Chamfer & F1-Score & Chamfer & F1-Score & Chamfer & F1-Score \\
         & mean/median & 0.01 & mean/median & 0.01 & mean/median & 0.01 \\ \midrule
         VF & 1.116/0.491 & 75.49 & 0.299/0.192 & 82.16 & 0.081/0.063 & 94.38 \\
         Planar VF & 0.558/0.313 & 77.06 & 0.282/0.176 & 83.87 & 0.074/0.060 & 95.76 \\ \bottomrule
    \end{tabular}}
    \caption{Reconstruction results on the mainly planar ShapeNet~\cite{cheng2015shapenet} categories of bookshelf, cabinet and laptop. While the VF already shows good performance on piecewise planar shapes, we can observe even better performance with Planar VF.}
    \label{tab:planar_result}
\end{table}

\begin{figure}[!ht]
\centering
\begin{subfigure}{0.18\linewidth}
\includegraphics[width=\textwidth]{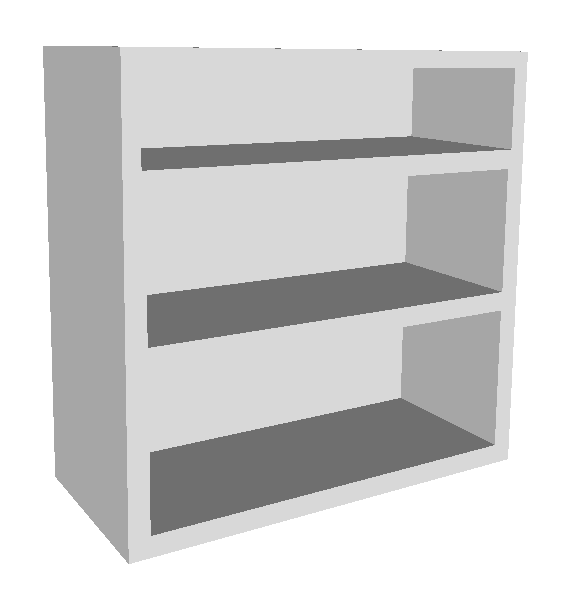}
\end{subfigure}
\hspace{0.01\linewidth}
\begin{subfigure}{0.35\linewidth}
\includegraphics[width=\textwidth]{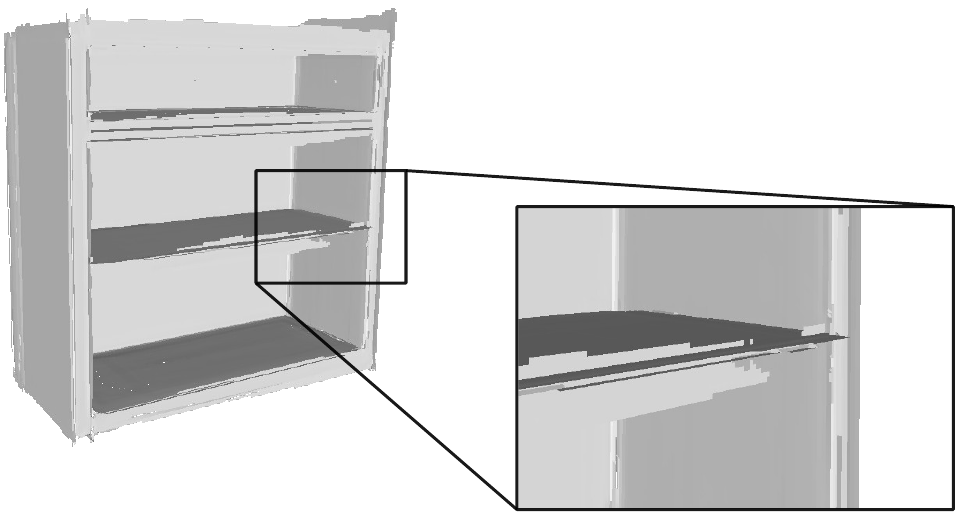}
\end{subfigure}
\hspace{0.01\linewidth}
\begin{subfigure}{0.35\linewidth}
\includegraphics[width=\textwidth]{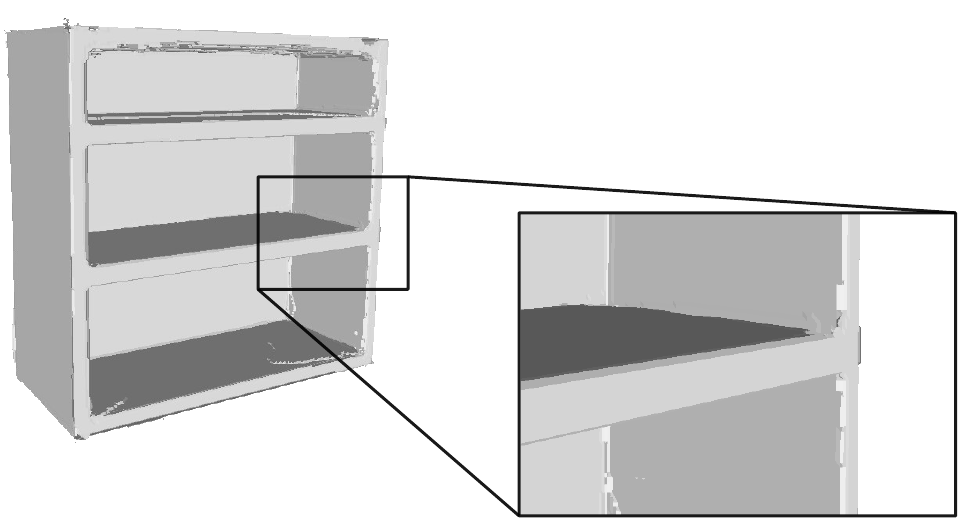}
\end{subfigure}
\begin{subfigure}{0.18\linewidth}
\includegraphics[width=\textwidth]{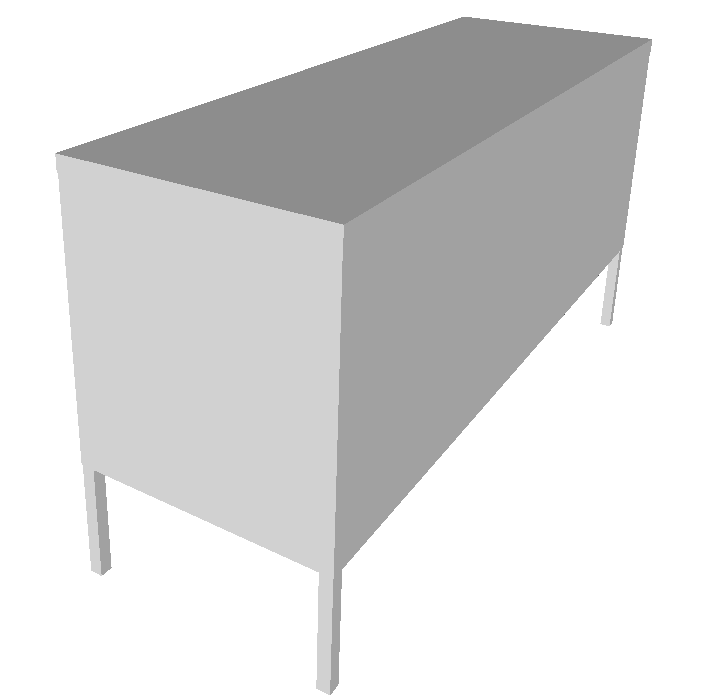}
\caption{GT}
\end{subfigure}
\hspace{0.01\linewidth}
\begin{subfigure}{0.35\linewidth}
\includegraphics[width=\textwidth]{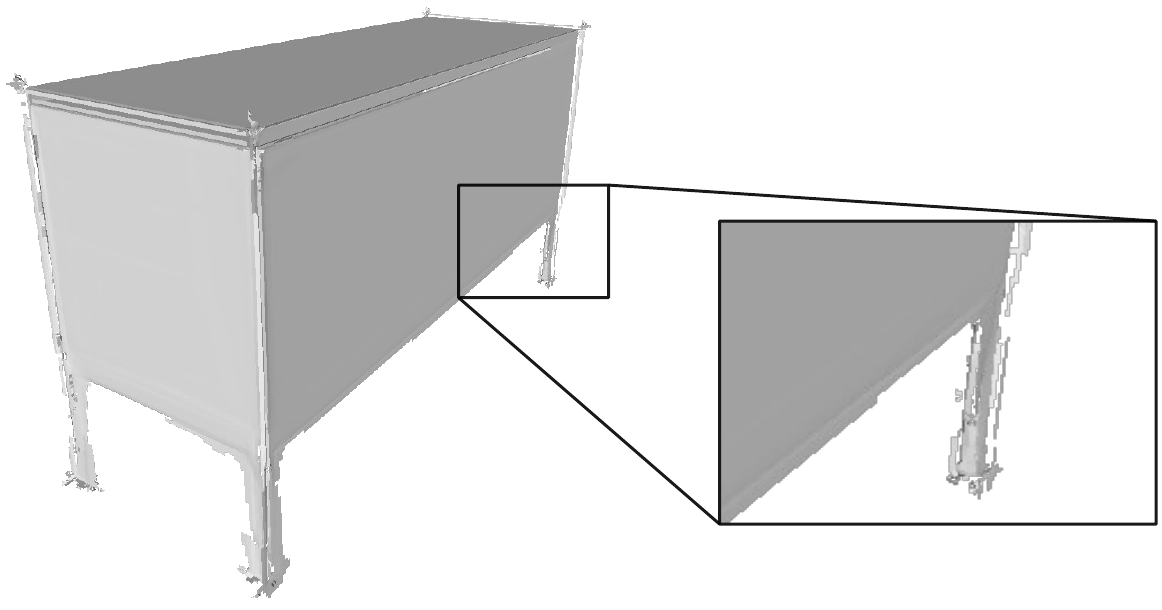}
\caption{VF}
\end{subfigure}
\hspace{0.01\linewidth}
\begin{subfigure}{0.35\linewidth}
\includegraphics[width=\textwidth]{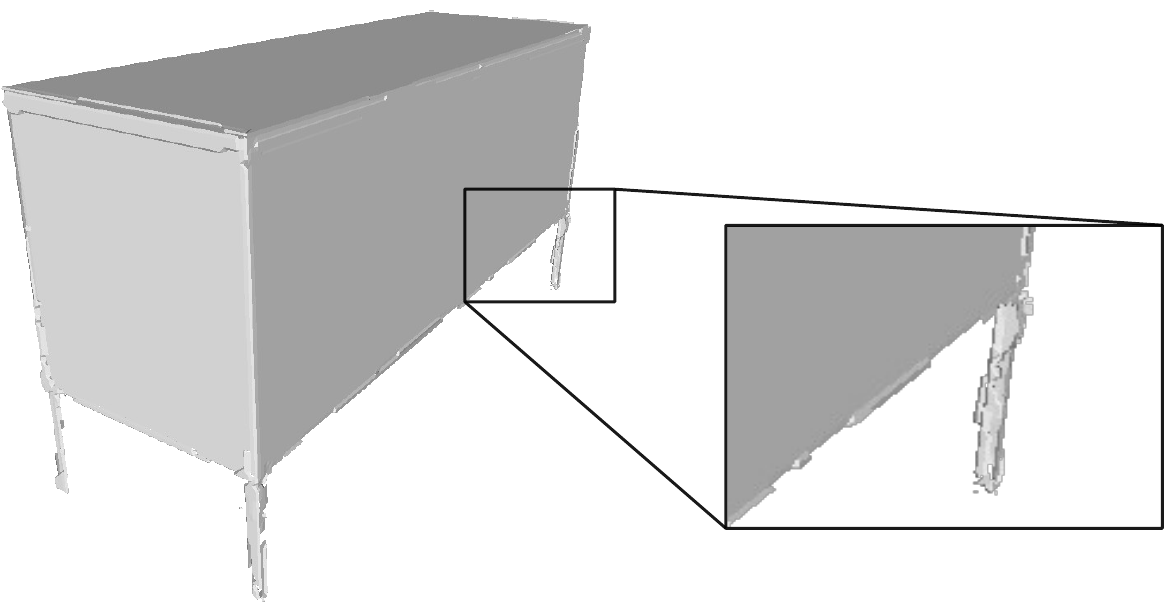} 
\caption{Planar VF}
\end{subfigure}
\caption{\textbf{Piecewise planar shape representation}: Planar VF represent highly planar objects with high accuracy, preserving straight angles at a detail level even better than VF.}
\label{fig:planar_qual}
\end{figure}

Here, we evaluate qualitatively and quantitatively the performance of applying the planar prior on VF - called Planar VF - on piecewise planar shapes and compare it to VF. This is done on the raw meshes from the \textit{bookshelves}, \textit{cabinets} and \textit{laptops} classes of Shapenet dataset \cite{cheng2015shapenet}.

In Figure \ref{fig:planar_qual}, we qualitatively compare the meshes generated with VF and Planar VF. Even though both methods are able to represent shapes accurately, Planar VF preserves even small planar regions better and can represent square angles sharply, showing the potential surface regularization capability of VF normals. This can be seen for example in the legs of the \textit{cabinet} or in the details of the edges. The higher qualitative accuracy is also supported by the quantitative evaluation in Table \ref{tab:planar_result}. These results show an additional advantage of being able to explicitly model directions, which allows one to adapt to the desired target shapes effectively.

\paragraph{Limitations.} Despite the significant results, VF does not ensure watertight meshes even when required and necessitates a non-standard MC algorithm for mesh inference. Furthermore, compared to the decades old binary occupancy and half a century old SDF, many of its properties and potential benefits/drawbacks are yet to be understood, which may be important in many downstream tasks.
\section{Conclusion}
In this paper, we revisit the INRs for representing 3D shapes and analyze the drawbacks of currently available representations. Popular representations, such as the SDF or binary occupancy, lack the expressive power to represent open or non-watertight surfaces. On the other hand, previous INR methods for open surfaces have problems with fast inference, accuracy, or training complexity. We propose to solve open surface representation by considering the novel VF and proved its theoretical suitability. We further showed the possibility to easily apply a shape prior on VF, allowing us to effectively fine-tune the method for piecewise planar objects. Our INR method using VF on standard architecture outperforms existing representations by a significant margin and opens up several new avenues in surface regularization and reconstruction.

\paragraph{\textbf{Acknowledgements}.} This research was funded by Align Technology Switzerland GmbH (project AlignTech-ETH). Research was also funded by the EU Horizon 2020 grant agreement No. 820434.

{\small
\bibliographystyle{ieee_fullname}
\bibliography{egbib}

\begin{thebibliography}{10}\itemsep=-1pt

\bibitem{anokhin2021image}
Ivan Anokhin, Kirill Demochkin, Taras Khakhulin, Gleb Sterkin, Victor
  Lempitsky, and Denis Korzhenkov.
\newblock Image generators with conditionally-independent pixel synthesis.
\newblock In {\em Proceedings of the IEEE/CVF Conference on Computer Vision and
  Pattern Recognition}, pages 14278--14287, 2021.

\bibitem{atzmon2020sal}
Matan Atzmon and Yaron Lipman.
\newblock Sal: Sign agnostic learning of shapes from raw data.
\newblock In {\em Proceedings of the IEEE/CVF Conference on Computer Vision and
  Pattern Recognition}, pages 2565--2574, 2020.

\bibitem{atzmon2021sald}
Matan Atzmon and Yaron Lipman.
\newblock {SALD:} sign agnostic learning with derivatives.
\newblock In {\em 9th International Conference on Learning Representations,
  {ICLR} 2021}, 2021.

\bibitem{ben2022digs}
Yizhak Ben-Shabat, Chamin~Hewa Koneputugodage, and Stephen Gould.
\newblock Digs: Divergence guided shape implicit neural representation for
  unoriented point clouds.
\newblock In {\em Proceedings of the IEEE/CVF Conference on Computer Vision and
  Pattern Recognition}, pages 19323--19332, 2022.

\bibitem{bhatnagar2019multi}
Bharat~Lal Bhatnagar, Garvita Tiwari, Christian Theobalt, and Gerard Pons-Moll.
\newblock Multi-garment net: Learning to dress 3d people from images.
\newblock In {\em proceedings of the IEEE/CVF international conference on
  computer vision}, pages 5420--5430, 2019.

\bibitem{Bhatnagar_2019_ICCV}
Bharat~Lal Bhatnagar, Garvita Tiwari, Christian Theobalt, and Gerard Pons-Moll.
\newblock Multi-garment net: Learning to dress 3d people from images.
\newblock In {\em Proceedings of the IEEE/CVF International Conference on
  Computer Vision (ICCV)}, October 2019.

\bibitem{chan2022efficient}
Eric~R Chan, Connor~Z Lin, Matthew~A Chan, Koki Nagano, Boxiao Pan, Shalini
  De~Mello, Orazio Gallo, Leonidas~J Guibas, Jonathan Tremblay, Sameh Khamis,
  et~al.
\newblock Efficient geometry-aware 3d generative adversarial networks.
\newblock In {\em Proceedings of the IEEE/CVF Conference on Computer Vision and
  Pattern Recognition}, pages 16123--16133, 2022.

\bibitem{cheng2015shapenet}
Angel~X Chang, Thomas Funkhouser, Leonidas Guibas, Pat Hanrahan, Qixing Huang,
  Zimo Li, Silvio Savarese, Manolis Savva, Shuran Song, Hao Su, et~al.
\newblock Shapenet: An information-rich 3d model repository.
\newblock {\em arXiv preprint arXiv:1512.03012}, 2015.

\bibitem{chen2021learning}
Yinbo Chen, Sifei Liu, and Xiaolong Wang.
\newblock Learning continuous image representation with local implicit image
  function.
\newblock In {\em Proceedings of the IEEE/CVF conference on computer vision and
  pattern recognition}, pages 8628--8638, 2021.

\bibitem{chen2019implicit}
Zhiqin Chen and Hao Zhang.
\newblock Learning implicit fields for generative shape modeling.
\newblock In {\em CVPR}, 2019.

\bibitem{chen2019implicitsdf1}
Zhiqin Chen and Hao Zhang.
\newblock Learning implicit fields for generative shape modeling.
\newblock In {\em Proceedings of the IEEE/CVF Conference on Computer Vision and
  Pattern Recognition}, pages 5939--5948, 2019.

\bibitem{chibane2020ndf}
Julian Chibane, Mohamad~Aymen mir, and Gerard Pons-Moll.
\newblock Neural unsigned distance fields for implicit function learning.
\newblock In H. Larochelle, M. Ranzato, R. Hadsell, M.~F. Balcan, and H. Lin,
  editors, {\em Advances in Neural Information Processing Systems}, volume~33,
  pages 21638--21652. Curran Associates, Inc., 2020.

\bibitem{choy2016voxelsmall1}
Christopher~B. Choy, Danfei Xu, JunYoung Gwak, Kevin Chen, and Silvio Savarese.
\newblock 3d-r2n2: A unified approach for single and multi-view 3d object
  reconstruction.
\newblock In Bastian Leibe, Jiri Matas, Nicu Sebe, and Max Welling, editors,
  {\em Computer Vision -- ECCV 2016}, pages 628--644, Cham, 2016. Springer
  International Publishing.

\bibitem{coughlan1999manhattan}
James~M Coughlan and Alan~L Yuille.
\newblock Manhattan world: Compass direction from a single image by bayesian
  inference.
\newblock In {\em Proceedings of the seventh IEEE international conference on
  computer vision}, volume~2, pages 941--947. IEEE, 1999.

\bibitem{deng2020hybrid_cvxnet}
Boyang Deng, Kyle Genova, Soroosh Yazdani, Sofien Bouaziz, Geoffrey Hinton, and
  Andrea Tagliasacchi.
\newblock Cvxnet: Learnable convex decomposition.
\newblock In {\em Proceedings of the IEEE/CVF Conference on Computer Vision and
  Pattern Recognition}, pages 31--44, 2020.

\bibitem{dupont2021coin}
Emilien Dupont, Adam Goli{\'n}ski, Milad Alizadeh, Yee~Whye Teh, and Arnaud
  Doucet.
\newblock Coin: Compression with implicit neural representations.
\newblock {\em arXiv preprint arXiv:2103.03123}, 2021.

\bibitem{eisenberger2020smooth}
Marvin Eisenberger, Zorah Lahner, and Daniel Cremers.
\newblock Smooth shells: Multi-scale shape registration with functional maps.
\newblock In {\em Proceedings of the IEEE/CVF Conference on Computer Vision and
  Pattern Recognition}, pages 12265--12274, 2020.

\bibitem{Feng2022}
Brandon~Y. Feng, Yinda Zhang, Danhang Tang, Ruofei Du, and Amitabh Varshney.
\newblock Prif: Primary ray-based implicit function.
\newblock In {\em Proceedings of the European Conference on Computer Vision
  (ECCV)}, October 2022.

\bibitem{gallup2010piecewise}
David Gallup, Jan-Michael Frahm, and Marc Pollefeys.
\newblock Piecewise planar and non-planar stereo for urban scene
  reconstruction.
\newblock In {\em 2010 IEEE computer society conference on computer vision and
  pattern recognition}, pages 1418--1425. IEEE, 2010.

\bibitem{Genova_2020_CVPR}
Kyle Genova, Forrester Cole, Avneesh Sud, Aaron Sarna, and Thomas Funkhouser.
\newblock Local deep implicit functions for 3d shape.
\newblock In {\em Proceedings of the IEEE/CVF Conference on Computer Vision and
  Pattern Recognition (CVPR)}, June 2020.

\bibitem{groueix20183d}
Thibault Groueix, Matthew Fisher, Vladimir~G Kim, Bryan~C Russell, and Mathieu
  Aubry.
\newblock 3d-coded: 3d correspondences by deep deformation.
\newblock In {\em ECCV}, 2018.

\bibitem{groueix2018papier}
Thibault Groueix, Matthew Fisher, Vladimir~G Kim, Bryan~C Russell, and Mathieu
  Aubry.
\newblock A papier-m{\^a}ch{\'e} approach to learning 3d surface generation.
\newblock In {\em Proceedings of the IEEE conference on computer vision and
  pattern recognition}, pages 216--224, 2018.

\bibitem{guillard2021meshudf}
Beno{\^{\i}}t Guillard, Federico Stella, and Pascal Fua.
\newblock Meshudf: Fast and differentiable meshing of unsigned distance field
  networks.
\newblock {\em CoRR}, abs/2111.14549, 2021.

\bibitem{guo2022neural}
Haoyu Guo, Sida Peng, Haotong Lin, Qianqian Wang, Guofeng Zhang, Hujun Bao, and
  Xiaowei Zhou.
\newblock Neural 3d scene reconstruction with the manhattan-world assumption.
\newblock In {\em Proceedings of the IEEE/CVF Conference on Computer Vision and
  Pattern Recognition}, pages 5511--5520, 2022.

\bibitem{hui2022neural}
Ka-Hei Hui, Ruihui Li, Jingyu Hu, and Chi-Wing Fu.
\newblock Neural wavelet-domain diffusion for 3d shape generation.
\newblock In {\em SIGGRAPH Asia 2022 Conference Papers}, pages 1--9, 2022.

\bibitem{ji2017voxel1}
Mengqi Ji, Juergen Gall, Haitian Zheng, Yebin Liu, and Lu Fang.
\newblock Surfacenet: An end-to-end 3d neural network for multiview stereopsis.
\newblock {\em 2017 IEEE International Conference on Computer Vision (ICCV)},
  Oct 2017.

\bibitem{kar2017voxel2}
Abhishek Kar, Christian H\"{a}ne, and Jitendra Malik.
\newblock Learning a multi-view stereo machine.
\newblock In I. Guyon, U.~V. Luxburg, S. Bengio, H. Wallach, R. Fergus, S.
  Vishwanathan, and R. Garnett, editors, {\em Advances in Neural Information
  Processing Systems}, volume~30. Curran Associates, Inc., 2017.

\bibitem{lei2022cadex}
Jiahui Lei and Kostas Daniilidis.
\newblock Cadex: Learning canonical deformation coordinate space for dynamic
  surface representation via neural homeomorphism.
\newblock In {\em Proceedings of the IEEE/CVF Conference on Computer Vision and
  Pattern Recognition}, pages 6624--6634, 2022.

\bibitem{lewiner2003efficient}
Thomas Lewiner, H{\'e}lio Lopes, Ant{\^o}nio~Wilson Vieira, and Geovan Tavares.
\newblock Efficient implementation of marching cubes' cases with topological
  guarantees.
\newblock {\em Journal of graphics tools}, 8(2):1--15, 2003.

\bibitem{litany2017deep}
Or Litany, Tal Remez, Emanuele Rodola, Alex Bronstein, and Michael Bronstein.
\newblock Deep functional maps: Structured prediction for dense shape
  correspondence.
\newblock In {\em ICCV}, 2017.

\bibitem{liu2020neural}
Lingjie Liu, Jiatao Gu, Kyaw~Zaw Lin, Tat-Seng Chua, and Christian Theobalt.
\newblock Neural sparse voxel fields.
\newblock {\em NeurIPS}, 2020.

\bibitem{maxwell1873treatise}
James~Clerk Maxwell.
\newblock {\em A treatise on electricity and magnetism}, volume~1.
\newblock Oxford: Clarendon Press, 1873.

\bibitem{mescheder2019occnet}
Lars Mescheder, Michael Oechsle, Michael Niemeyer, Sebastian Nowozin, and
  Andreas Geiger.
\newblock Occupancy networks: Learning 3d reconstruction in function space.
\newblock In {\em Proceedings of the IEEE/CVF Conference on Computer Vision and
  Pattern Recognition (CVPR)}, June 2019.

\bibitem{mildenhall2020nerf}
Ben Mildenhall, Pratul~P Srinivasan, Matthew Tancik, Jonathan~T Barron, Ravi
  Ramamoorthi, and Ren Ng.
\newblock Nerf: Representing scenes as neural radiance fields for view
  synthesis.
\newblock In {\em European conference on computer vision}, pages 405--421.
  Springer, 2020.

\bibitem{mueller2022instant}
Thomas M\"uller, Alex Evans, Christoph Schied, and Alexander Keller.
\newblock Instant neural graphics primitives with a multiresolution hash
  encoding.
\newblock {\em ACM Trans. Graph.}, 41(4):102:1--102:15, July 2022.

\bibitem{niemeyer2019occupancy}
Michael Niemeyer, Lars Mescheder, Michael Oechsle, and Andreas Geiger.
\newblock Occupancy flow: 4d reconstruction by learning particle dynamics.
\newblock In {\em Proceedings of the IEEE/CVF international conference on
  computer vision}, pages 5379--5389, 2019.

\bibitem{or2022stylesdf}
Roy Or-El, Xuan Luo, Mengyi Shan, Eli Shechtman, Jeong~Joon Park, and Ira
  Kemelmacher-Shlizerman.
\newblock Stylesdf: High-resolution 3d-consistent image and geometry
  generation.
\newblock In {\em Proceedings of the IEEE/CVF Conference on Computer Vision and
  Pattern Recognition}, pages 13503--13513, 2022.

\bibitem{osher2003level}
Stanley Osher and Ronald~P Fedkiw.
\newblock {\em Level set methods and dynamic implicit surfaces}, volume 153.
\newblock Springer, 2003.

\bibitem{Osher1988-fronts}
Stanley Osher and James~A Sethian.
\newblock Fronts propagating with curvature-dependent speed: Algorithms based
  on hamilton-jacobi formulations.
\newblock {\em Journal of Computational Physics}, 79(1):12--49, 1988.

\bibitem{ovsjanikov2012functional}
Maks Ovsjanikov, Mirela Ben-Chen, Justin Solomon, Adrian Butscher, and Leonidas
  Guibas.
\newblock Functional maps: a flexible representation of maps between shapes.
\newblock {\em ACM Transactions on Graphics (TOG)}, 31(4):1--11, 2012.

\bibitem{park2019deepsdf}
Jeong~Joon Park, Peter Florence, Julian Straub, Richard Newcombe, and Steven
  Lovegrove.
\newblock Deepsdf: Learning continuous signed distance functions for shape
  representation.
\newblock In {\em Proceedings of the IEEE/CVF Conference on Computer Vision and
  Pattern Recognition (CVPR)}, June 2019.

\bibitem{rella2022zero}
Edoardo~Mello Rella, Ajad Chhatkuli, Yun Liu, Ender Konukoglu, and Luc~Van
  Gool.
\newblock Zero pixel directional boundary by vector transform.
\newblock In {\em International Conference on Learning Representations}, 2022.

\bibitem{romanoni2019tapa}
Andrea Romanoni and Matteo Matteucci.
\newblock Tapa-mvs: Textureless-aware patchmatch multi-view stereo.
\newblock In {\em Proceedings of the IEEE/CVF International Conference on
  Computer Vision}, pages 10413--10422, 2019.

\bibitem{rozen2021moser}
Noam Rozen, Aditya Grover, Maximilian Nickel, and Yaron Lipman.
\newblock Moser flow: Divergence-based generative modeling on manifolds.
\newblock {\em Advances in Neural Information Processing Systems}, 34, 2021.

\bibitem{rudin1976principles}
Walter Rudin et~al.
\newblock {\em Principles of mathematical analysis}, volume~3.
\newblock McGraw-hill New York, 1976.

\bibitem{saito2019implicitbin3}
Shunsuke Saito, Zeng Huang, Ryota Natsume, Shigeo Morishima, Angjoo Kanazawa,
  and Hao Li.
\newblock Pifu: Pixel-aligned implicit function for high-resolution clothed
  human digitization.
\newblock In {\em Proceedings of the IEEE/CVF International Conference on
  Computer Vision}, pages 2304--2314, 2019.

\bibitem{Shaham_2021_CVPR}
Tamar~Rott Shaham, Michael Gharbi, Richard Zhang, Eli Shechtman, and Tomer
  Michaeli.
\newblock Spatially-adaptive pixelwise networks for fast image translation.
\newblock In {\em Proceedings of the IEEE/CVF Conference on Computer Vision and
  Pattern Recognition (CVPR)}, pages 14882--14891, June 2021.

\bibitem{sitzmann2020implicit}
Vincent Sitzmann, Julien Martel, Alexander Bergman, David Lindell, and Gordon
  Wetzstein.
\newblock Implicit neural representations with periodic activation functions.
\newblock {\em Advances in Neural Information Processing Systems},
  33:7462--7473, 2020.

\bibitem{sitzmann2021light}
Vincent Sitzmann, Semon Rezchikov, Bill Freeman, Josh Tenenbaum, and Fredo
  Durand.
\newblock Light field networks: Neural scene representations with
  single-evaluation rendering.
\newblock {\em Advances in Neural Information Processing Systems},
  34:19313--19325, 2021.

\bibitem{sommersang2022}
Christiane Sommer, Lu Sang, David Schubert, and Daniel Cremers.
\newblock Gradient-sdf: A semi-implicit surface representation for 3d
  reconstruction.
\newblock In {\em IEEE/CVF International Conference on Computer Vision and
  Pattern Recognition (CVPR)}, 2022.

\bibitem{sun2022direct}
Cheng Sun, Min Sun, and Hwann-Tzong Chen.
\newblock Direct voxel grid optimization: Super-fast convergence for radiance
  fields reconstruction.
\newblock In {\em Proceedings of the IEEE/CVF Conference on Computer Vision and
  Pattern Recognition}, pages 5459--5469, 2022.

\bibitem{tancik2020fourier}
Matthew Tancik, Pratul Srinivasan, Ben Mildenhall, Sara Fridovich-Keil, Nithin
  Raghavan, Utkarsh Singhal, Ravi Ramamoorthi, Jonathan Barron, and Ren Ng.
\newblock Fourier features let networks learn high frequency functions in low
  dimensional domains.
\newblock {\em Advances in Neural Information Processing Systems},
  33:7537--7547, 2020.

\bibitem{tulsiani2017voxelsmall2}
Shubham Tulsiani, Tinghui Zhou, Alexei~A. Efros, and Jitendra Malik.
\newblock Multi-view supervision for single-view reconstruction via
  differentiable ray consistency.
\newblock In {\em Proceedings of the IEEE Conference on Computer Vision and
  Pattern Recognition (CVPR)}, July 2017.

\bibitem{ueda2022neural}
Itsuki Ueda, Yoshihiro Fukuhara, Hirokatsu Kataoka, Hiroaki Aizawa, Hidehiko
  Shishido, and Itaru Kitahara.
\newblock Neural density-distance fields.
\newblock In {\em Proceedings of the European Conference on Computer Vision},
  2022.

\bibitem{venkatesh2020dude}
Rahul Venkatesh, Sarthak Sharma, Aurobrata Ghosh, Laszlo Jeni, and Maneesh
  Singh.
\newblock Dude: Deep unsigned distance embeddings for hi-fidelity
  representation of complex 3d surfaces.
\newblock {\em arXiv preprint arXiv:2011.02570}, 2020.

\bibitem{wu2017bigvoxel1}
Jiajun Wu, Yifan Wang, Tianfan Xue, Xingyuan Sun, Bill Freeman, and Josh
  Tenenbaum.
\newblock Marrnet: 3d shape reconstruction via 2.5d sketches.
\newblock In I. Guyon, U.~V. Luxburg, S. Bengio, H. Wallach, R. Fergus, S.
  Vishwanathan, and R. Garnett, editors, {\em Advances in Neural Information
  Processing Systems}, volume~30. Curran Associates, Inc., 2017.

\bibitem{wu2015smallvoxel3}
Zhirong Wu, Shuran Song, Aditya Khosla, Fisher Yu, Linguang Zhang, Xiaoou Tang,
  and Jianxiong Xiao.
\newblock 3d shapenets: A deep representation for volumetric shapes.
\newblock In {\em Proceedings of the IEEE Conference on Computer Vision and
  Pattern Recognition (CVPR)}, June 2015.

\bibitem{Xiu_2022_CVPR}
Yuliang Xiu, Jinlong Yang, Dimitrios Tzionas, and Michael~J. Black.
\newblock Icon: Implicit clothed humans obtained from normals.
\newblock In {\em Proceedings of the IEEE/CVF Conference on Computer Vision and
  Pattern Recognition (CVPR)}, pages 13296--13306, June 2022.

\bibitem{xu2020planar}
Qingshan Xu and Wenbing Tao.
\newblock Planar prior assisted patchmatch multi-view stereo.
\newblock In {\em Proceedings of the AAAI Conference on Artificial
  Intelligence}, volume~34, pages 12516--12523, 2020.

\bibitem{yariv2021volume}
Lior Yariv, Jiatao Gu, Yoni Kasten, and Yaron Lipman.
\newblock Volume rendering of neural implicit surfaces.
\newblock {\em Advances in Neural Information Processing Systems},
  34:4805--4815, 2021.

\bibitem{Ye_2022gifs}
Jianglong Ye, Yuntao Chen, Naiyan Wang, and Xiaolong Wang.
\newblock Gifs: Neural implicit function for general shape representation.
\newblock In {\em Proceedings of the IEEE/CVF Conference on Computer Vision and
  Pattern Recognition (CVPR)}, pages 12829--12839, June 2022.

\bibitem{yifan2020neural}
Wang Yifan, Noam Aigerman, Vladimir~G Kim, Siddhartha Chaudhuri, and Olga
  Sorkine-Hornung.
\newblock Neural cages for detail-preserving 3d deformations.
\newblock In {\em Proceedings of the IEEE/CVF Conference on Computer Vision and
  Pattern Recognition}, pages 75--83, 2020.

\bibitem{yu2021plenoxels}
Alex Yu, Sara Fridovich-Keil, Matthew Tancik, Qinhong Chen, Benjamin Recht, and
  Angjoo Kanazawa.
\newblock Plenoxels: Radiance fields without neural networks.
\newblock {\em arXiv preprint arXiv:2112.05131}, 2021.

\bibitem{zhang2018bigvoxel2}
Xiuming Zhang, Zhoutong Zhang, Chengkai Zhang, Josh Tenenbaum, Bill Freeman,
  and Jiajun Wu.
\newblock Learning to reconstruct shapes from unseen classes.
\newblock In S. Bengio, H. Wallach, H. Larochelle, K. Grauman, N. Cesa-Bianchi,
  and R. Garnett, editors, {\em Advances in Neural Information Processing
  Systems}, volume~31. Curran Associates, Inc., 2018.

\bibitem{zheng2021deep}
Zerong Zheng, Tao Yu, Qionghai Dai, and Yebin Liu.
\newblock Deep implicit templates for 3d shape representation.
\newblock In {\em Proceedings of the IEEE/CVF Conference on Computer Vision and
  Pattern Recognition}, pages 1429--1439, 2021.

\end{thebibliography}
}

\renewcommand\thesection{\Alph{section}}
\section{Proofs of VF Properties}
In this Section, we provide the proofs of Properties stated in the main text in Section \ref{sec:prop}. For clarity, we restate the representation definition and the properties followed by the actual proofs.

We define VF (see Section \ref{sec:prop}) as follows.
\begin{align}
\begin{split}
    f(\mathsf{x}) &= \mathsf{v} \quad \text{with} \quad \mathsf{x}\in\Omega, \quad \mathsf{v}\in\Gamma \\
    \mathsf{v} &= -\frac{\mathsf{x}- \mathsf{\hat{x}}_S}{\Vert \mathsf{x} - \mathsf{\hat{x}}_S\Vert} \ \ \text{and}\\
    \mathsf{\hat{x}}_S &= \argmin _{\mathsf{x}_S\in \Pi } \Vert \mathsf{x}-\mathsf{x}_S\Vert \quad \text{if} \quad \mathsf{x} \notin \Pi,\  \text{otherwise:} \\
    \mathsf{v} &= -\frac{\mathsf{x}'- \mathsf{\hat{x}}_S}{\Vert \mathsf{x}'- \mathsf{\hat{x}}_S\Vert} \quad \text{with} \\ \mathsf{x}' &= \lim_{\Vert \mathsf{\epsilon} \Vert \to 0}{\mathsf{x} + \mathsf{\epsilon}}; \ \mathsf{x}' \notin \Pi \ \text{and} \ \epsilon > 0.
\end{split}
\label{eq:vectransforms}
\tag{1}
\end{align}
The arbitrary constraint $\epsilon > 0$ implies that each component is greater than 0. Based on the definition of VF, we prove the following properties, listed in Section \ref{sec:prop}.

\begin{figure*}[t]
\centering
\begin{subfigure}{0.16\linewidth}
\includegraphics[width=\textwidth]{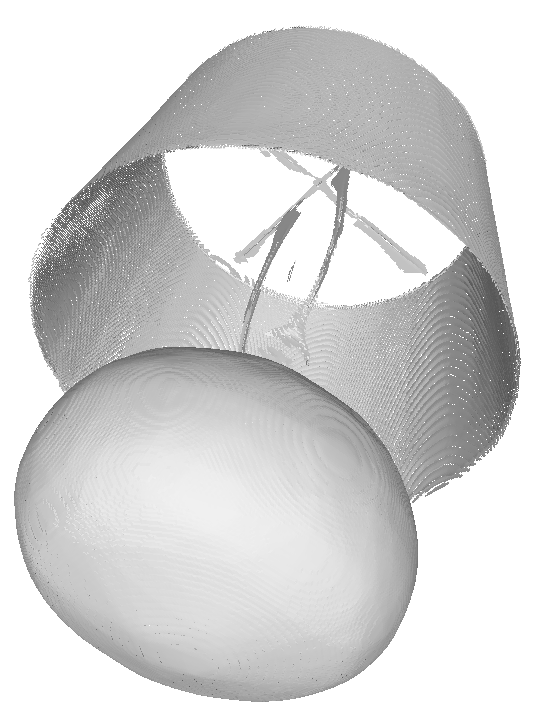}
\end{subfigure}
\hspace{0.03\linewidth}
\begin{subfigure}{0.23\linewidth}
\includegraphics[width=\textwidth]{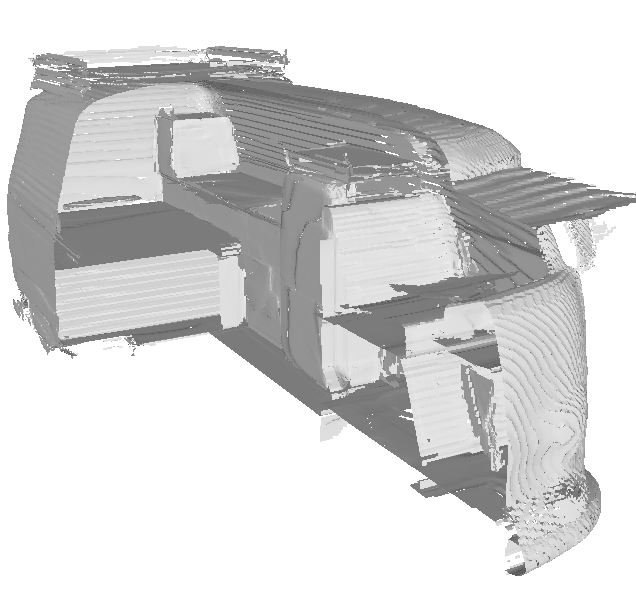}
\end{subfigure}
\hspace{0.03\linewidth}
\begin{subfigure}{0.20\linewidth}
\includegraphics[width=\textwidth]{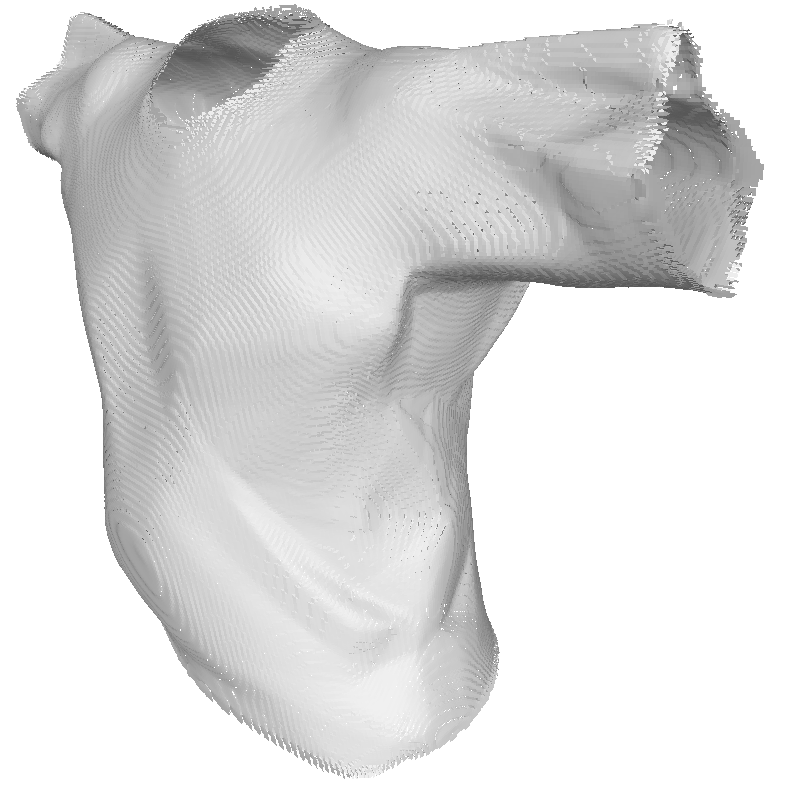} 
\end{subfigure}
\hspace{0.03\linewidth}
\begin{subfigure}{0.23\linewidth}
\includegraphics[width=\textwidth]{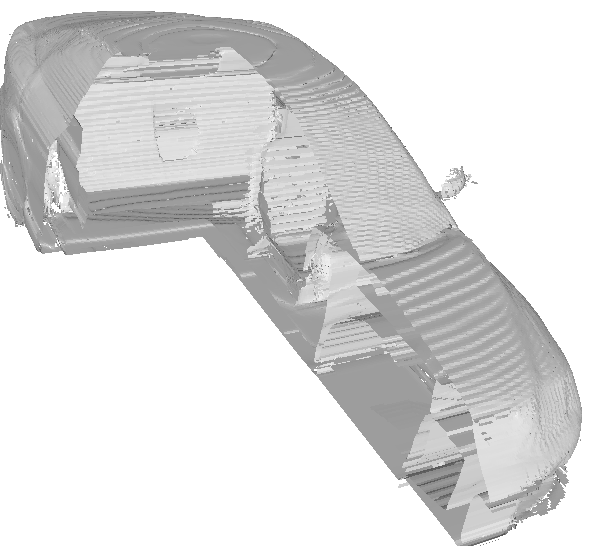} 
\end{subfigure}
\caption{\textbf{General surface visualizations.} The proposed VF representation can encode any type of surface and can be converted to a mesh representation.}
\label{fig:good_qual}
\end{figure*}

\paragraph{Property 3.1}
\emph{The vector field $\mathsf{v} = f(\mathsf{x})$ is equal to the negative gradient of the unsigned distance field, \ie,  $-\nabla u(\mathsf{x})$, except at the discontinuities at the surface points $\{\mathsf{x}_S\}$ and at points at equal distance from multiple surface points.}

\begin{proof} Consider $u(\mathsf{x})=\Vert \mathsf{x}- \mathsf{\hat{x}}_S \Vert$. Except at the discontinuities on the surface and at points at equal distance from multiple surfaces, $\nabla u(\mathsf{x})$ is defined as,
\begin{equation}
    u(\mathsf{x}) = \Vert \mathsf{x} - \hat{\mathsf{x}}_S \Vert.
\end{equation}
The gradient of the distance field is given by,
$\nabla u(\mathsf{x}) = \frac{\mathsf{x}- \mathsf{\hat{x}}_S}{\Vert \mathsf{x} - \mathsf{\hat{x}}_S\Vert}$, thus $\nabla u(\mathsf{x}) = -\mathsf{v}$ for points outside the surface. Please refer to \cite{osher2003level}, page 9 regarding a discussion on the gradient of the distance field. The gradient points to the direction of increasing distance. The result is valid for all smooth surfaces. For piece-wise smooth surfaces, additional discontinuities may arise, specifically where the vector field $\mathsf{v}$ diverges. Note that at the discontinuities, we can still consider the property to be true provided that we use the same one hand-side limit, thanks to the definitions 4.1 and 5.1 in Rudin~\cite{rudin1976principles}.
\end{proof}

\paragraph{Property 3.2}
\emph{The vector field $\mathsf{v} = f(\mathsf{x})$ is equal to the surface normal as it approaches the surface.}

\begin{proof}
For any implicit function $F(\mathsf{x})=0$, the gradient is equal to the surface normal as $\mathsf{x}$ approaches the surface (please refer to \cite{osher2003level}, pages 9 -- 12). Note that at the surface, there are two opposite directions depending on the limit chosen, and both are surface normals, equally valid for our purpose.
\end{proof}

\paragraph{Property 3.3}
\emph{Consider the VF representation $\mathsf{v} = f(\mathsf{x})$ of a piecewise smooth surface as defined in Eq.~\eqref{eq:vectransforms}, and the following transform:}
\begin{align}
\label{eq:divvts}
\begin{split}
    & g(\mathsf{x}) = D_\Phi f(\mathsf{x}). \\
\end{split}
\tag{2}
\end{align}
\emph{Here, $D_\Phi$ is the operator for flux density, defined as flux per unit cross-sectional area \cite{maxwell1873treatise}, measured using an infinitesimal spherical surface.
A point $\mathsf{x} \in \mathbb{R}^3$ then is a surface point, $\mathsf{x} \in \Pi$, if and only if it belongs to the zero level set of $g(\mathsf{x})+1$, \ie, }
\begin{equation}
\label{eq:invtransforms}
\Pi = L_0(g+1).    \tag{3}
\end{equation}

Before proving Prop. 3.3, we clarify the meaning of flux density and how it is computed in practice. The flux $\Phi$ of a vector field through an oriented surface is the surface integral of the inner product between the vector field and the unit vector normal to the surface. We now give three examples. \emph{i}: Given a flat square surface of side $L$ and a constant vector field $\mathsf{z}$ with angle $\theta$ between it and the surface normal, the flux is $\Phi = L^2 \cos{\theta} \cdot \|\mathsf{z}\|$. \emph{ii}: If we consider a sphere with radius $R$ in the same constant vector field $\mathsf{z}$, then the flux is inward in half the sphere and outward in the other half. By computing the surface integral, we find that both the inward and outward components have magnitude $\Phi = \pi R^2 \cdot \|\mathsf{z}\|$ but opposite sign so that the resulting flux is $0$. \emph{iii}: As a third example, we take the same sphere with radius $R$ and place it in a radial vector field $\mathsf{z}$ pointing outward with constant magnitude $\|\mathsf{z}\| = \zeta$ and centered on the sphere center. Here, the field vectors are perpendicular to the sphere on its surface so that the overall flux is $\Phi = 4 \pi R^2 \cdot \|\mathsf{z}\|$. Starting from the flux, flux density is the amount of flux divided by the unit area. Similarly to the flux, we define the area $A$ as the surface integral of the absolute value of the inner product between the unit vector parallel to the vector field in each point and the unit vector normal to the surface. By applying this to the three previous examples, we obtain: \emph{i} $A = L^2 \cos{\theta}$, \emph{ii} $A = 2 \pi R^2$, \emph{iii} $A = 4 \pi R^2$. We note that here (\emph{ii}) we do not have components with opposite sign and in all the three examples the area is not proportional to the magnitude of the field. From the definitions of flux and area it is then possible to compute the flux density $D_\Phi$ in the three examples as the ratio between the flux $\Phi$ and the area $A$. In practice, we obtain \emph{i} $D_\Phi = \|\mathsf{z}\|$, \emph{ii} $D_\Phi = 0$, \emph{iii} $D_\Phi = \|\mathsf{z}\|$. We note that, given the vector field properties and the sphere positioning chosen, $D_\Phi$ does not depend on the size of the surface or the relative angle $\theta$ (in example \emph{i}). Thus, the value does not change when the size of the spherical surface tends to zero as is the case here. Furthermore, given the properties of VF and the smoothness assumption, we can consider that at any point in space (except at equal distance from multiple surface points, a case that we handle in the proof) the VF is constant in direction in an infinitesimal sphere. Therefore, in all these points, the area is $A = 2 \pi R^2$, which is the cross-sectional area of the two sides. Having clarified the flux density as the flux through a spherical surface divided by its cross-sectional area when the radius of such sphere tends to zero, we can prove Property 3.3.

\begin{figure}[ht]
\centering
\begin{subfigure}{0.8\linewidth}
    \includegraphics[width=\textwidth]{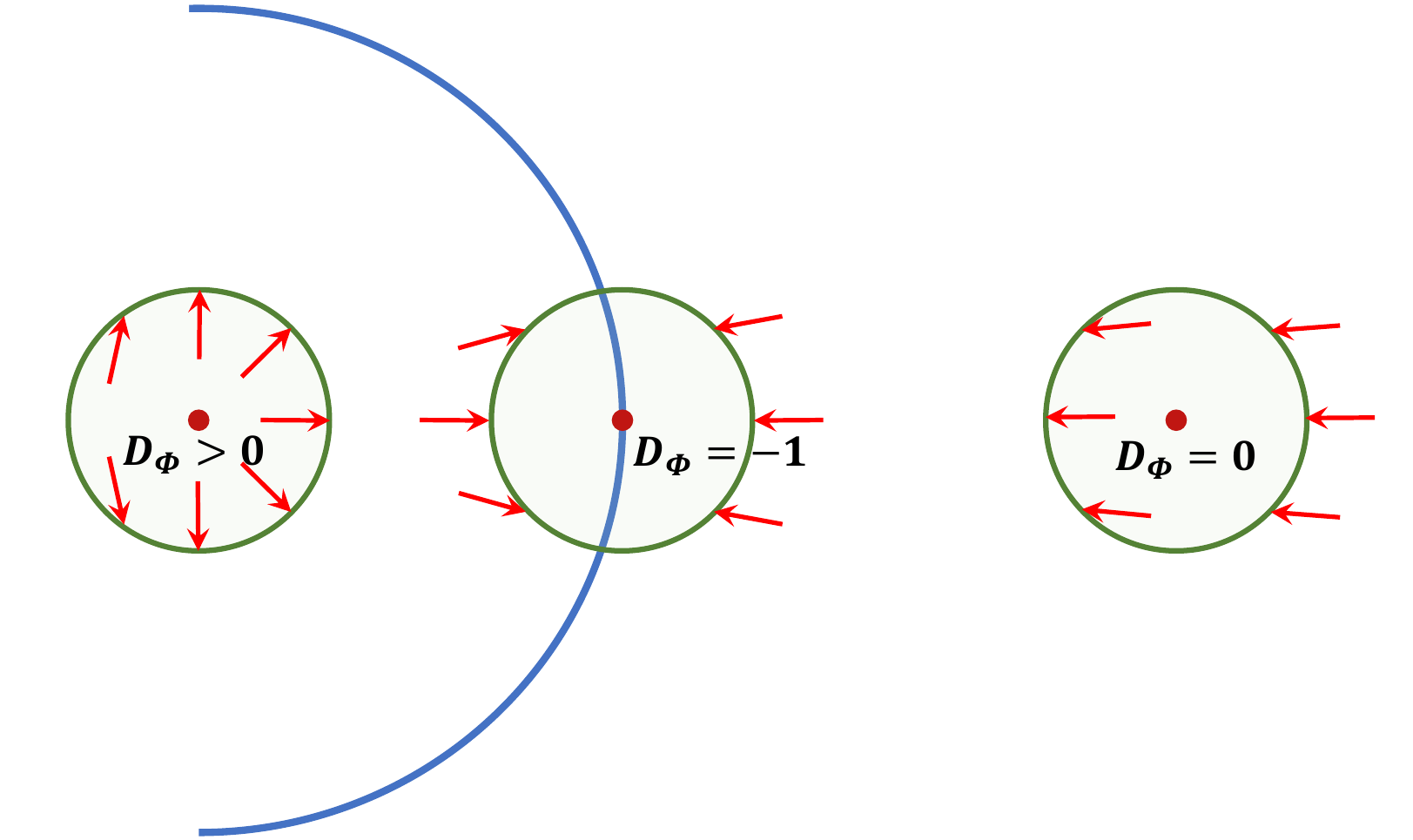}
    \caption{\textbf{Proof visualization.} We visualize the three cases considered in the proof of Property 3.3. Note that it does not matter on which side of the object-surface the points lie.}
    \label{fig:proof}
\end{subfigure} \\
\begin{subfigure}{0.8\linewidth}
    \includegraphics[width=\textwidth]{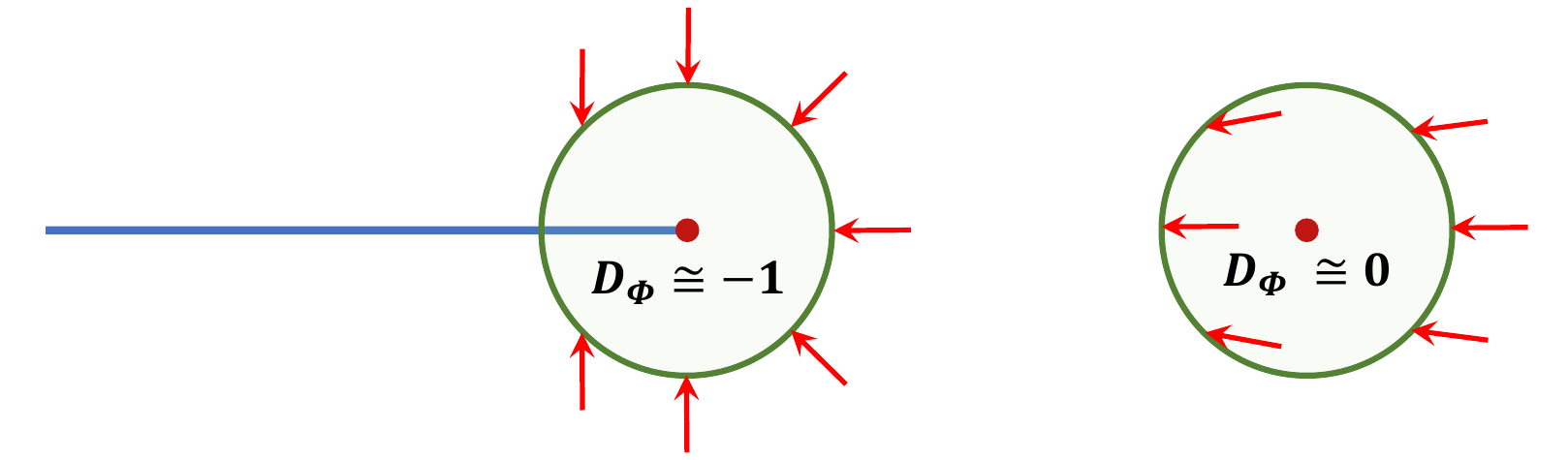}
    \caption{\textbf{Discontinuities.} We visualize the case of the presence of discontinuous object-surfaces to show that flux density can be used to detect then as well.}
    \label{fig:ext_vf}
\end{subfigure} \\
\caption{\textbf{Flux density visualization.} We show a 2D visualization of how flux density is computed on VF. Note that the actual flux density is computed as the sphere-surface radius tends to zero which makes the effect of the object-surface curvature on VF direction negligible. This allows us to use the assumption that at an infinitesimal level, the direction of VF remains constant with only the normal flip being the non-neglible change. Only discontinuity points are excluded from this assumption.}
\label{fig:fl_dens}
\end{figure}

\begin{proof}
To avoid confusion throughout the proof, we refer to the surface of the object to represent as object-surface and to the surface of the sphere used to compute the flux density as sphere-surface.
Let us start proving that, if a point belongs to the object-surface, then it is in the zero level set of Eq.~\eqref{eq:invtransforms}. We measure the flux density as the flux through a spherical surface divided by its cross-sectional area when the radius of such sphere tends to zero. In the infinitesimal sphere around any object-surface point, using Property 3.2, the only component of VF present is normal to the object-surface, with tangential components approaching $0$. Given the assumption of piecewise smooth object-surfaces, the field can be considered constant in an infinitesimal neighborhood, except from the orientation flip at the object-surface. By taking as positive the orientation outward of the sphere-surface, the flux through such infinitesimal sphere-surface centered on an object-surface point is $ - 2 \pi R^2$. Similarly, the cross-sectional area of the sphere-surface is $2 \pi R^2$. By taking the ratio of the two values, we obtain the flux density of $-1$, which demonstrates that the point on the object-surface belongs to the zero level set $L_0(g+1)$.

In order to prove that only object-surface points are in the level set, we can identify two cases of points $\mathsf{\hat{x}}$ not belonging to the object-surface. We consider case \emph{I} when the given point $\mathsf{\hat{x}}$ is equidistant to multiple object-surface points, multiple outward vectors stem from its infinitesimal neighborhood, consequently producing a positive flux density value. We consider case \emph{II} when a point $\mathsf{\hat{x}}$ has a single closest object-surface point. Given the smoothness assumption, the field can be considered constant inside the sphere-surface as its radius tends to zero. Therefore, in case \emph{II} the flux density is zero because in each hemisphere the flux has magnitude $\pi R^2$ but opposite sign. Fig. \ref{fig:proof} gives an intuition of the levels of flux density in the three types of points we explained.
\end{proof}

Having proved property 3.3, we show intuitively that it is possible to extend it to discontinuities without differences in its practical application. There is always a sizable gap in flux density between points that lie on the object-surface and points outside. As shown in Fig. \ref{fig:ext_vf}, this can be understood by observing that when an infinitesimal sphere-surface is built around a point on the object-surface (even at a discontinuity), then the VF always points inward on such sphere-surface. Therefore, we always obtain flux density values of $-1$. On the other hand, when considering an infinitesimal sphere-surface centered on points outside the object-surface, there is always a component that is pointing outward of the sphere-surface which brings the flux density value around $0$.

\begin{figure*}[t!]
\centering
\begin{subfigure}{0.13\linewidth}
\includegraphics[width=\textwidth]{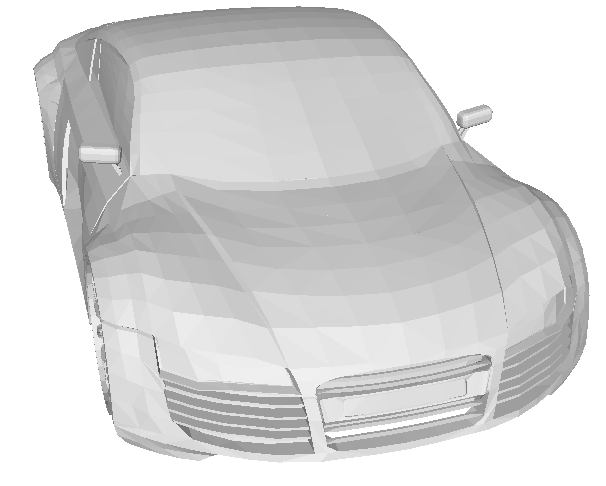}
\end{subfigure}
\hspace{0.02\linewidth}
\begin{subfigure}{0.13\linewidth}
\includegraphics[width=\textwidth]{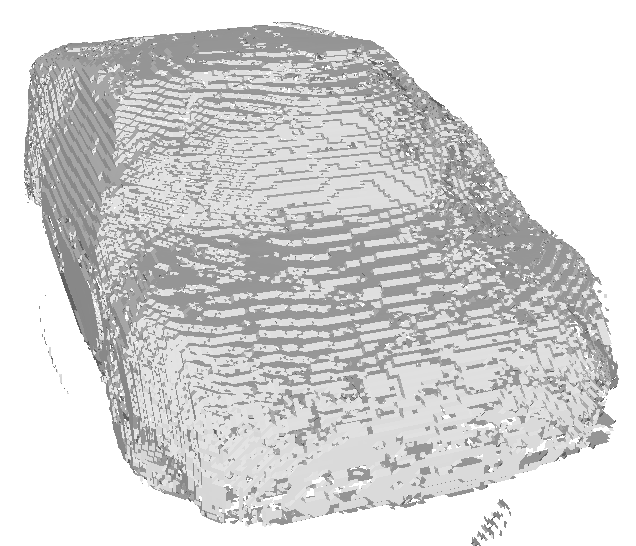}
\end{subfigure}
\hspace{0.02\linewidth}
\begin{subfigure}{0.13\linewidth}
\includegraphics[width=\textwidth]{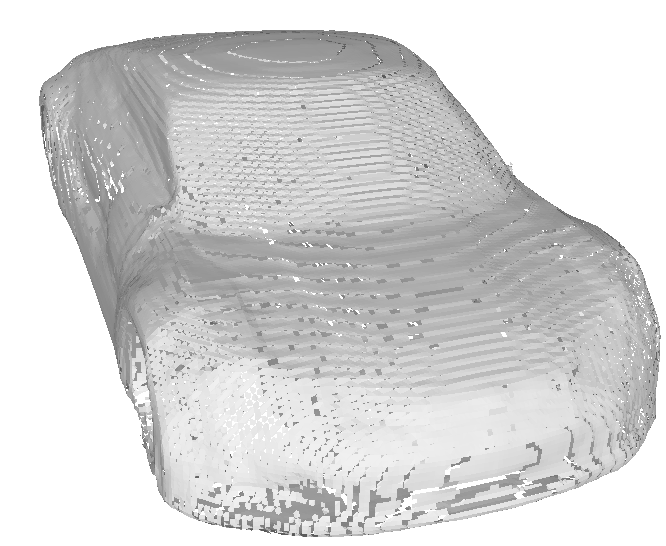} 
\end{subfigure}
\hspace{0.02\linewidth}
\begin{subfigure}{0.13\linewidth}
\includegraphics[width=\textwidth]{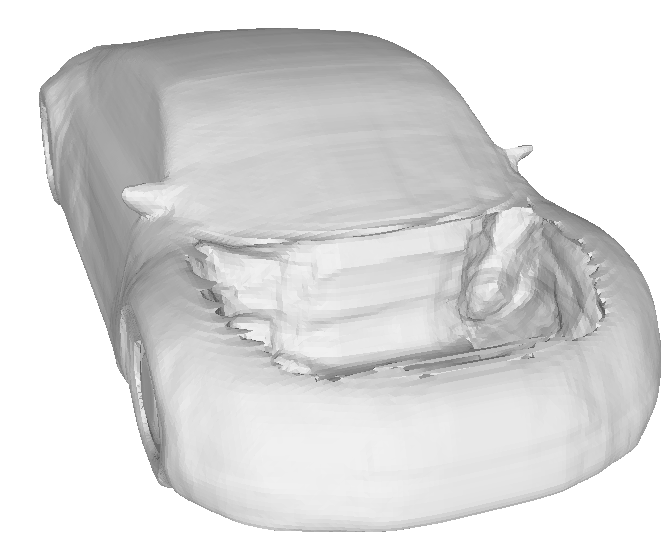} 
\end{subfigure}
\hspace{0.02\linewidth}
\begin{subfigure}{0.13\linewidth}
\includegraphics[width=\textwidth]{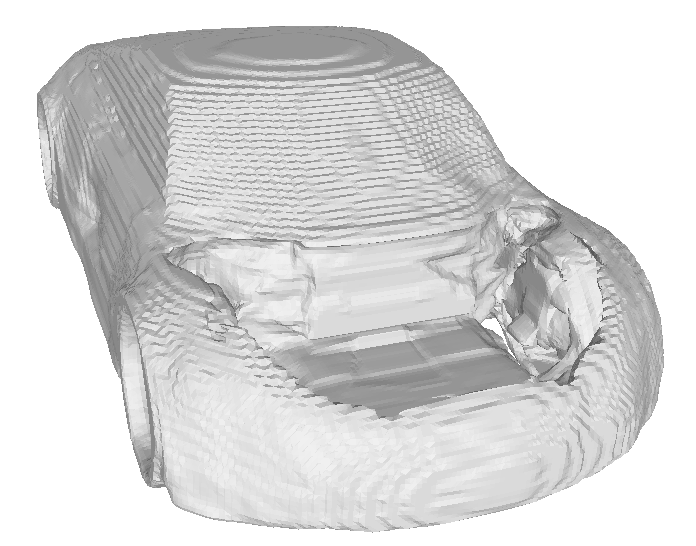} 
\end{subfigure}
\hspace{0.02\linewidth}
\begin{subfigure}{0.13\linewidth}
\includegraphics[width=\textwidth]{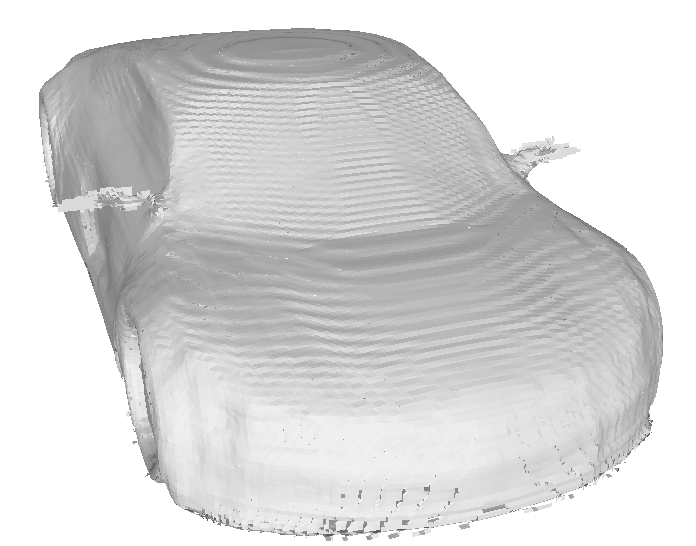} 
\end{subfigure} 
\begin{subfigure}{0.13\linewidth}
\includegraphics[width=\textwidth]{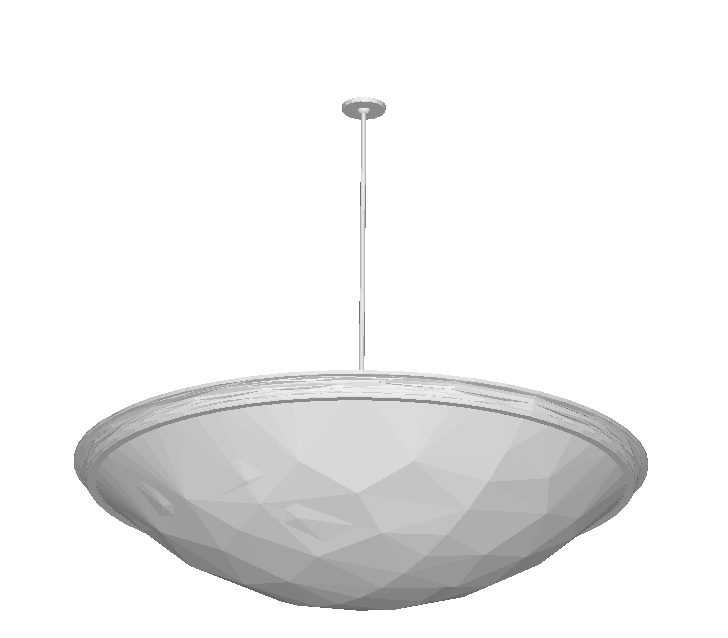}
\end{subfigure}
\hspace{0.02\linewidth}
\begin{subfigure}{0.13\linewidth}
\includegraphics[width=\textwidth]{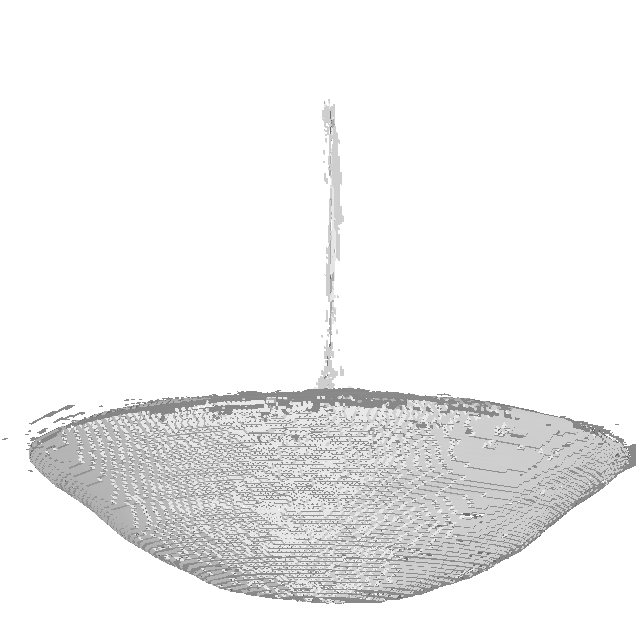}
\end{subfigure}
\hspace{0.02\linewidth}
\begin{subfigure}{0.13\linewidth}
\includegraphics[width=\textwidth]{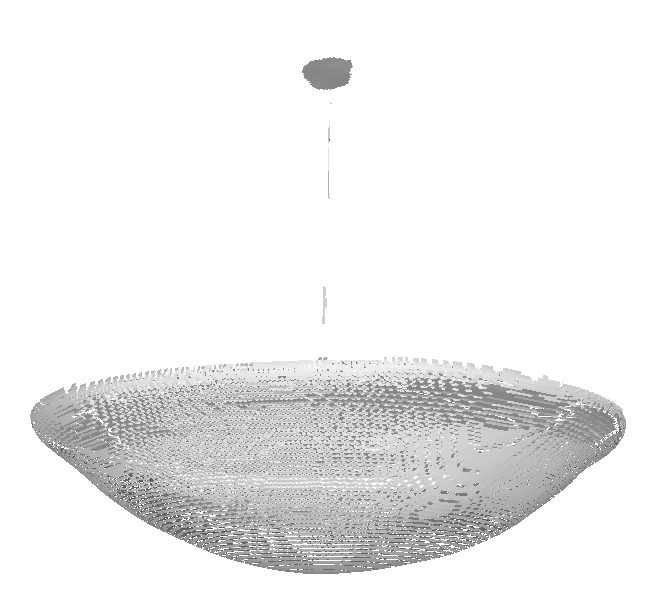} 
\end{subfigure}
\hspace{0.02\linewidth}
\begin{subfigure}{0.13\linewidth}
\includegraphics[width=\textwidth]{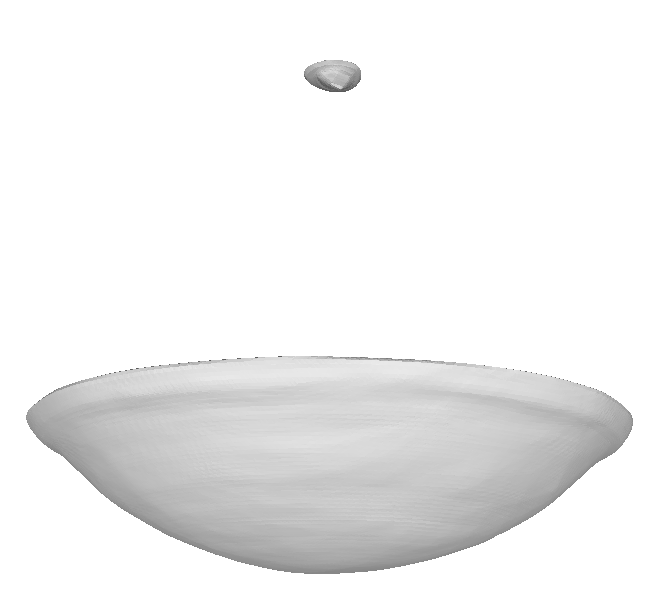} 
\end{subfigure}
\hspace{0.02\linewidth}
\begin{subfigure}{0.13\linewidth}
\includegraphics[width=\textwidth]{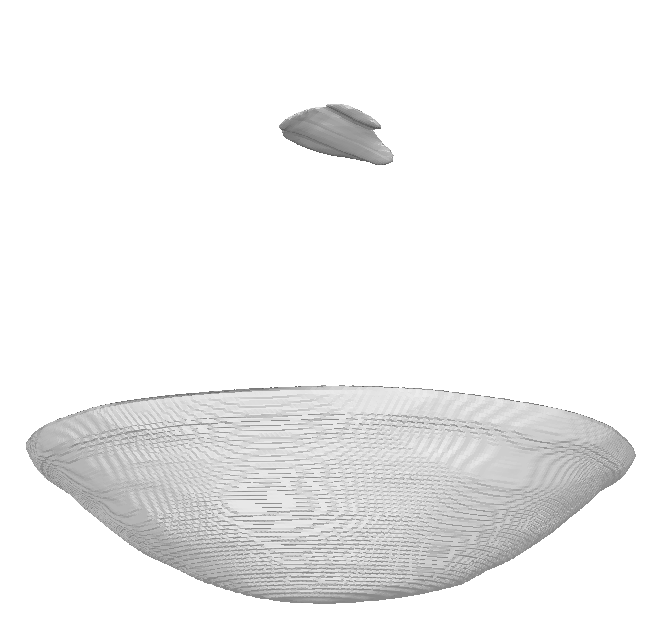} 
\end{subfigure}
\hspace{0.02\linewidth}
\begin{subfigure}{0.13\linewidth}
\includegraphics[width=\textwidth]{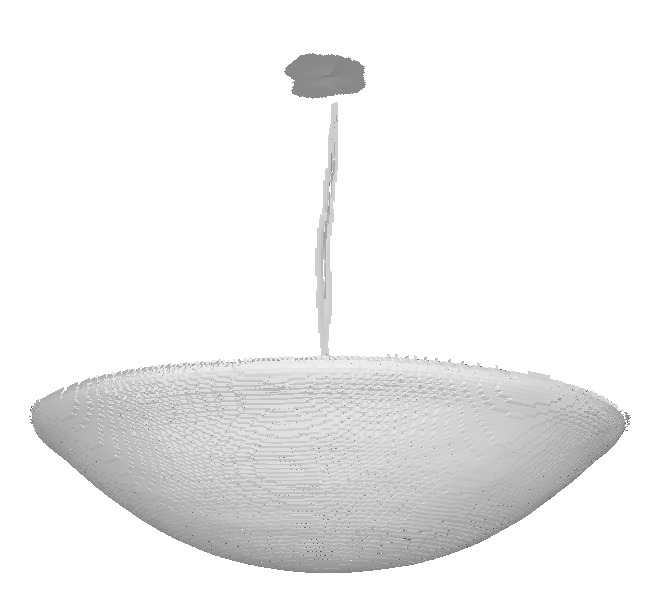} 
\end{subfigure} 
\begin{subfigure}{0.13\linewidth}
\includegraphics[width=\textwidth]{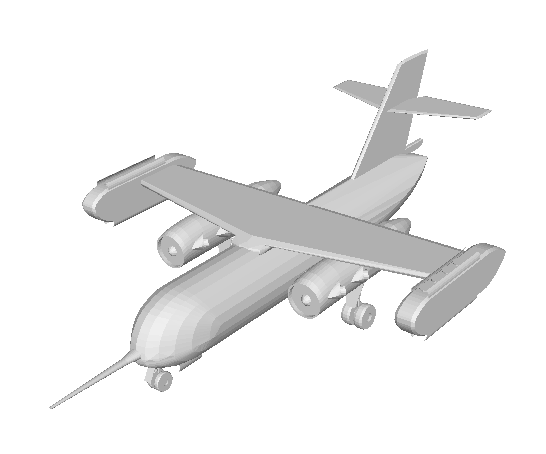}
\end{subfigure}
\hspace{0.02\linewidth}
\begin{subfigure}{0.13\linewidth}
\includegraphics[width=\textwidth]{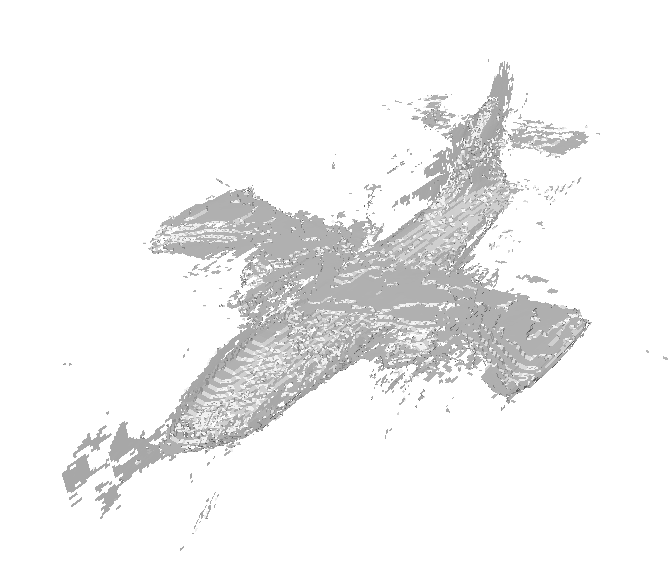}
\end{subfigure}
\hspace{0.02\linewidth}
\begin{subfigure}{0.13\linewidth}
\includegraphics[width=\textwidth]{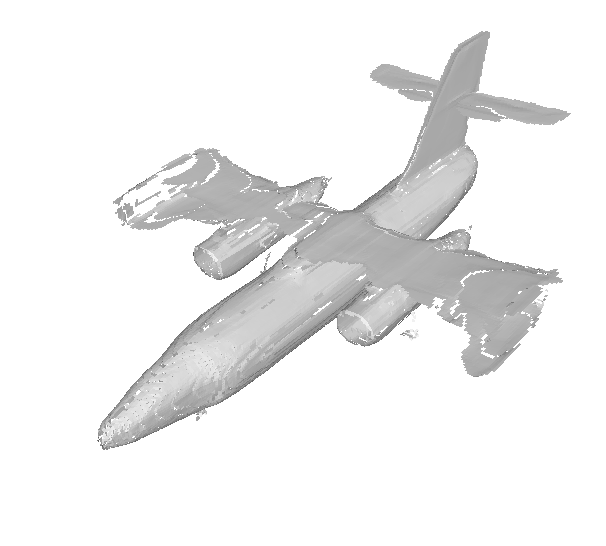} 
\end{subfigure}
\hspace{0.02\linewidth}
\begin{subfigure}{0.13\linewidth}
\includegraphics[width=\textwidth]{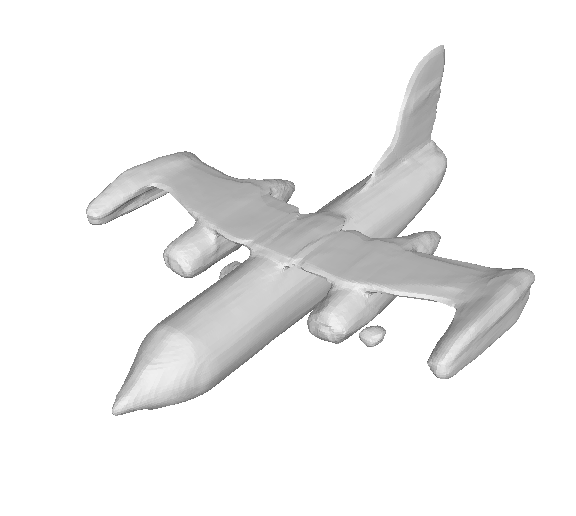} 
\end{subfigure}
\hspace{0.02\linewidth}
\begin{subfigure}{0.13\linewidth}
\includegraphics[width=\textwidth]{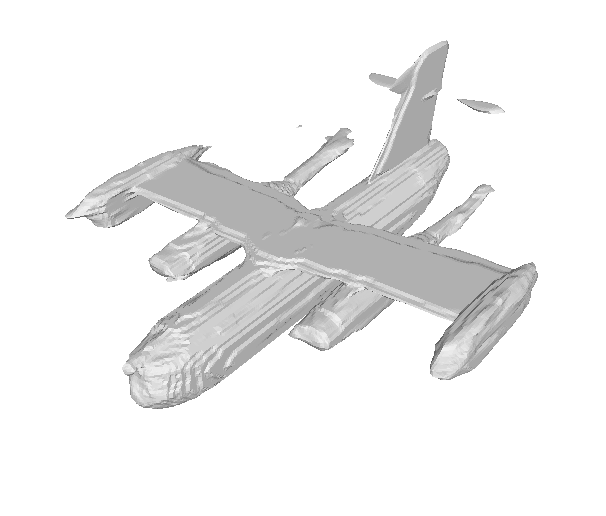} 
\end{subfigure}
\hspace{0.02\linewidth}
\begin{subfigure}{0.13\linewidth}
\includegraphics[width=\textwidth]{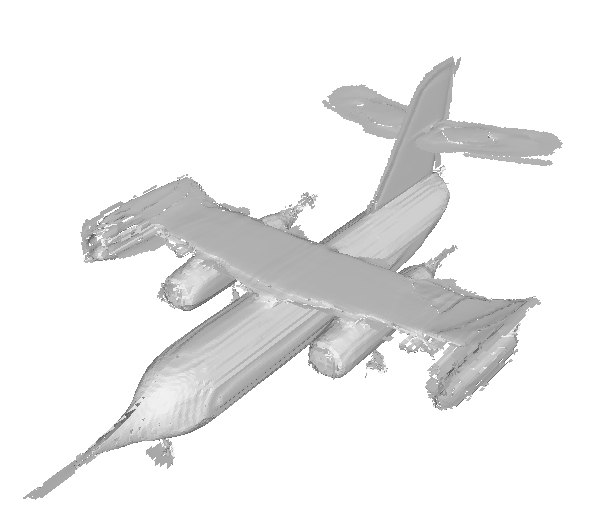} 
\end{subfigure} 
\begin{subfigure}{0.13\linewidth}
\includegraphics[width=\textwidth]{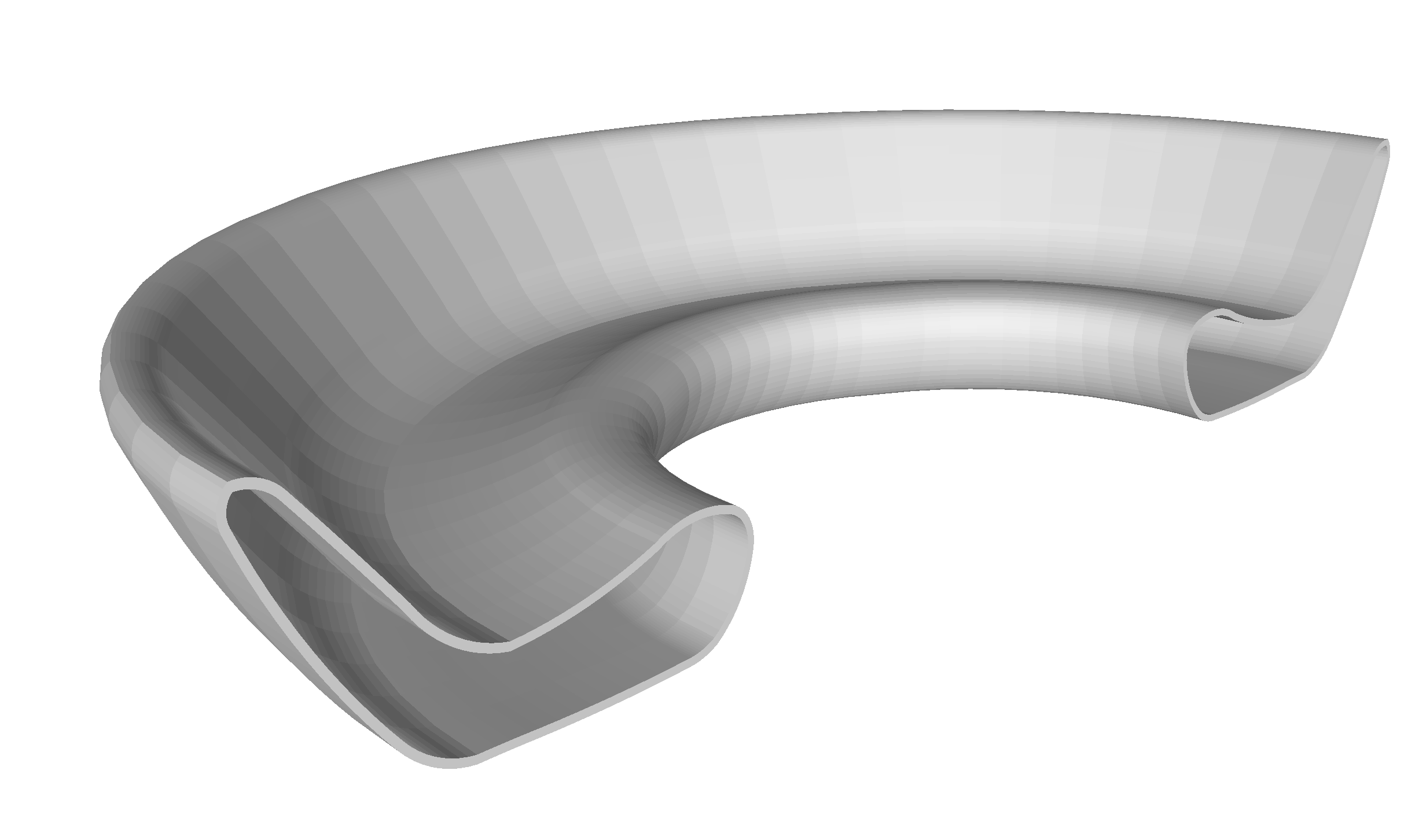}
\end{subfigure}
\hspace{0.02\linewidth}
\begin{subfigure}{0.13\linewidth}
\includegraphics[width=\textwidth]{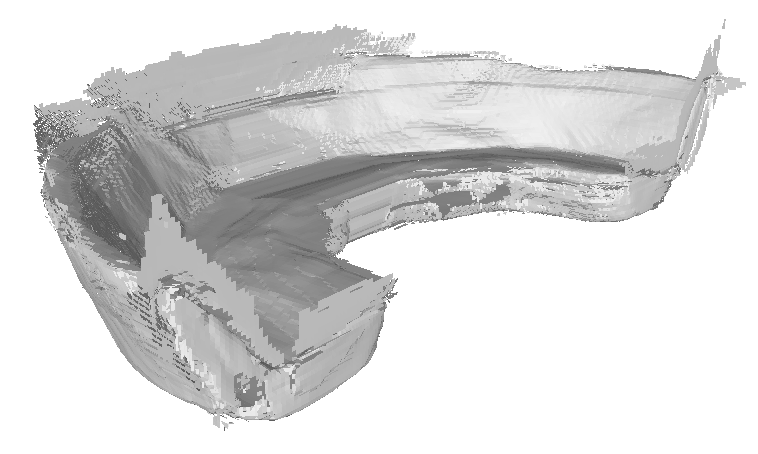}
\end{subfigure}
\hspace{0.02\linewidth}
\begin{subfigure}{0.13\linewidth}
\includegraphics[width=\textwidth]{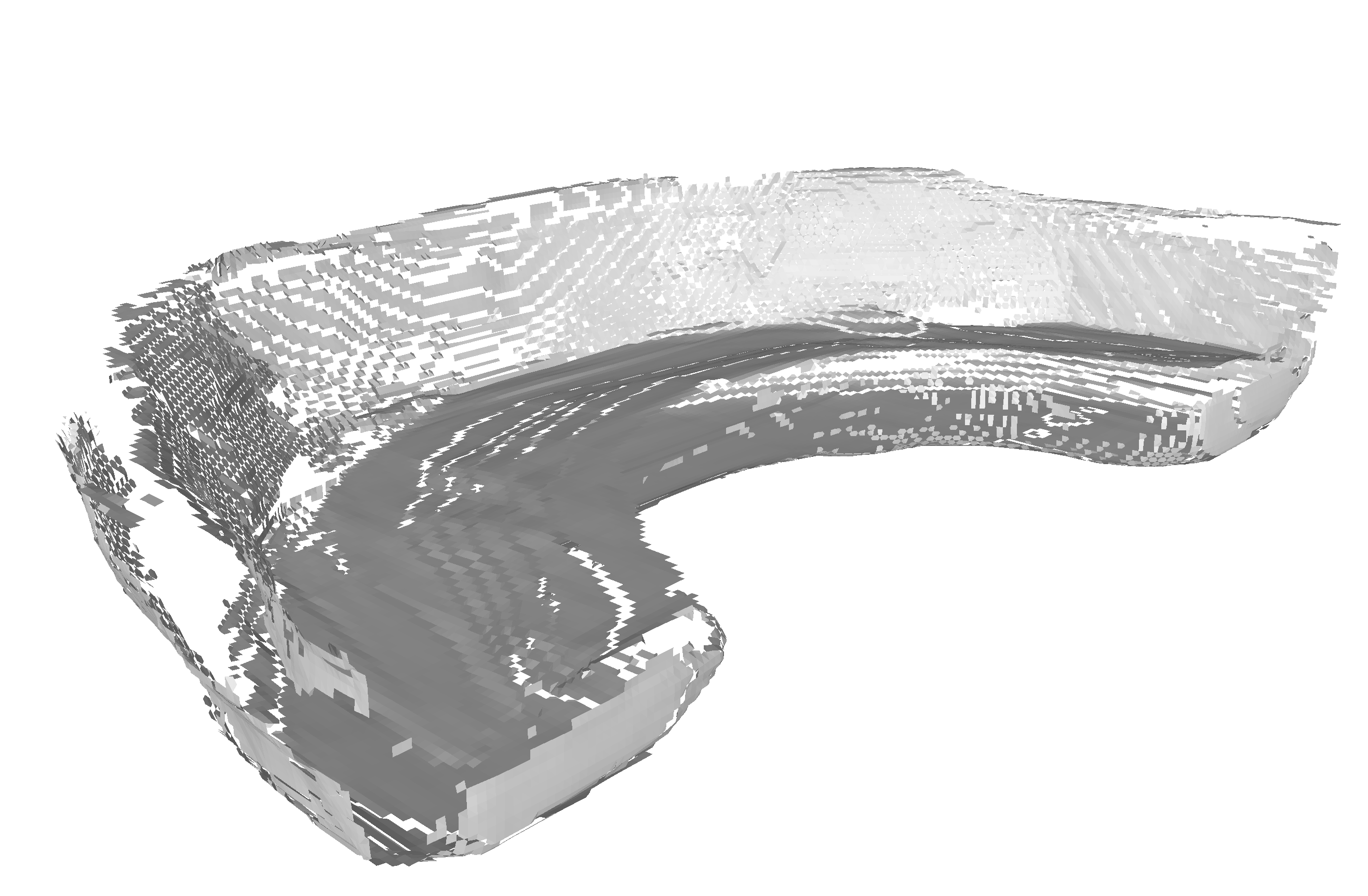} 
\end{subfigure}
\hspace{0.02\linewidth}
\begin{subfigure}{0.13\linewidth}
\includegraphics[width=\textwidth]{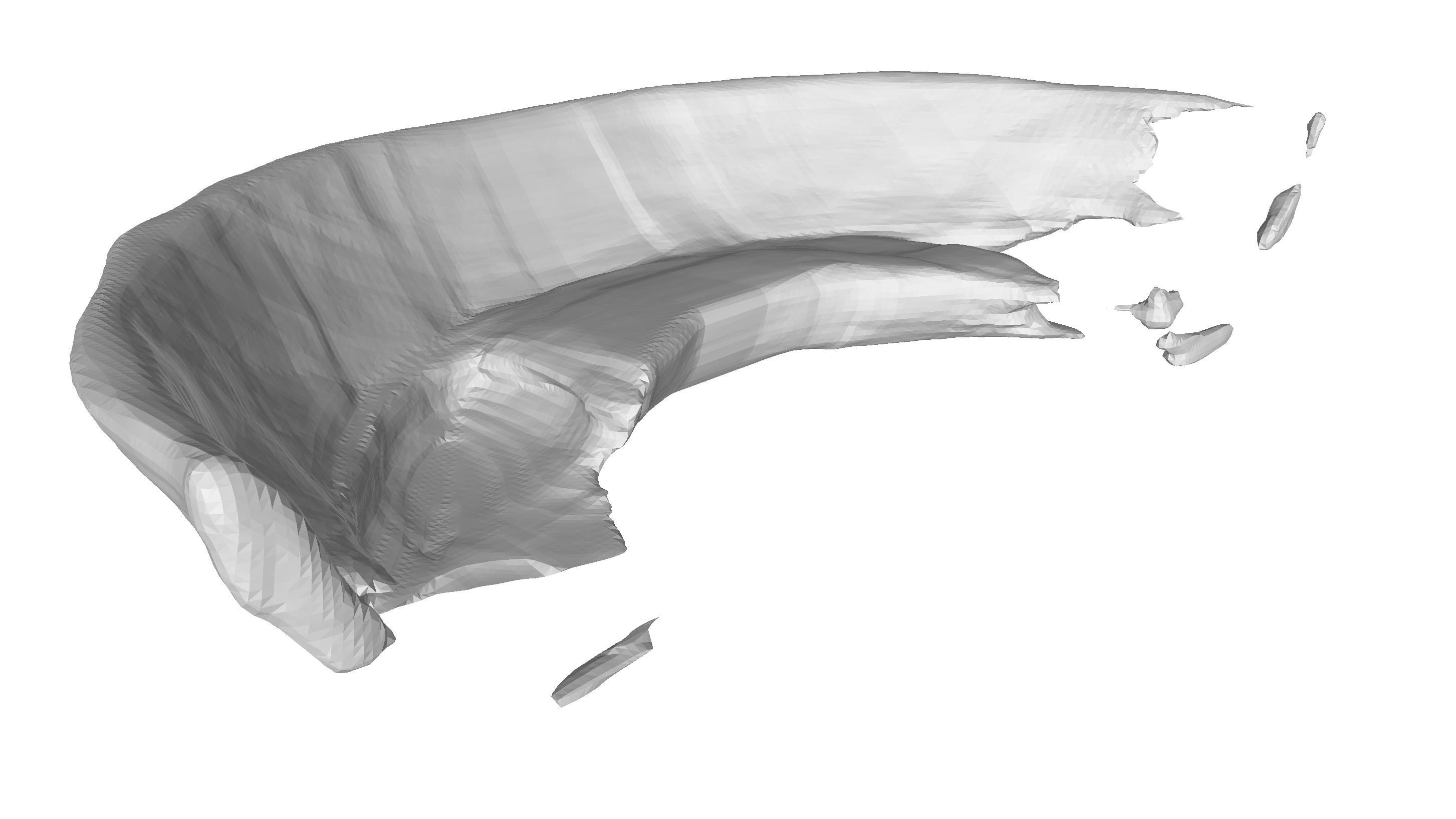} 
\end{subfigure}
\hspace{0.02\linewidth}
\begin{subfigure}{0.13\linewidth}
\includegraphics[width=\textwidth]{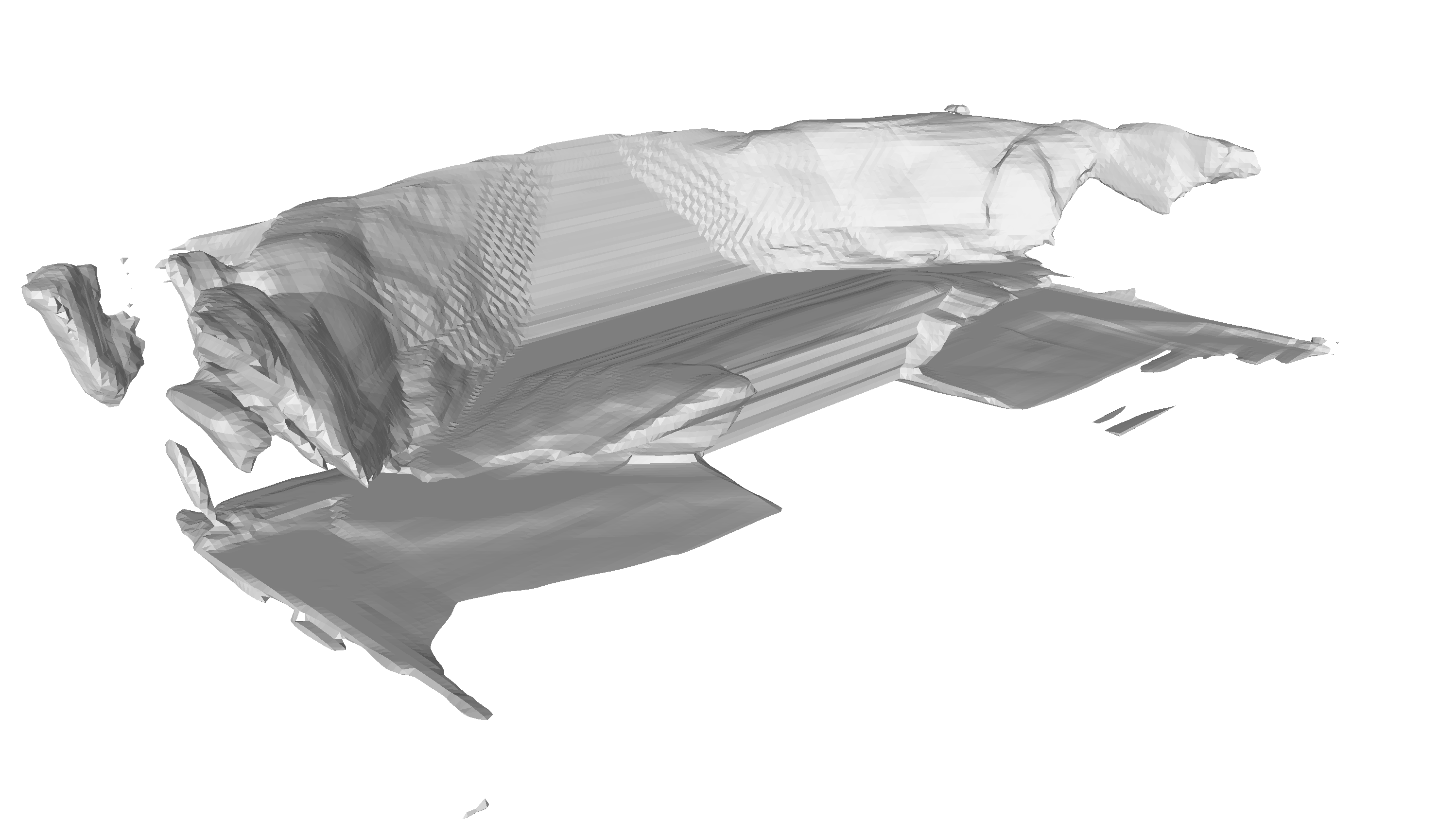} 
\end{subfigure}
\hspace{0.02\linewidth}
\begin{subfigure}{0.13\linewidth}
\includegraphics[width=\textwidth]{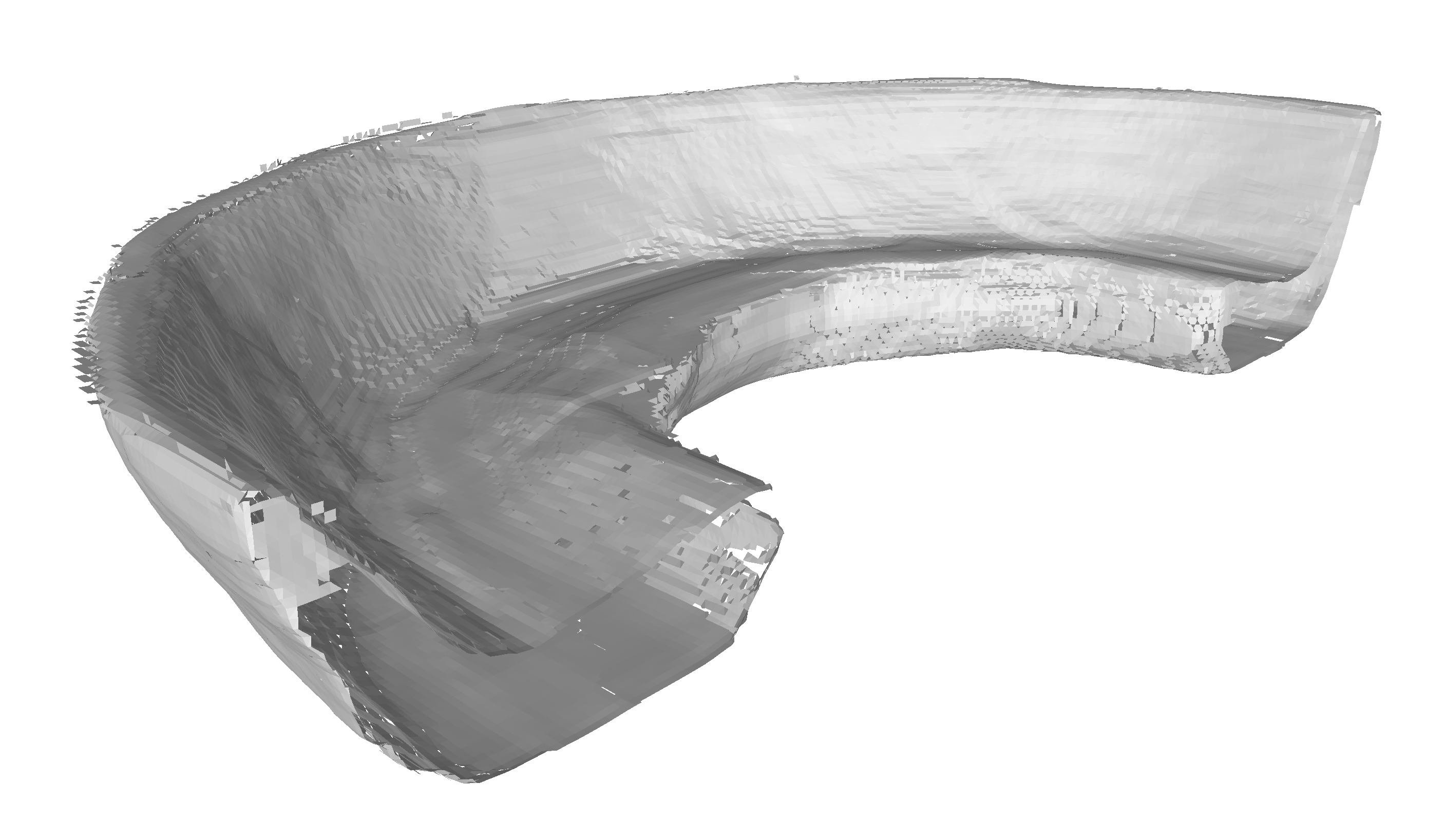} 
\end{subfigure} 
\begin{subfigure}{0.13\linewidth}
\includegraphics[width=\textwidth]{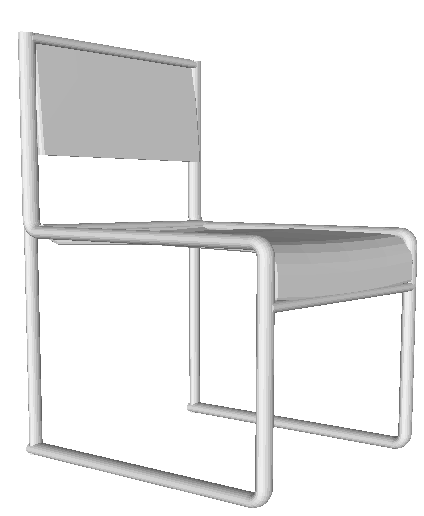}
\caption{GT}
\end{subfigure}
\hspace{0.02\linewidth}
\begin{subfigure}{0.13\linewidth}
\includegraphics[width=\textwidth]{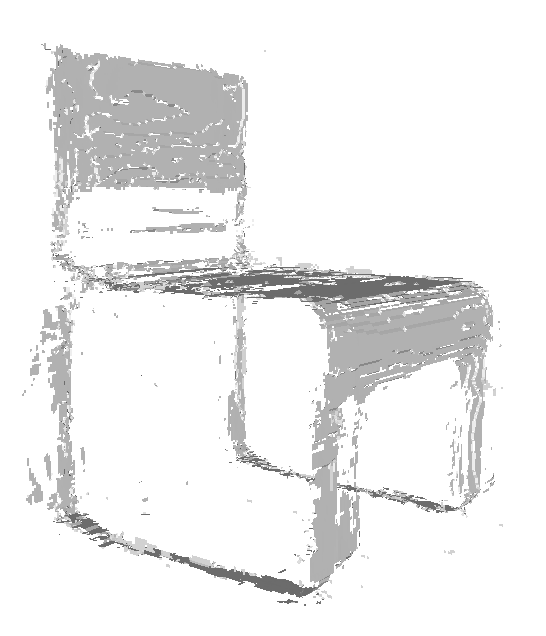}
\caption{NDF \cite{chibane2020ndf}}
\end{subfigure}
\hspace{0.02\linewidth}
\begin{subfigure}{0.13\linewidth}
\includegraphics[width=\textwidth]{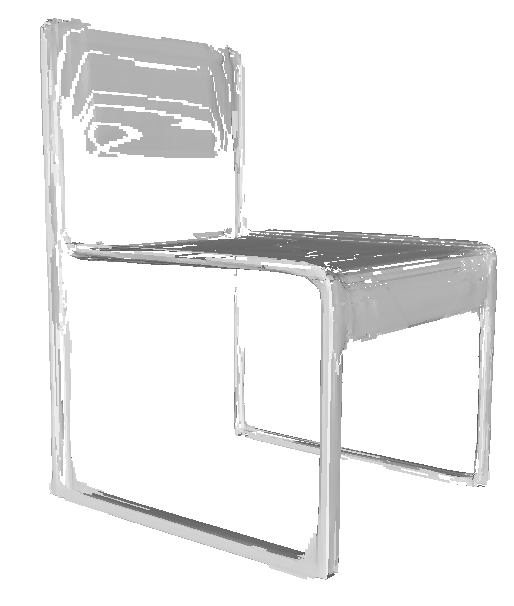} 
\caption{GIFS \cite{Ye_2022gifs}}
\end{subfigure}
\hspace{0.02\linewidth}
\begin{subfigure}{0.13\linewidth}
\includegraphics[width=\textwidth]{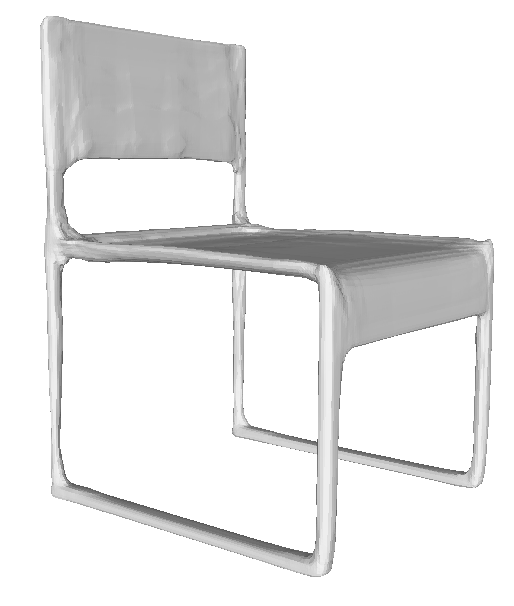} 
\caption{DeepSDF \cite{park2019deepsdf}}
\end{subfigure}
\hspace{0.02\linewidth}
\begin{subfigure}{0.13\linewidth}
\includegraphics[width=\textwidth]{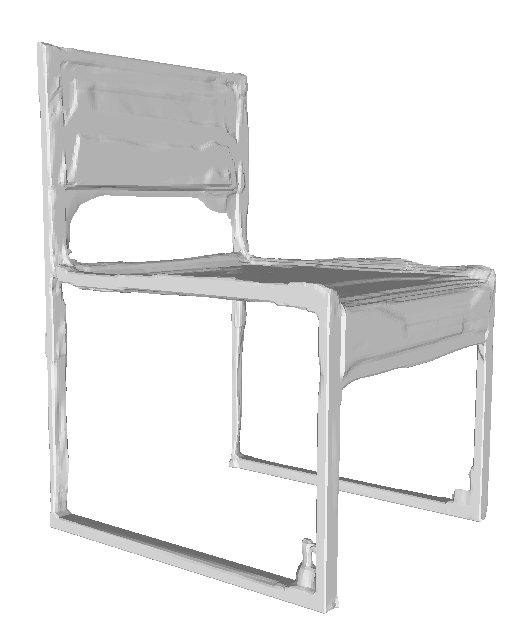} 
\caption{OccNet \cite{mescheder2019occnet}}
\end{subfigure}
\hspace{0.02\linewidth}
\begin{subfigure}{0.13\linewidth}
\includegraphics[width=\textwidth]{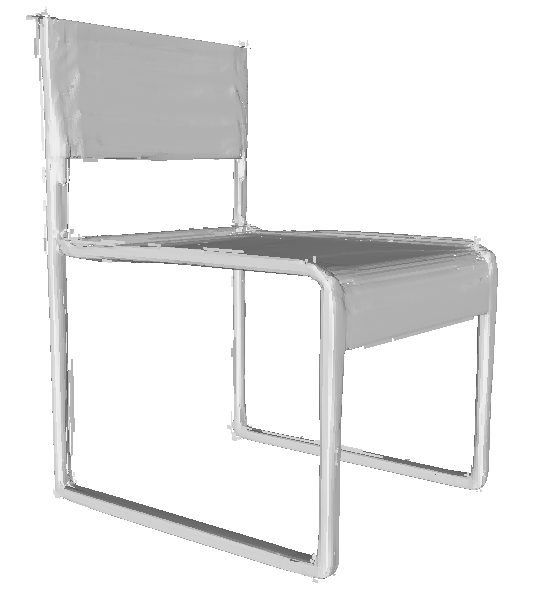}
\caption{VF}
\end{subfigure} 
\caption{\textbf{Qualitative reconstruction.} We compare each representation method in reconstructing shapes from each used ShapeNet \cite{cheng2015shapenet} class. VF performs significantly better than the other representations in any shape and on par with DeepSDF on the chair example. Regarding the qualitatively poor performance of DeepSDF \cite{park2019deepsdf} and OccNet \cite{mescheder2019occnet} in the \textit{sofas} class (fourth row), we point out that the shown example is a challenging shape as it is empty inside and thin (see GT shape), which does not allow it to be easily represented as watertight. Regarding the performance on \textit{cars}, it needs to be noted that the watertight shape includes the empty engine space visible from the front grid. For this reason, DeepSDF \cite{park2019deepsdf} and OccNet \cite{mescheder2019occnet}, struggle to reconstruct the thin bonnet accurately.}
\label{fig:qual_sup}
\end{figure*}

\begin{figure}[ht]
\centering
\begin{subfigure}{0.21\linewidth}
\includegraphics[width=\textwidth]{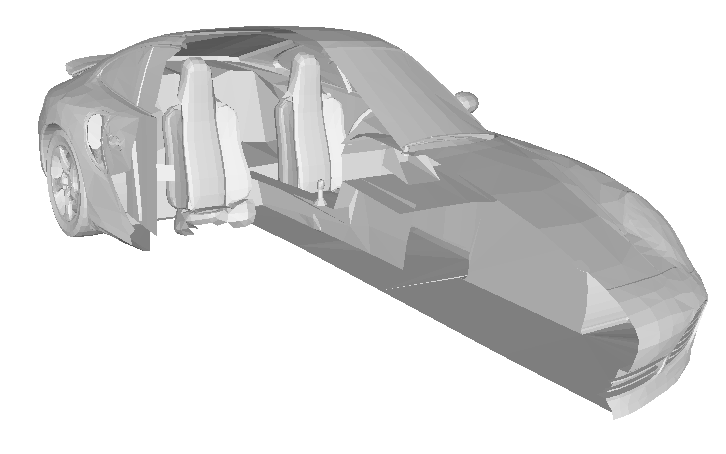}
\end{subfigure}
\hspace{0.02\linewidth}
\begin{subfigure}{0.21\linewidth}
\includegraphics[width=\textwidth]{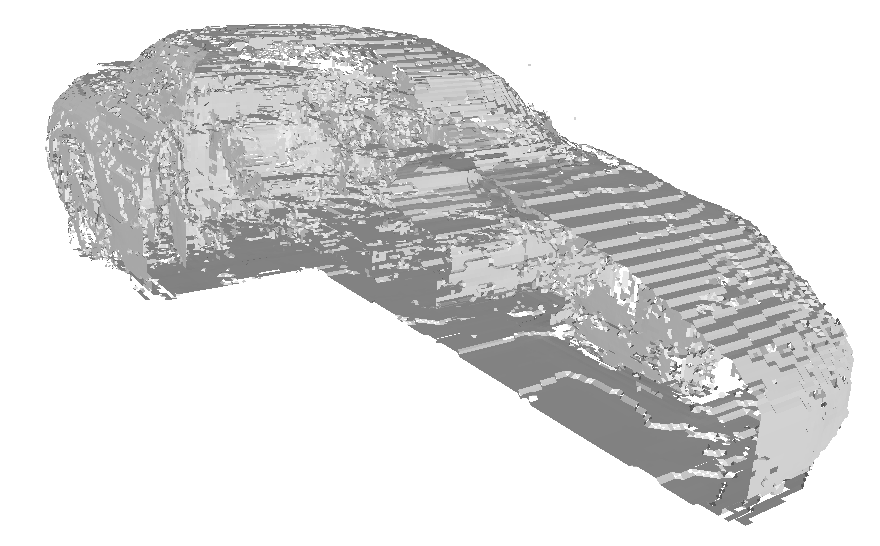}
\end{subfigure}
\hspace{0.02\linewidth}
\begin{subfigure}{0.21\linewidth}
\includegraphics[width=\textwidth]{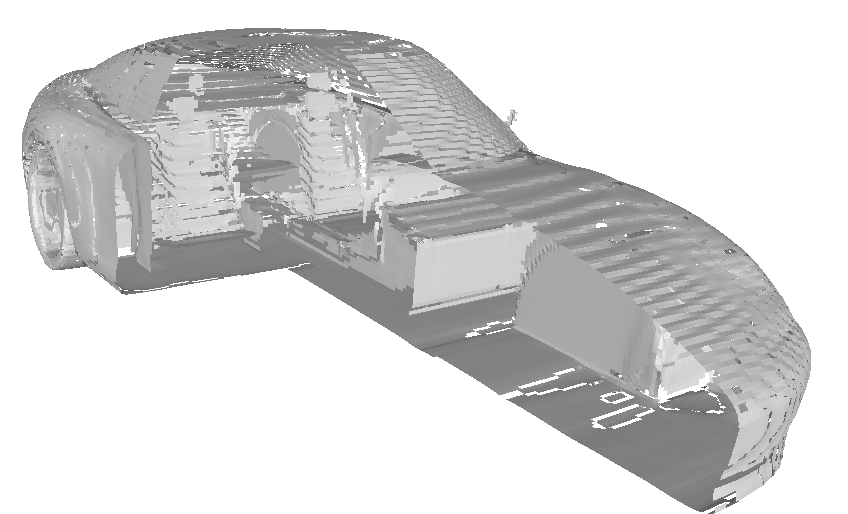} 
\end{subfigure}
\hspace{0.02\linewidth}
\begin{subfigure}{0.21\linewidth}
\includegraphics[width=\textwidth]{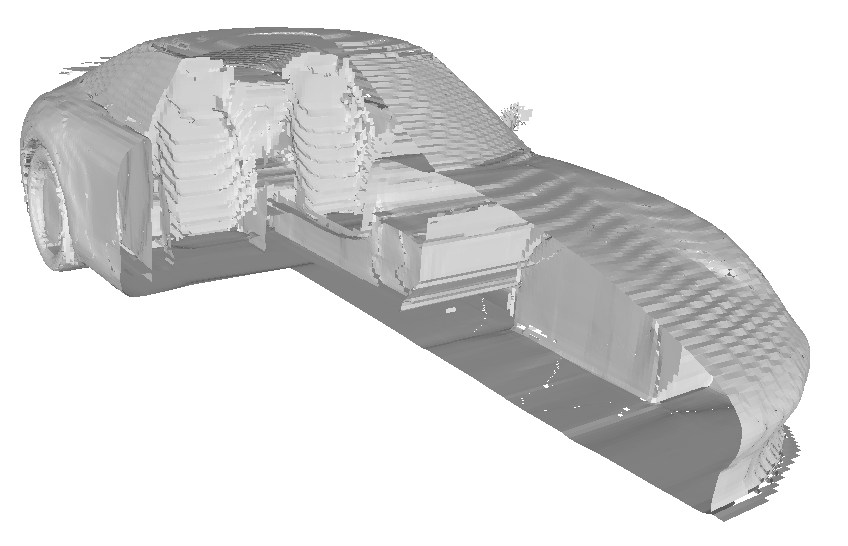} 
\end{subfigure} 
\begin{subfigure}{0.21\linewidth}
\includegraphics[width=\textwidth]{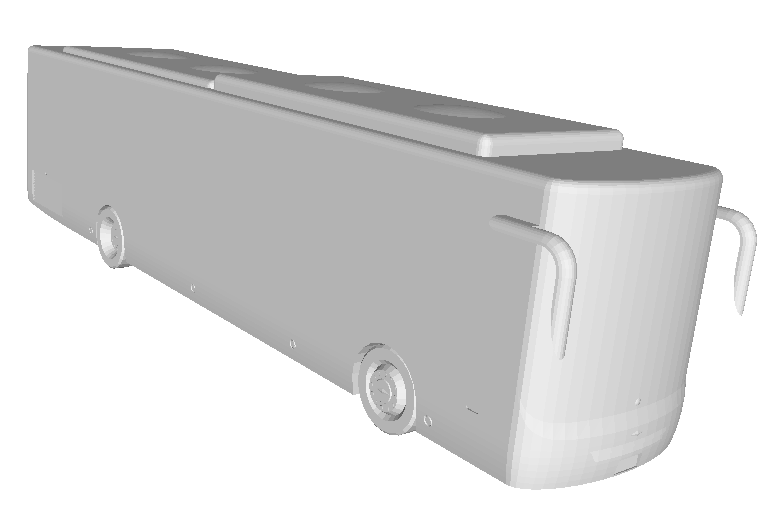}
\end{subfigure}
\hspace{0.02\linewidth}
\begin{subfigure}{0.21\linewidth}
\includegraphics[width=\textwidth]{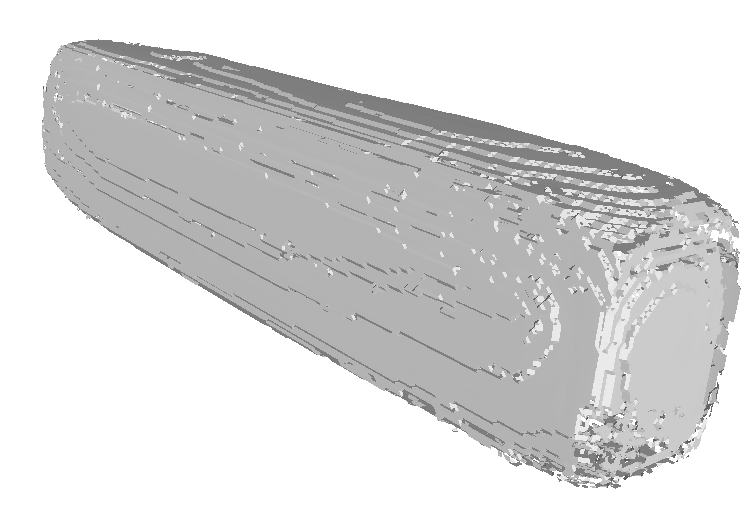}
\end{subfigure}
\hspace{0.02\linewidth}
\begin{subfigure}{0.21\linewidth}
\includegraphics[width=\textwidth]{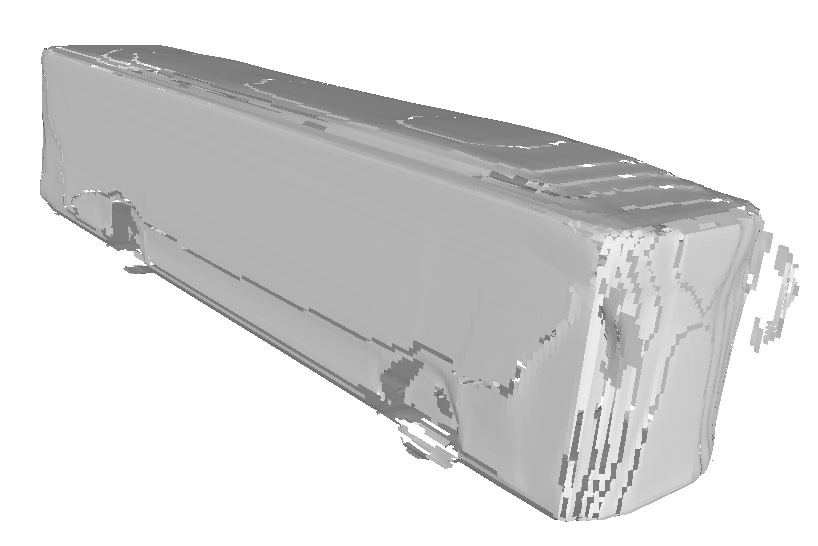} 
\end{subfigure}
\hspace{0.02\linewidth}
\begin{subfigure}{0.21\linewidth}
\includegraphics[width=\textwidth]{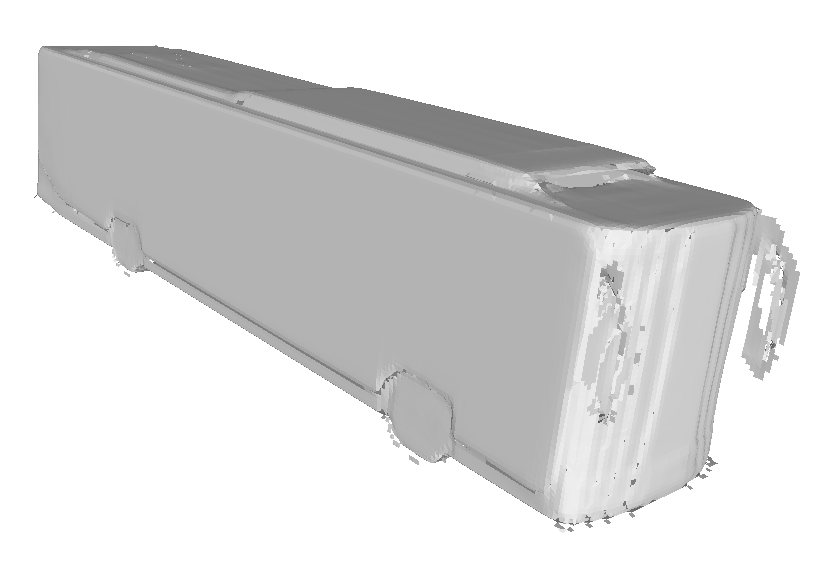} 
\end{subfigure} 
\begin{subfigure}{0.17\linewidth}
\includegraphics[width=\textwidth]{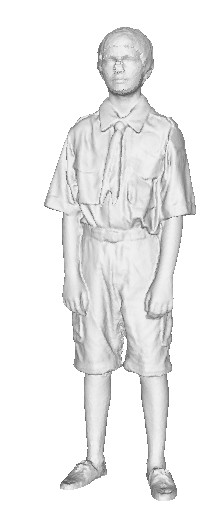}
\caption{GT}
\end{subfigure}
\hspace{0.06\linewidth}
\begin{subfigure}{0.19\linewidth}
\includegraphics[width=\textwidth]{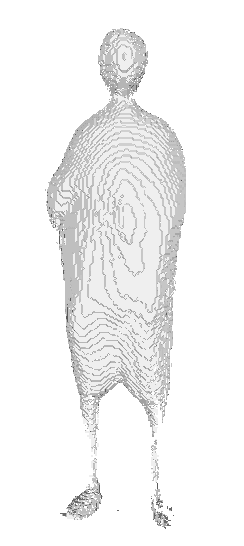}
\caption{NDF \cite{chibane2020ndf}}
\end{subfigure}
\hspace{0.06\linewidth}
\begin{subfigure}{0.19\linewidth}
\includegraphics[width=\textwidth]{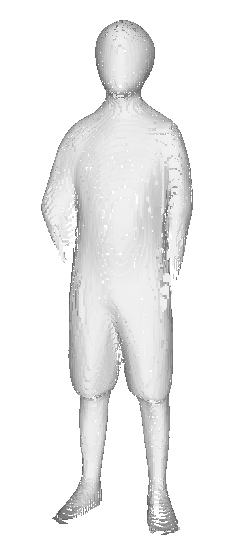} 
\caption{GIFS \cite{Ye_2022gifs}}
\end{subfigure}
\hspace{0.06\linewidth}
\begin{subfigure}{0.16\linewidth}
\includegraphics[width=\textwidth]{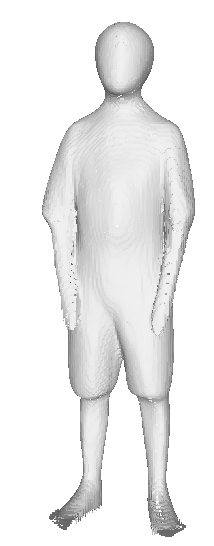}
\caption{VF}
\end{subfigure} 
\caption{\textbf{General shape representation}: VF can represent both open and multi-layered shapes. VF can outperform the other representations across every shape class. We note VF reconstructs with accuracy even the inside of the shown car.}
\label{fig:sup_open_qual}
\end{figure}

\section{Qualitative Results Analysis}

Here we provide additional visualizations of the results obtained with VF and compare to the other representations.
Figure \ref{fig:good_qual} shows the results achieved by VF on some challenging open and multi-layered examples to further demonstrate the ability of VF to generate accurate predictions in every shape class. In particular it is worth noting the accuracy in reconstructing the very thin parts in the lamp structure and the inside of the car and the mini-bus.

In Figure \ref{fig:qual_sup}, we provide additional qualitative comparison to the other existing methods. Similarly to the results shown previously, this highlights the different properties of the methods. As volumetric representations, like DeepSDF \cite{park2019deepsdf} and OccNet \cite{mescheder2019occnet}, predict smooth watertight meshes, they achieve high-quality results in smooth classes such as \textit{chairs}. On the other hand, this property shows to be a limitation when representing thin object parts, such as in the \textit{sofas} or \textit{planes} example. Overall, they also struggle in predicting sharp details as highlighted throughout every class.

More example comparison between VF, NDF \cite{chibane2020ndf}, and GIFS \cite{Ye_2022gifs} can be seen in Figure \ref{fig:sup_open_qual}. Here again, we can notice the similar expressive power between methods, with VF that can outperform the other representations at the detail level and with sharper results. In particular, NDF \cite{chibane2020ndf} is significantly more affected by noise compared to VF, which leads to small holes and artifacts. As shown in the main paper and in Section \ref{sec:sup_qual}, a probable cause being the higher level of noise observed in the UDF\footnote{The representation used by NDF \cite{chibane2020ndf}} and its gradient. GIFS \cite{Ye_2022gifs}, on the other hand, is less affected by noise but is still less effective in representing details. This might be caused by the higher complexity of having 2 separate predictions and of learning a function that relates point pairs. Furthermore, it does not allow to easily model surface properties, such as planarity, restricting its possible future applications.

\section{Implementation Details and Ablation}

\begin{figure}[ht]
\centering
\begin{subfigure}{0.99\linewidth}
\includegraphics[width=\textwidth]{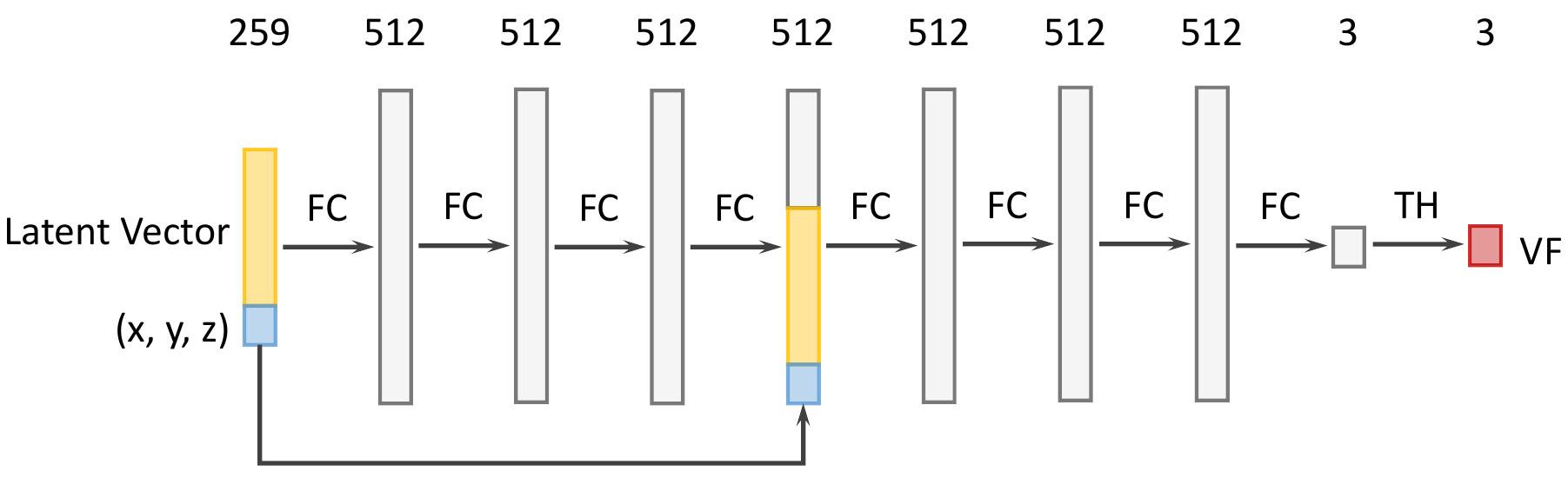} 
\caption{VF Network}
\label{fig:vf_net}
\end{subfigure}
\begin{subfigure}{0.99\linewidth}
\includegraphics[width=\textwidth]{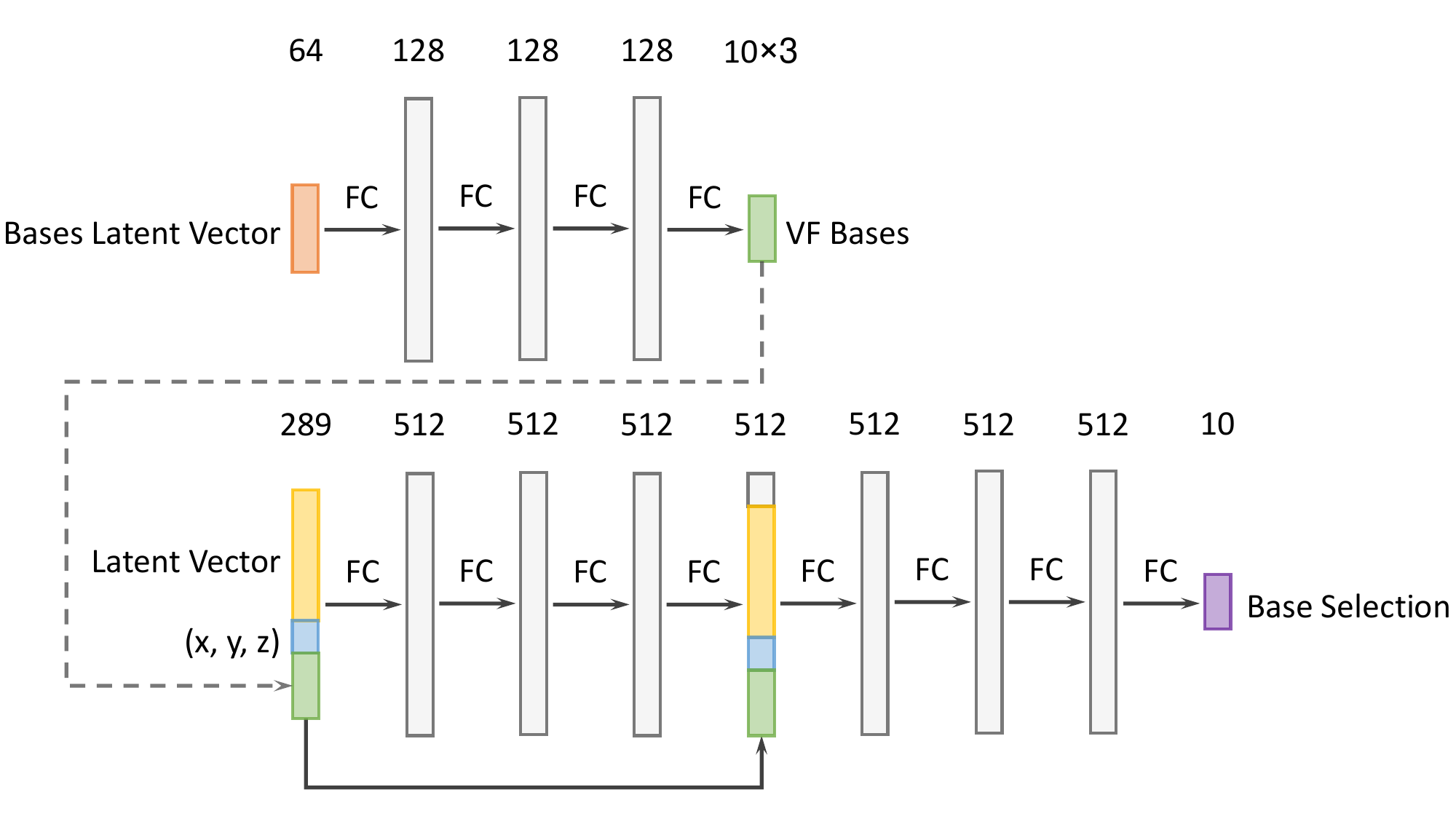}
\caption{Planar VF Network}
\label{fig:plan_vf_net}
\end{subfigure} 
\caption{Auto-decoder network structure as used for VF and Planar VF. FC indicates a Fully Connected layer and TH the hyperbolic tangent activation}
\label{fig:net_strcuture}
\end{figure}

In this section, we show the specifics of the network structure used for VF and Planar VF and its training. We further ablate the network structure showing that VF can benefit from larger architectures, such as the one proposed by \cite{chibane2020ndf} on the \textit{cars} ShapeNet \cite{cheng2015shapenet} category.

As described in the main text, the architecture structure used for the experiments with VF and every other method is an auto-decoder network~\cite{park2019deepsdf} The specifics of its number of layers and size can be seen in Figure \ref{fig:vf_net}. It is trained for $2\,000$ epochs with samples from $64$ scenes in each batch and $16\,384$ points per scene. The optimizer used is Adam, with a starting learning rate of $0.001$ which is decreased by a factor of $2$ every $500$ epochs. The same optimizer and scheduler are used also for the latent vector with a starting learning rate of $0.0005$. In addition to the $\ell_1$ loss used to train VF, a $\ell_2$ regularization loss is applied to the latent vector to keep its norm small. Weight normalization and $0.2$ dropout are applied to each network layer, as well as the ReLU activation except for the output layer.

During inference, only the latent vector is optimized for each shape. It is initialized as a zero-mean Gaussian random vector with $0.01$ standard deviation, and then optimized for $800$ iterations with initial learning rate of $0.005$, decreased by a factor of $10$ after $400$ iterations. In each iteration, $800$ randomly sampled points are selected for the optimization. In order to obtain the reconstructed mesh, the auto-decoder is queried using each position on a voxel grid, together with the optimized latent vector. The norm of VF is then taken and used as input to the MC algorithm, while the normalized VF is used to compute the discrete flux density.

Planar VF uses a similar architecture as shown in Figure \ref{fig:plan_vf_net}. It is trained and optimized in an equivalent manner as explained for VF with the same hyperparameters highlighted for the network and latent vector optimization. As an important training detail, we note that the basis vectors $\mathsf{b}$ are first shuffled in their order before being fed to the main network branch. This allows the network to learn to select the appropriate basis and not overfit on the most common set of bases.

Table \ref{tab:ablation} shows the results obtained using the larger network proposed in \cite{chibane2020ndf}. It is significantly larger compared to the standard auto-decoder architecture commonly used for the task, and requires significantly longer training - $2 \times$ increase in both memory and training time. From Table \ref{tab:ablation}, it can be seen that VF benefits from the larger network, and keeps an edge over the compared methods in most metrics.

\begin{table}[!h]
    \centering
    \resizebox{0.7\linewidth}{!}{
    \begin{tabular}{ c | c c | c c }\toprule
         \multirow{2}{*}{\textbf{Method}} & \multicolumn{2}{c|}{Chamfer} & \multicolumn{2}{c}{F1-Score} \\
         & mean & median & 0.005 & 0.01 \\ \midrule
         NDF \cite{chibane2020ndf} & 0.0126 & 0.0120 & 88.09 & \textbf{99.54} \\
         GIFS \cite{Ye_2022gifs} & 0.0128 & 0.0123 & 88.05 & 99.31 \\ \midrule
         \textbf{VF} & \textbf{0.0122} & \textbf{0.0116} & \textbf{88.23} & 99.47 \\ \bottomrule
    \end{tabular}}
    \caption{\emph{Network ablation}: Reconstruction results on unprocessed ShapeNet~\cite{cheng2015shapenet} cars with the larger backbone used in \cite{chibane2020ndf} \cite{Ye_2022gifs}.}
    \label{tab:ablation}
\end{table}

\section{Shape Representation Properties}
\label{sec:sup_qual}

\begin{figure}[ht]
\centering
\begin{subfigure}{0.32\linewidth}
\includegraphics[width=\textwidth]{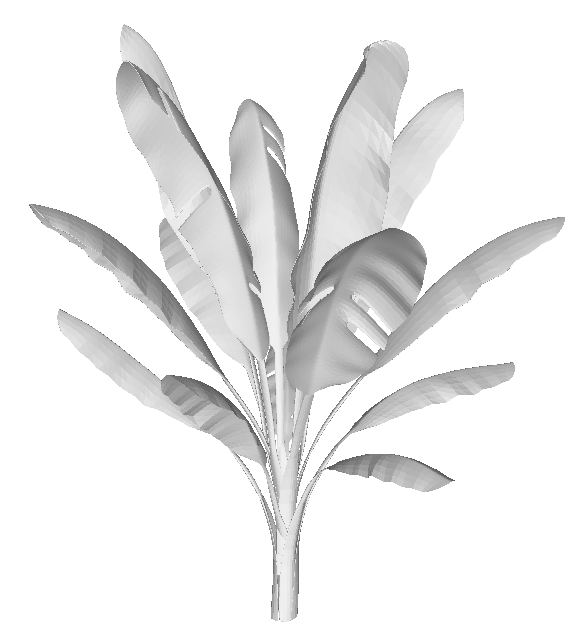}
\end{subfigure}
\begin{subfigure}{0.32\linewidth}
\includegraphics[width=\textwidth]{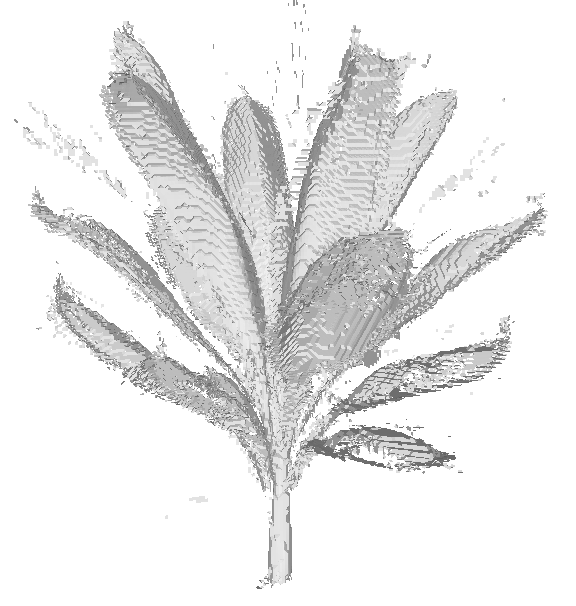}
\end{subfigure}
\begin{subfigure}{0.32\linewidth}
\includegraphics[width=\textwidth]{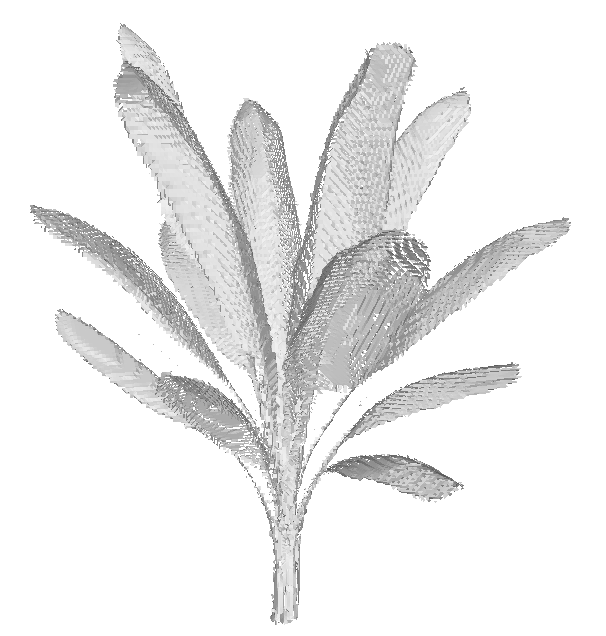}
\end{subfigure}
\begin{subfigure}{0.32\linewidth}
\includegraphics[width=\textwidth]{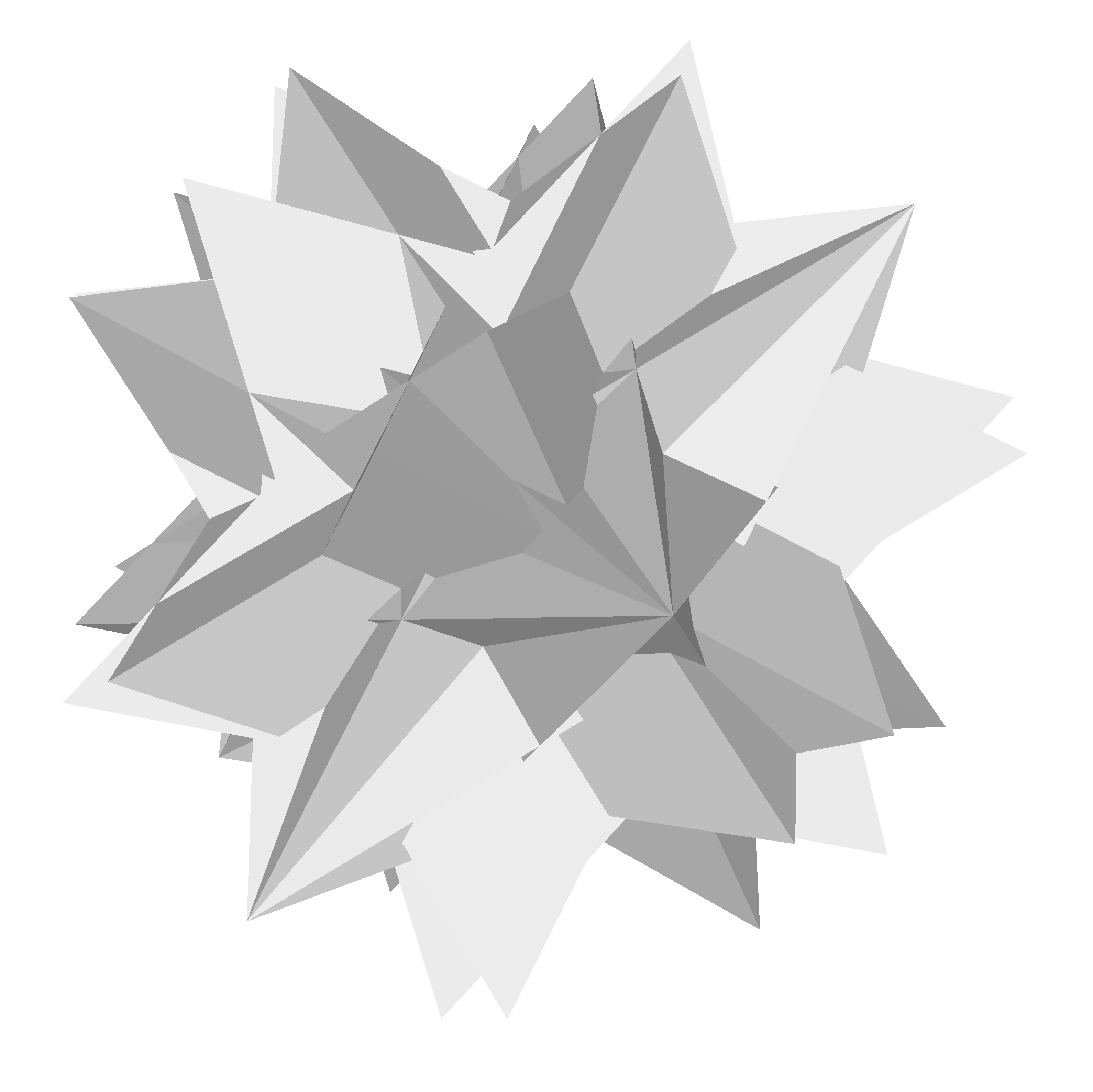} 
\caption{GT}
\end{subfigure}
\begin{subfigure}{0.32\linewidth}
\includegraphics[width=\textwidth]{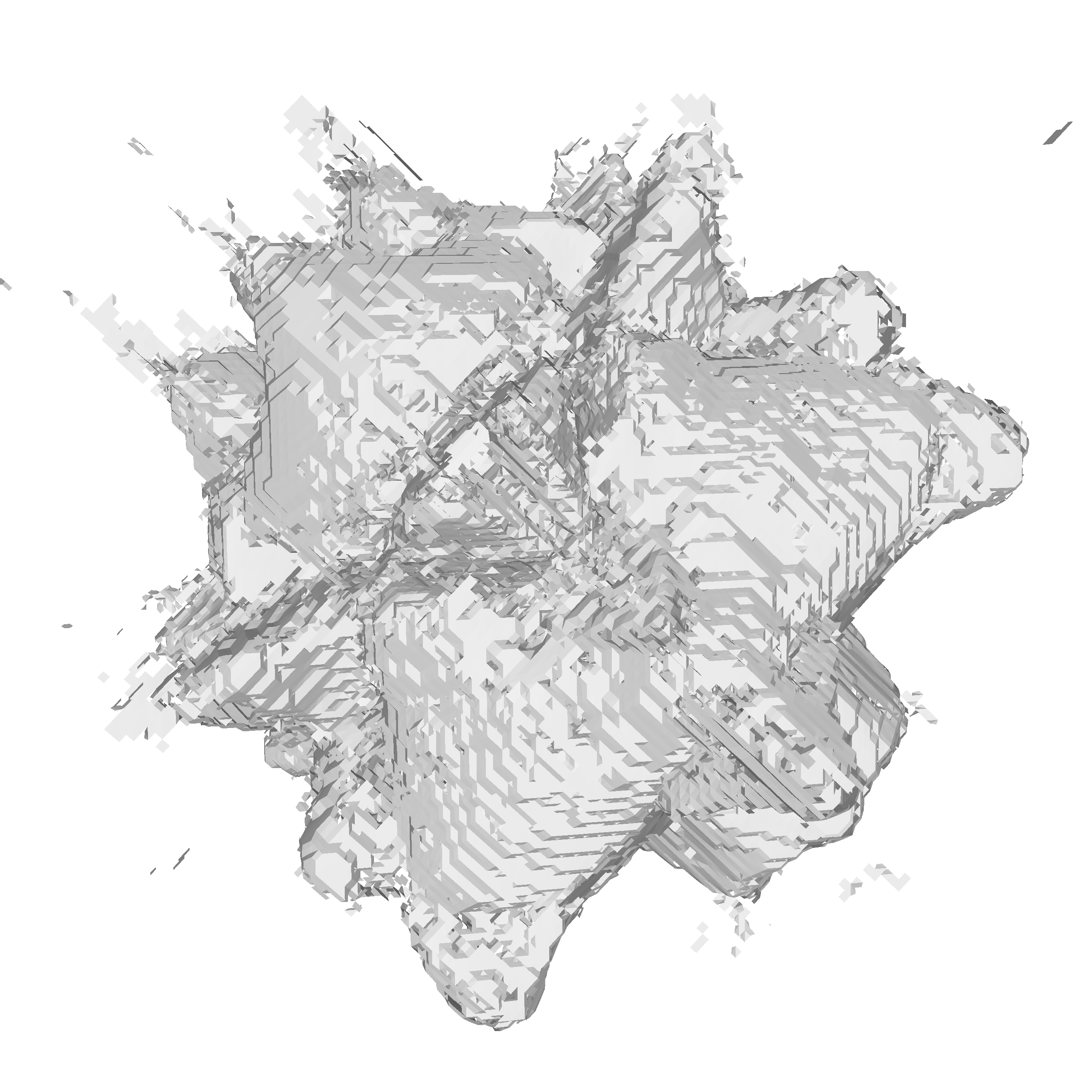}
\caption{UDF}
\end{subfigure} 
\begin{subfigure}{0.32\linewidth}
\includegraphics[width=\textwidth]{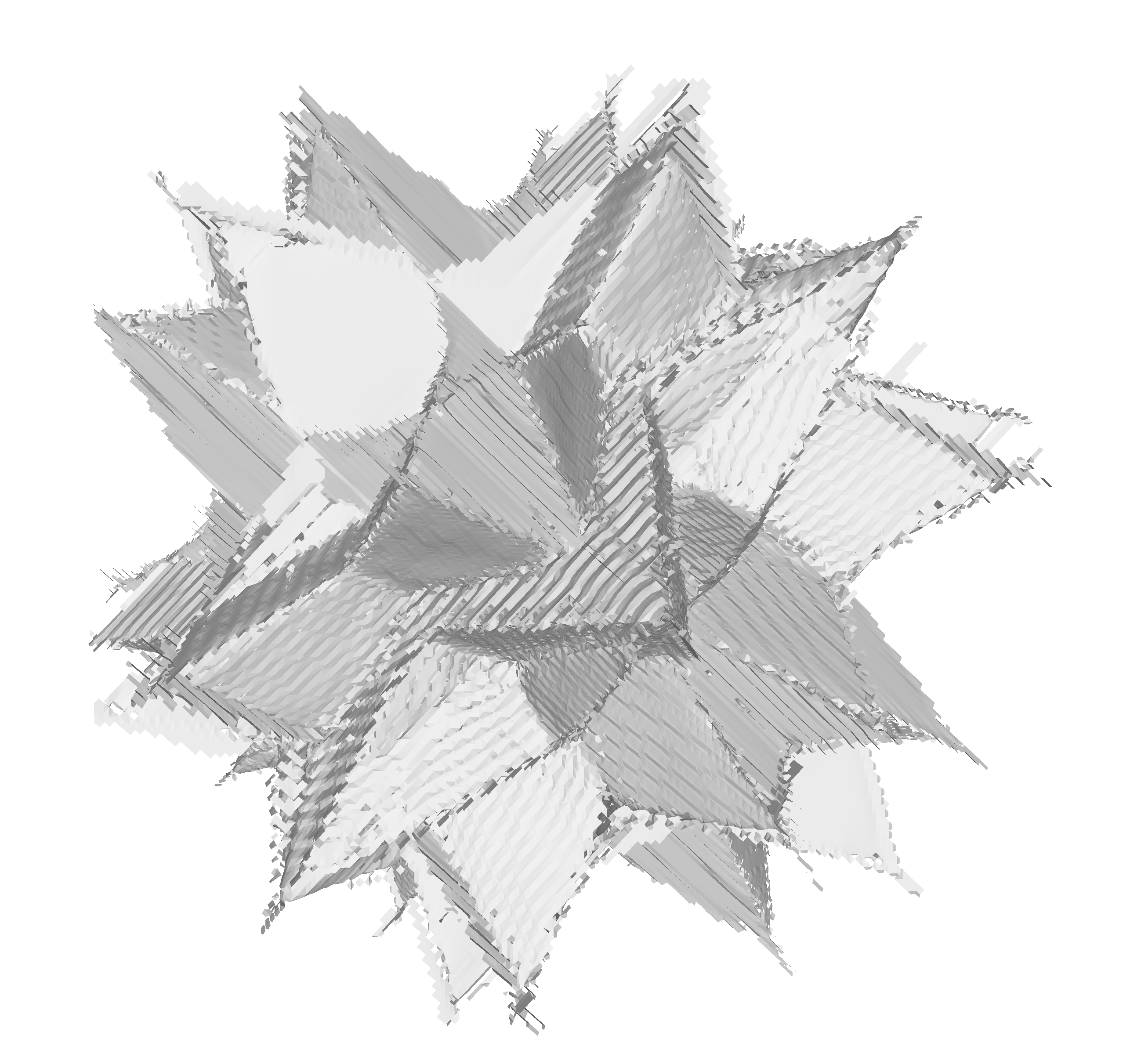}
\caption{VF}
\end{subfigure} 
\caption{Stress test on a tetrahedron-based complex shape and on a palm tree model. The tetrahedron-based shape is obtained by taking a tetrahedron and its dual (oriented with top and bottom on the $z$ axis) and replicating the shape by rotating of $60^{\circ}$ around the $x$ axis multiple times until returning to the original position. VF strongly outperforms UDF being able to effectively grasp the complexity of the shapes and avoiding noise and artifacts.}
\label{fig:complex_tetra}
\end{figure}

In this section, we show experiments on simple shapes as in Section 3.3 to support the validity of VF and compare against UDF. To highlight the differences, we try to reconstruct with a small network two tetrahedron-based shapes and analyze the accuracy and noise levels on them. The first toy experiment uses a tetrahedron overlapped with its dual shape; the second, which can be considered a stress test of the representation's capabilities, consists of trying to reconstruct a more complex shape, obtained by overlapping multiple rotated tetrahedrons (as explained in Fig. \ref{fig:complex_tetra}).

The network used, takes as input the 3D coordinates, has $5$ fully connected (FC) layers with a hidden size of $512$, and outputs the VF or UDF prediction.

\begin{figure}[ht]
\centering
\begin{subfigure}{0.32\linewidth}
\includegraphics[width=\textwidth]{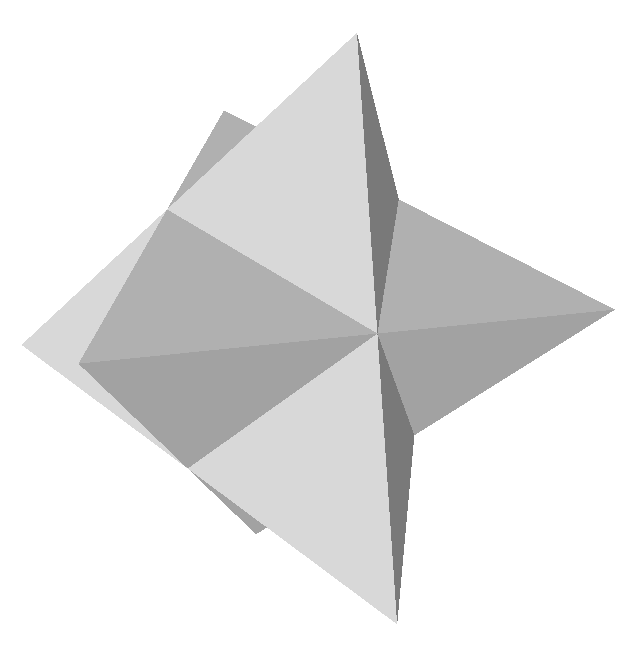} 
\caption{GT}
\end{subfigure}
\begin{subfigure}{0.32\linewidth}
\includegraphics[width=\textwidth]{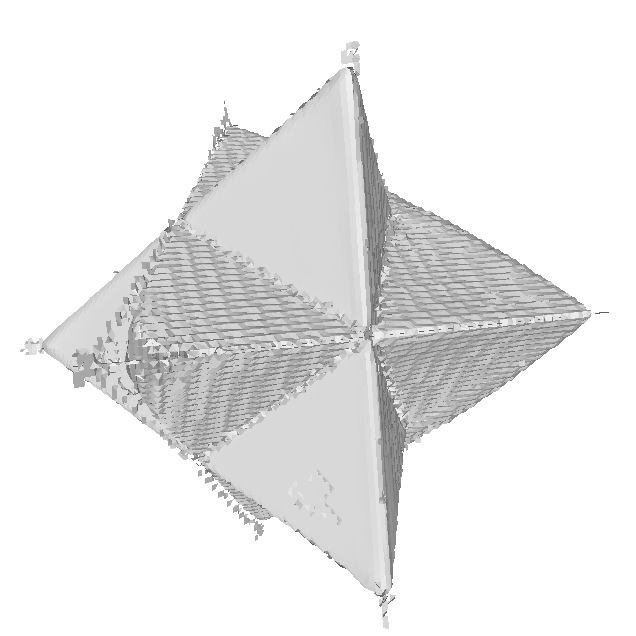}
\caption{UDF}
\end{subfigure} 
\begin{subfigure}{0.32\linewidth}
\includegraphics[width=\textwidth]{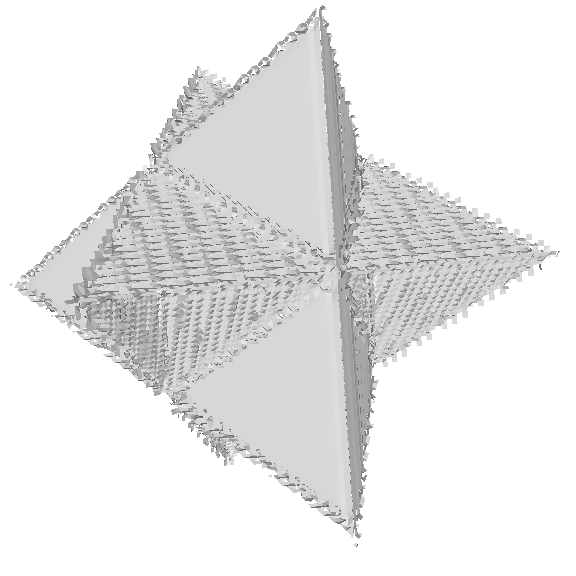}
\caption{VF}
\end{subfigure} 
\begin{subfigure}{0.8\linewidth}
\includegraphics[width=\textwidth]{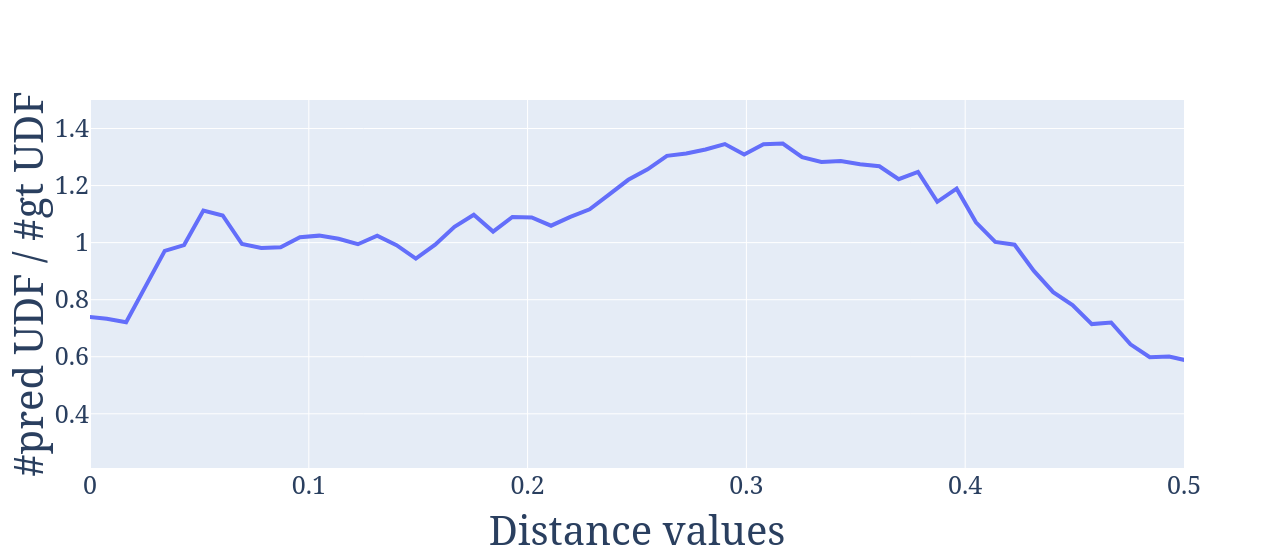}
\caption{UDF Bias Analysis}
\label{fig:bias_plot}
\end{subfigure} 
\caption{Representation comparison on a double tetrahedron shape. Top row: Reconstruction results; VF better preserves planar sides compared to UDF and does not have holes. Bottom row: Effect of UDF bias on the prediction. It shows the number of UDF predictions in a small range divided by the number of gt UDF in the same range. Values above $1$ at a certain distance implies that the number of predictions around that value is higher than the actual number of points at that distance; the opposite is true for values below $1$. For a perfect UDF surface extraction, one desires the value at distance 0 to be 1 (necessary condition). Instead the value is much lower indicating an unknown bias.
}
\label{fig:simple_tetra}
\end{figure}

Figure \ref{fig:simple_tetra} shows the performance on the 2 overlapped tetrahedrons. We observe that on this slightly more complex shape, UDF struggles significantly more than VF in preserving all the flat surfaces and the edges accurately. This is reflected on the accuracy in predicting the correct directions towards the surface. VF achieves an average $\ell_1$ error of $0.1831$, while the UDF gradients performs almost 3 times worse with an error of $0.4862$. Figure \ref{fig:bias_plot} shows the effect of the bias towards the mid range of the predicted values that affects UDF, as noted in the main text. It is evident that the model predicts values in the middle of the range much more often than they actually appear. Instead, the number of predictions of values at the extremes of the range is significantly smaller than in the ground truth. This reduced capability to predict accurately, is a further reason for the difference in performance between VF and UDF.

Figure \ref{fig:complex_tetra} shows the stress test on a complex toy shape. Qualitatively we can see a significant difference in the representation ability of VF when compared to UDF. This is reflected in the noise value, with the average $\ell_1$ error of VF at $0.3085$, much lower than the one of the UDF gradient at $0.7952$. Regarding Figure \ref{fig:complex_tetra}, we note that the high complexity is not just given by the exterior of the shape, but also by all the intersection of the planes under the surface; both these make the function representing the object extremely hard to learn. Similar results are observed on the palm tree representation in Fig. \ref{fig:complex_tetra} with VF that reconstructs the thin leaves with higher fidelity and avoiding artifacts.

\begin{table}[ht]
    \scriptsize
    \centering
    \setlength{\tabcolsep}{2mm}
    \resizebox{0.90\columnwidth}{!}{
    \begin{tabular}{ c | c c c c c }\toprule
         \textbf{Method} & \textit{chairs} & \textit{lamps} & \textit{planes} & \textit{sofas} & \textit{cars} \\ \midrule
         SDF & 0.874 & 0.836 & 0.867 & 0.890 & 0.848 \\  
         UDF & 0.810 & 0.831 & 0.829 & 0.803 & 0.851 \\ \midrule
         \textbf{VF} & \textbf{0.898} & \textbf{0.877} & \textbf{0.919} & \textbf{0.910} & \textbf{0.918} \\ \midrule
         & \textit{cars} & \textit{busses} & \textit{lamps} & \textit{clothes} & \\ \midrule
         UDF & 0.809 & 0.869 & 0.837 & 0.876 & \\ \midrule
         \textbf{VF} & \textbf{0.893} & \textbf{0.915} & \textbf{0.886} & \textbf{0.909} & \\ \midrule
    \end{tabular}}
    \caption{Normal consistency evaluation computed as cosine similarity (higher is better). First line of results is on watertight meshes \cite{cheng2015shapenet} and the second one on general objects \cite{cheng2015shapenet,Bhatnagar_2019_ICCV}.}
    \label{tab:normcon}
\end{table}

Table \ref{tab:normcon} further highlights the better performance of VF in comparison to UDF and SDF, and shows that explicitly supervising the normal prediction significantly improves normal consistency. This is evaluated by computing the cosine similarity between the normal at the surface and the normal predicted with VF or obtained by differentiating SDF\footnote{Given that SDF does not flip direction at the surface, the direction is evaluated by always taking the orientation in agreement with the ground truth} or UDF. Across all categories, VF outperforms the distance based counterparts even in categories in which the Chamfer distance or the F1-Score were comparable.

\section{Adapted Marching Cubes}

A challenging part of implicit 3D representation is an easy transformation to standard mesh representation. Mesh allows rendering and manipulation of 3D shapes using standard graphics tools. For the purpose of going from implicit to mesh representation, a traditionally successful algorithm has been Marching Cubes (MC). Current MC algorithms are developed to produce smooth surfaces without holes and with continuously changing normals. This produces visually appealing results but constitutes a challenge when adapting the algorithm for different types of representation.

The standard MC algorithm is applied to a scalar field to produce triangles at the positions where the scalar field crosses a predefined value. Considering SDF, the value used is $0$ and the resulting polygonal surface approximates the zero level set of SDF. This is done by first voxelizing the space and evaluating the field values in these voxels. Furthermore, in order to choose between the multiple possible mesh configurations that arise from the vertex assignments, neighboring voxels are used to ensure surface continuity. To secure smoothly changing normals, the MC algorithms compute them based on a neighborhood around each surface position and locate the mesh faces in each voxel with a trilinear interpolation of the field.

\begin{figure}[!ht]
\centering
\begin{subfigure}{0.30\linewidth}
\includegraphics[width=\textwidth]{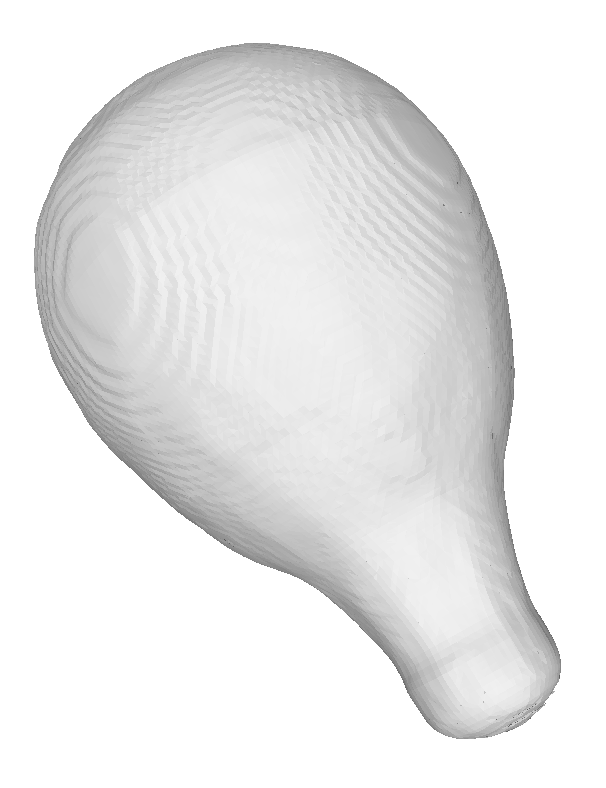}
\end{subfigure}
\hspace{0.1\linewidth}
\begin{subfigure}{0.30\linewidth}
\includegraphics[width=\textwidth]{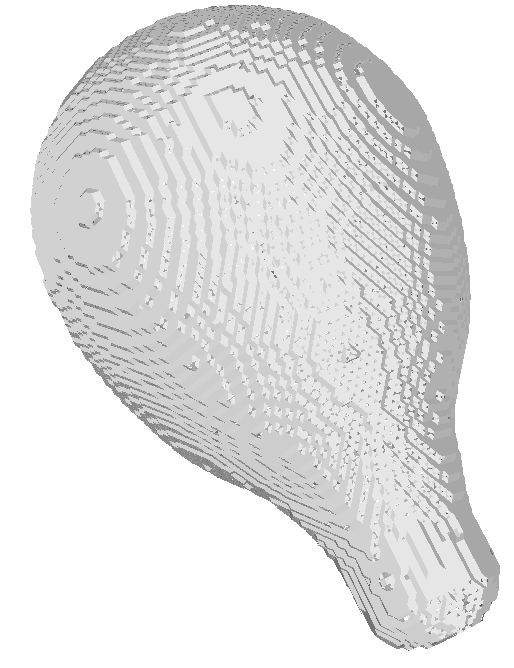}
\end{subfigure}
\caption{MC comparison on VF with (left) and without (right) the use of its norm. VF norm constitutes an effective measure of the distance from the surface and helps smooth the reconstructed meshes.}
\label{fig:mc_vt}
\end{figure}

In short, MC needs a way to define which voxels contain a surface, which vertices are inside and which outside the shape, and what is the distance of each vertex from the surface. For SDF all the three aspects are solved taking the distance: if there are vertices with different signs then there is a zero-crossing and a surface; the sign of the distance indicates whether a pixel is inside or outside; the absolute value of SDF is the distance from the surface.
To adapt the MC algorithm to VF, we now assess the three aspects just highlighted:
\begin{itemize}
    \item \textbf{Surface voxels}: as VF does not have a level set that identifies surface voxels that can be used in MC, we need to define a way to indicate such voxels. For this, we use Property 3.3 and identify the surface voxels as the ones that have a flux density smaller than or close to $-1$. In practice, we explicitly compute the flux density in each voxel and assign a flag to ones where the value is lower than $-0.7$\footnote{As explained in the main paper, we use $-0.7$ instead of $-1$ to account for small mis-predictions.}; those are surface voxels. From a computational standpoint, discrete flux density can be computed in a highly parallel manner inside the MC computation.
    
    \item \textbf{Inside-Outside clustering}: for each surface voxel, the directions of the vertices are then clustered into two groups, based on their cosine similarity. Then the two clusters are randomly selected to be considered inside or outside. In practice, the two clusters can be identified by taking the two vectors among the 8 of the voxel with the lowest cosine similarity between each other; these identify the two dominant opposite directions. The 6 remaining vectors are then associated to the cluster with which they share the highest cosine similarity. This is an effective clustering technique with very low computational cost.
        
    \item \textbf{Surface distance}: the final step to adapt is the use of distance values for the vertices to exploit the trilinear interpolation inside MC. In the case of VF, there is no exact representation of the distance to the surface as it does not explicitly encode it. However, we can use the continuity property of INRs; as the surface is defined by points where field directions flip, we observe that the norm of predicted VF is reduced around the surface points. We note that, even though this measure is not the actual distance, it monotonically changes close to the surface and hence can be used for our purpose. The same effect is also used when applying the traditional MC algorithm on the binary occupancy field~\cite{mescheder2019occnet}. Figure \ref{fig:mc_vt} shows that this can be effectively used for the purpose, as the resulting mesh is much smoother.
\end{itemize}
Our modified MC method is described in Algorithm~\ref{alg:MCvector}.

\begin{algorithm}[ht]
{\small
\caption{\small MCvector $(\mathcal{X},\mathsf{f}_{VF}(\mathsf{x}))$}
\label{alg:MCvector}
\begin{algorithmic}
 \STATE 1. Sample point set $\mathcal{X}\subset\mathbb{R}^3$ to form a voxel grid.\\
 \STATE 2. For each vertex $\mathsf{x}\in \mathcal{X}$, compute VF, compute the flux density, and flag voxels with a flux density value lower than $-0.7$. These are surface voxels.\\
 \STATE 3. For flagged voxels, cluster vertices into two groups using normals. One group of vertices is arbitrarily assigned to be inside and other the outside the surface.\\
 \STATE 4. Assign the norm of the predicted VF to the vertices and generate the surface.
\end{algorithmic}
}
\end{algorithm}

\end{document}